\pdfoutput=1

\documentclass[11pt]{article}

\usepackage{acl}

\usepackage{times}
\usepackage{latexsym}

\usepackage[T1]{fontenc}

\usepackage[utf8]{inputenc}

\usepackage{microtype}

\usepackage{inconsolata}
\usepackage{amsmath, amssymb}
\IfFileExists{makecell.sty}{\usepackage{makecell}}{}

\usepackage{amsfonts}  
\usepackage{graphicx}  
\IfFileExists{boldline.sty}{\usepackage{boldline}}{}  
\providecommand{\hlineB}[1]{\hline}
\IfFileExists{multirow.sty}{\usepackage{multirow}}{}  
\providecommand{\multirow}[3]{#3}
\usepackage{booktabs}  
\usepackage{array}     
\usepackage{xcolor}    
\IfFileExists{arydshln.sty}{\usepackage{arydshln}}{}

\IfFileExists{makecell.sty}{\usepackage{makecell}}{}
\IfFileExists{floatrow.sty}{\usepackage{floatrow}}{}  
\usepackage{subcaption} 
\usepackage{placeins} 
\usepackage{float} 
\IfFileExists{ulem.sty}{\usepackage[normalem]{ulem}}{} 


\newcommand\blfootnote[1]{%
  \begingroup
  \renewcommand\thefootnote{}\footnote{#1}%
  \addtocounter{footnote}{-1}%
  \endgroup
}

\title{Reasoning Depth and Environment Complexity: A Controlled Study of RLVR Data Allocation across Logical Reasoning Tasks}

\makeatletter
\renewcommand{\thefootnote}{\fnsymbol{footnote}}
\makeatother

\author{
Yihua Zhu$^{1,3}$ \qquad  Qianying Liu$^{3}$\textsuperscript{\dag} \qquad Fei Cheng$^{1}$  \qquad Jiaxin Wang$^{1}$ \qquad \\
\textbf{Akiko Aizawa}$^{2,3}$ \qquad \textbf{Sadao Kurohashi}$^{3}$ \qquad \textbf{Hidetoshi Shimodaira}$^{1,4}$
 \\
$^1$Kyoto University \qquad $^2$University of Tokyo \qquad $^3$NII LLMC \qquad $^4$RIKEN\\
\texttt{\{zhu.yihua.22h, wang.jiaxin.77y\}@st.kyoto-u.ac.jp}\\ 
\texttt{\{feicheng, shimo\}@i.kyoto-u.ac.jp} \texttt{\{ying, aizawa, kurohashi\}@nii.ac.jp}}

\date{}

\begin{document}

\maketitle

\footnotetext[2]{Corresponding author.}
\blfootnote{Our code is available at \url{https://github.com/YihuaZhu111/kg-rlvr-allocation}.}

\begin{abstract}

Reinforcement learning with verifiable rewards (RLVR) has become central to post-training reasoning models, yet a key limitation of existing studies is their narrow view of the reasoning space: difficulty is treated as reasoning depth alone, and reward is concentrated on forward deductive state tracking. We instead characterize the reasoning space along two dimensions. \textbf{Difficulty.} Beyond reasoning depth, we study environment complexity, where models must identify the correct path amid distractors and interacting structures. \textbf{Rewarded reasoning form.} We consider four abilities core to real-world reasoning: deductive state tracking, abductive recovery of hidden events or facts, inductive rule induction, and analogical transfer. To disentangle these factors, we construct a synthetic knowledge-graph environment with controlled pre- and post-training distributions, where each instance varies along depth, complexity, and task family. Three findings emerge: joint depth--complexity coverage outperforms single-axis recipes; reasoning families respond non-uniformly, with abductive reasoning degrading outside the RL-covered region and task correlations clustering into deductive--abductive and inductive--analogy pairs; and uniform mixing outperforms staged curricula under a fixed budget. 
We also find that recent off-the-shelf models exhibit the same deductive-over-abductive asymmetry, suggesting that this gap is not merely an artifact of our controlled setup.

\end{abstract}

\section{Introduction} \label{sec:intro}

\begin{figure*}[!t]
\centering
\includegraphics[width=\textwidth]{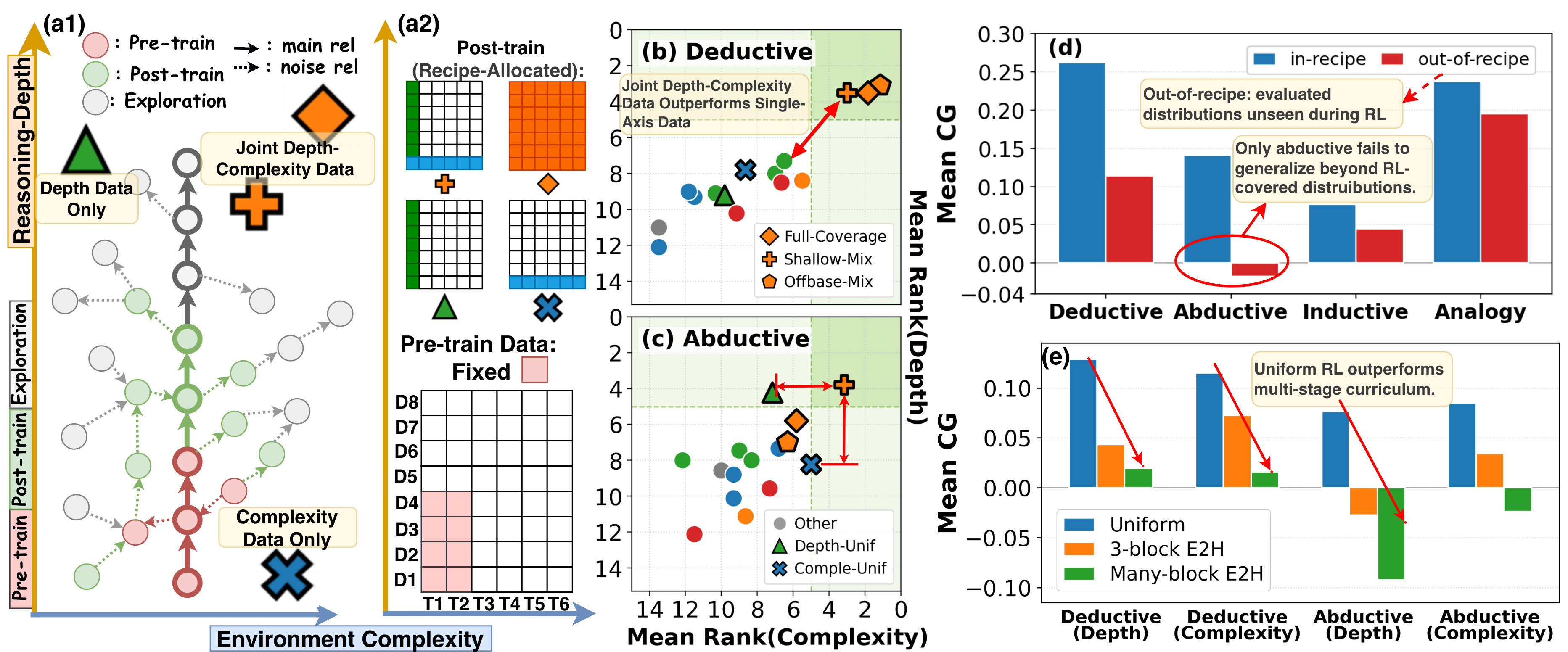}
\caption{Two-axis data distribution and three main findings.
(a) A knowledge-graph schematic shows reasoning depth and environment complexity.
(b,c) Finding 1: joint depth--complexity coverage outperforms single-axis recipes.
(d) Finding 2: abductive reasoning generalizes poorly outside the RL-covered data
distribution.
(e) Finding 3: uniform RL outperforms a multi-stage curriculum under a fixed budget.}
\label{fig:intro-main-findings}
\end{figure*}

Reinforcement learning with verifiable rewards has become a central post-training paradigm for improving the reasoning ability of language models. By rewarding outputs that can be automatically verified, RLVR has enabled substantial progress on mathematical reasoning, code generation, STEM problem solving, and long-horizon reasoning~\citep{guo2025deepseek,openai2026gpt55,anthropic2026opus47,
 openai2025o3o4mini}. However, the success of RLVR also exposes a more fundamental design question: \textbf{How should RL training examples be distributed across the reasoning space during post-training?} 
 This distribution determines which reasoning behaviors are reinforced and generalized, yet how to shape it across difficulty levels and forms of reasoning remains poorly understood.

A natural dimension of this distribution is reasoning difficulty. Existing post-training studies have examined difficulty through a depth-oriented lens, emphasizing longer reasoning chains, harder compositional structures, longer reasoning horizons, or training near the model's competence boundary~\citep{zhang2026on,lu2026rhorizon,yang2026depthbreadth,
huang2026on,wang2026can}. 
However, as illustrated in Figure~\ref{fig:intro-main-findings}(a), a model can fail along two distinct axes: the target reasoning path may be long, requiring many sequential inferences, or the path could be buried in a complex environment of distractors and interacting structures. Difficulty in RLVR data thus has two distinct axes: \textit{reasoning depth}, which stresses multi-step commitment, and \textit{environment complexity}, which stresses relevant-structure selection.
This motivates our first research question \textbf{RQ1: How should RL training examples be allocated across reasoning depth and environment complexity during RLVR post-training?} In particular, under a fixed post-training budget, should training data concentrate on one difficulty axis, jointly cover both axes, or follow a staged curriculum over the depth--complexity space?

Apart from difficulty, a second dimension is the form of reasoning that receives reward. Existing reasoning-oriented post-training mainly targets forward, deductive tasks such as mathematics and code~\citep{guo2025deepseek, abdin2025phi,tang2024mathscale}. However, logical reasoning is broader: abductive inference recovers hidden causes, inductive reasoning extracts rules from instances, and analogy transfers relational patterns across cases. This motivates our second research question \textbf{RQ2: Do different logical reasoning families behave uniformly under RLVR post-training?} We ask whether reasoning forms differ in how they use context, generalize beyond training coverage, and respond to the same post-training distribution.

Answering these questions with off-the-shelf models is difficult: their pre-training distributions are not observable, so post-training gains cannot be cleanly attributed to a specific RLVR data distribution. A controlled experimental setting is therefore necessary to isolate how that distribution shapes reasoning improvement.
We construct a synthetic knowledge-graph reasoning environment in which both pre-training and post-training distributions are explicitly specified. Each data instance is grounded in a knowledge graph that combines a dynamic event-state component with a static kinship component. As shown in Figure~\ref{fig:intro-main-findings}(a),
we vary two orthogonal dimensions of difficulty: reasoning depth \(D\), which controls the length of the target reasoning chain, and environment complexity \(T\), which controls the amount of distractor structure around that chain. Over this \(D\times T\) space, we define depth-focused, complexity-focused, jointly covered, and curriculum-based data recipes. 
From the same underlying world, we generate instances of four reasoning families: deductive state tracking, abductive recovery of missing events or facts, inductive rule induction, and analogy-based relation transfer. We first pre-train a model on shallow-depth, low-complexity regions, and then apply Group Relative Policy Optimization (GRPO)~\citep{shao2024deepseekmath} post-training under controlled data recipes that vary coverage over the \(D\times T\) grid and reasoning families.

This controlled setting yields three main findings. \textbf{Joint depth--complexity coverage produces broader RLVR gains.} (Figure~\ref{fig:intro-main-findings}(b,c)) Jointly covered recipes produce more robust gains across the grid space than depth-only or complexity-only recipes. \textbf{Reasoning families exhibit distinct post-training behavior.} The four task families are not interchangeable RL targets: the task-pair correlations cluster into deductive--abductive and inductive--analogy pairs. Abductive reasoning is distribution-sensitive and can even degrade outside the relevant training region (Figure~\ref{fig:intro-main-findings}(d)). Moreover, evaluation of off-the-shelf LLMs
reproduces such deductive-vs.-abductive asymmetry, performing poorly on abductive reasoning instances. \textbf{Uniform mixing is stronger than staged curricula under a fixed budget.} (Figure~\ref{fig:intro-main-findings}(e)) Easy-to-hard schedules help harder regions but lose global coverage. Together, these findings show that RLVR data distribution shapes not only how much reasoning improves, but also where improvement occurs, how it transfers, and which reasoning forms benefit.

\begin{figure*}[!t]
\centering
\includegraphics[width=\textwidth]{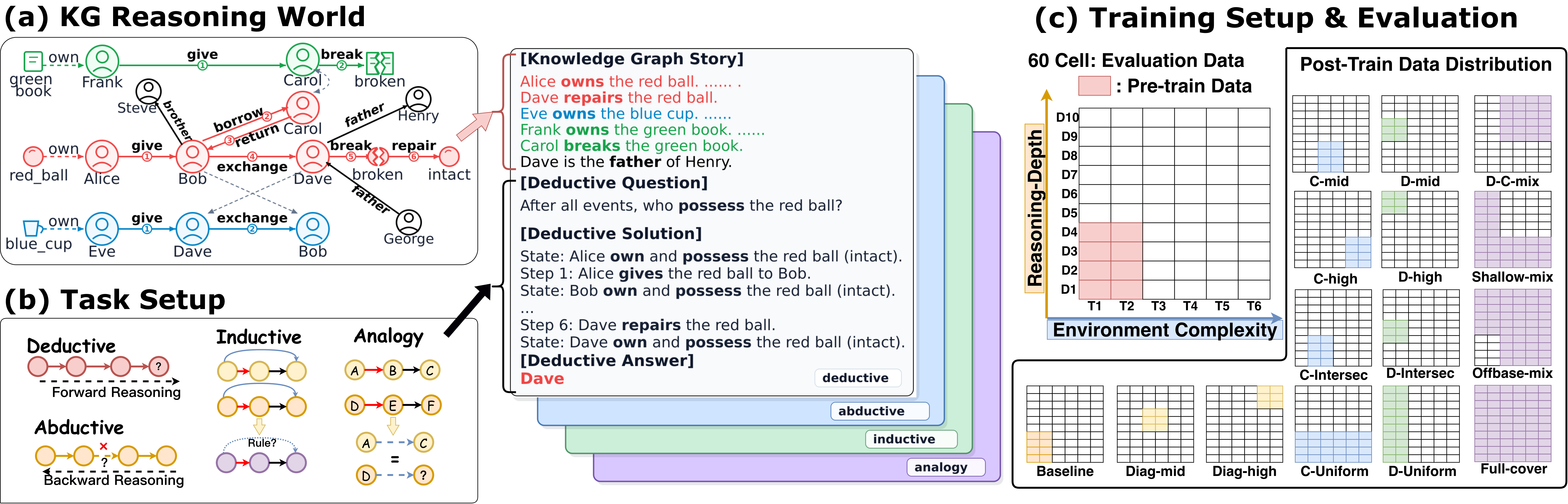}

\caption{Controlled KGRL framework. The figure summarizes the synthetic KG world,
four reasoning task families, and pre-training/RL post-training data distributions
used to study depth--complexity coverage.}
\label{fig:kgrl-framework}
\end{figure*}

\section{Methodology}
\label{sec:method}

\noindent We define a controlled KG reasoning framework. Appendices~\ref{app:data-generation-framework}
and~\ref{app:training-setup} give full data-generation and training details.

\subsection{Controllable KG Reasoning World}
\label{sec:controllable-kg-reasoning-dataset}

\paragraph{Graph Definition}
Each reasoning instance is grounded in a knowledge graph
\[
\mathcal{G} = (\mathcal{P}, \mathcal{O}, \mathcal{E}_{\mathrm{dyn}},
\mathcal{E}_{\mathrm{stat}}, s_0),
\]
where \(\mathcal{P}\) is a set of persons, \(\mathcal{O}\) is a set of objects,
\(\mathcal{E}_{\mathrm{dyn}}\) is a time-ordered sequence of dynamic events,
\(\mathcal{E}_{\mathrm{stat}}\) is a set of static kinship relations over persons, and
\(s_0\) is the initial world state.

\paragraph{Dynamic and Static Worlds}
The generator separates each story into dynamic and static components. The dynamic
event-state component updates object-centered facts such as ownership, possession, and
integrity, while the static kinship component stores fixed relations among people.
Figure~\ref{fig:kgrl-framework}(a) shows a toy instance: the red-ball chain tracks
events from Alice's initial ownership to Dave's final possession, and the black
kinship edges provide fixed relational facts in the same story.

\subsection{Task Setup}
\label{sec:task-setup}

\paragraph{Logical Reasoning Task}
We instantiate four logical task families from the same controlled graph world.
Deductive and abductive tasks use the dynamic component: deductive tasks track observed events
forward to a queried state value, while abductive tasks recover a missing event or
initial fact consistent with the observed states. Inductive and analogy tasks use the
static component: inductive tasks infer a shared relation rule from support examples,
while analogy tasks transfer a relation pattern from one entity pair to another.

\paragraph{Reasoning Depth}
Reasoning depth counts dynamic operation events on the causal chain of the target
object:
\[
D(\mathcal{G}) =
\left|\{e_i \in \mathcal{A}_{\mathrm{op}} : o^\star \in \mathrm{Obj}(e_i)\}\right|.
\]
Here \(o^\star\) is the target object, \(\mathrm{Obj}(e_i)\) is the set of objects
affected by event \(e_i\), and \(\mathcal{A}_{\mathrm{op}}\) is the set of dynamic
operation events.
Figure~\ref{fig:intro-main-findings}(a) visualizes this as the vertical reasoning-depth
axis.

\begin{figure*}[t]
\centering
\includegraphics[width=\textwidth]{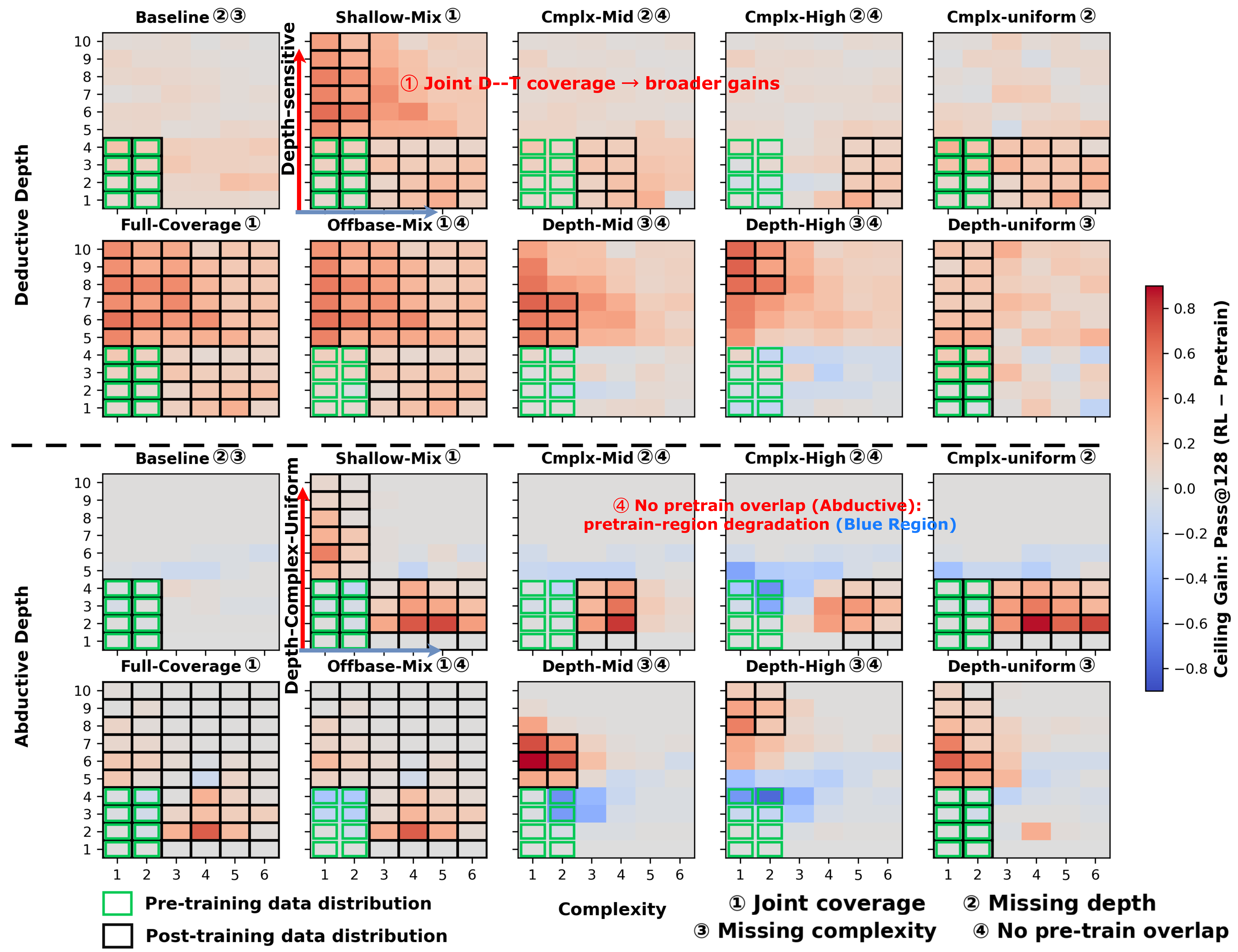}
\caption{Cell-level CG heatmaps on the 60-cell \((D,T)\) grid. Each panel
corresponds to one RL recipe; warmer colors and larger CG values indicate stronger
post-training gains, while blue indicates degradation below the pre-trained model.}
\label{fig:q1-marginal-heatmaps}
\end{figure*}

\paragraph{Environment Complexity}
The environment complexity \(T(\mathcal{G}) \in \{1,\ldots,6\}\) indexes a structural
parameter map:
\[
\begin{aligned}
\Theta(T) &=
(k_{\mathrm{chains}}, n_{\mathrm{persons}}, \rho_{\mathrm{crossover}},
\rho_{\mathrm{exchange}}).
\end{aligned}
\]

Here \(\Theta(T)\) controls the number of parallel chains, person scale, cross-chain
entity overlap, and exchange frequency. Higher \(T\) adds distractors and cross-chain
interactions around the target chain. Figure~\ref{fig:intro-main-findings}(a) provides a
schematic example.

\subsection{Multi-Grained Evaluation}
\label{sec:multi-grained-evaluation}

\paragraph{Strict Correctness}
Following prior work~\citep{zhang2026on}, we evaluate every generated solution
with a strict process-verified criterion. Given gold trace \(\pi\), gold answer
\(a^\star\), predicted trace \(\hat{\pi}\), and predicted answer \(\hat{a}\), a
solution is correct only when both process and answer match:
\[
\mathrm{correct}(\hat{\pi},\hat{a}\mid \pi,a^\star)
=
\mathrm{m}_{\mathrm{P}}(\hat{\pi},\pi)
\wedge
\mathrm{m}_{\mathrm{A}}(\hat{a},a^\star).
\]
This rule applies to all four task families. Here \(\mathrm{m}_{\mathrm{P}}\) is
task-specific process matching, and \(\mathrm{m}_{\mathrm{A}}\) is answer matching. We
report strict process-verified \(\mathrm{pass@}k\) for \(k\in\{1,\ldots,128\}\).
Ceiling Gain (CG) is the post-training gain in sampled reasoning ceiling:
\(\mathrm{CG}=\mathrm{pass@}128_{\mathrm{post}}-\mathrm{pass@}128_{\mathrm{pre}}\).
Single Gain (SG) is the corresponding single-sample gain:
\(\mathrm{SG}=\mathrm{pass@}1_{\mathrm{post}}-\mathrm{pass@}1_{\mathrm{pre}}\).

\paragraph{Grid-Level Reporting}
The evaluation split contains 4,800 graphs, with 80 graphs in each of the 60
\((D,T)\) cells. We report cell-level results before aggregating by depth, complexity,
task family, or recipe family, since post-training can improve one region while
suppressing another. Detailed verdict rules and coarser plane summaries are deferred
to Appendices~\ref{appx:experiment-evaluation-metrics}
and~\ref{appx:supplementary-experiments-results}.

\subsection{Training Setup}
\label{sec:train-setup}
We conduct experiments with a 107M-parameter Qwen2.5-style decoder-only
model~\citep{qwen2025qwen25technicalreport}. All stages use the same controlled KG generator and differ only
in \((D,T)\) coverage and per-cell budgets. Full model,
optimizer, recipe, and implementation details are given in
Appendix~\ref{app:training-setup}.

\paragraph{Pre-Training}
The pre-training split covers the shallow and low-complexity block
\(D\in\{1,\ldots,4\}\) and \(T\in\{1,2\}\), with 3.2 million graphs and 3.14B tokens
in total. This region teaches the model basic reasoning ability in our domain while
leaving deeper chains and higher-complexity environments unseen. This lets us test
whether RL post-training extends the model beyond the pre-training distribution.

\paragraph{Post-Training}
Post-training starts from the pre-trained model and uses
GRPO~\citep{shao2024deepseekmath}. We ask which \((D,T)\) regions should provide RL
signal after pre-training. We study 15 recipes: one baseline and 14 additional
data-distribution recipes in four families, as detailed in Figure~\ref{fig:kgrl-framework}(c).

Each recipe uses a fixed budget of 220K tasks, distributed uniformly over its selected
\((D,T)\) cells. The budget is split into 140K, 60K, 10K, and 10K tasks for deductive,
abductive, inductive, and analogy, respectively. This allocation reflects our intended
reasoning emphasis: deductive receives the largest share because much real-world
reasoning is deductive; abductive keeps a moderate share as the inverse dynamic
counterpart of deductive state tracking; inductive and analogy receive smaller shares
because they operate mainly on the static component and are less affected by reasoning
depth and environment complexity, making them less dependent on extensive RL training.

During GRPO, \(A=\mathrm{m}_{\mathrm{A}}(\hat{a},a^\star)\in\{0,1\}\) indicates
answer match and \(P=\mathrm{m}_{\mathrm{P}}(\hat{\pi},\pi)\in\{0,1\}\) indicates
full parsed-trace match. The outcome-gated reward is
\[
R =
\begin{cases}
0.8P+0.2A, & A=1,\\
0, & A=0.
\end{cases}
\]
Thus a response receives no reward when its final answer is wrong, even if the trace is
partially plausible.

\section{How Should Post-Training Data Be Distributed Across Reasoning Depth and Environment Complexity?}
\label{sec:recipe-design}

We study the question from global and local views. The main text focuses on deductive
and abductive; inductive and analogy results are reported in
Appendix~\ref{appx:inductive-analogy-experiments}.

\subsection{Task Setting}
For each task family, we ask which RL data distribution gives the strongest global
improvement across the 60-cell \((D,T)\) grid, and which recipes improve or suppress each
individual cell.

\paragraph{Task 1: Global Recipe Ranking}
We rank recipes along the depth and complexity axes using CG. For each task, recipes are
ranked by mean CG within every depth or complexity slice, and the slice-level ranks are
then averaged. For example, the depth slice \(D=1\) contains all cells with
\(T=1,\ldots,6\); averaging ranks over all 10 depth slices gives each recipe's mean
depth-rank.

\paragraph{Task 2: Local Cell Improvement}
The local view keeps full grid resolution: for every task family and valid \((D,T)\)
cell, we compute CG against the pre-trained model.

\subsection{Observations}
\paragraph{Task 1.}
Figure~\ref{fig:intro-main-findings}(b, c) shows that stronger recipes lie closer to the
upper-right corner. For deductive, Shallow-Mix, Offbase-Mix, and Full-Coverage outrank
all other RL recipes, including Cmplx-Uniform and Depth-Uniform, on both axes. For
abductive, Shallow-Mix beats Cmplx-Uniform on depth and Depth-Uniform on complexity,
while remaining competitive on each baseline's covered axis.

\paragraph{Task 2.}
Figure~\ref{fig:q1-marginal-heatmaps} shows that Shallow-Mix improves over Cmplx-Uniform
and Depth-Uniform on nearly all deductive cells, with the largest margins on deeper
low-complexity cells. On abductive, Shallow-Mix again gives broader gains than either
single-axis recipe, though with smaller margins. Paired bootstrap intervals confirm
that the overall Shallow-Mix advantage exceeds evaluation-sampling noise, while its
abductive slice is at parity (Appendix~\ref{appx:bootstrap}).

The heatmaps also reveal different axis sensitivities. Deductive gains are larger along
the depth extension than along the complexity extension, and Depth-family recipes yield
larger CG gains than Cmplx-family recipes. Abductive gains are more balanced across the
two axes, with no clear preference for either.

\subsection{Takeaways}
\paragraph{Post-training data that jointly cover reasoning depth and environment complexity outperform data that cover only one axis.}

\paragraph{Deductive benefits more from depth coverage, while abductive has no clear axis preference.}

\subsection{Discussion}
\paragraph{Joint depth--complexity coverage is necessary.}
Depth tests whether intermediate commitments can be preserved over longer chains,
whereas complexity tests relevant-structure selection under distractors. Single-axis
recipes reward only one demand, while joint coverage rewards both.

Figure~\ref{fig:q1-marginal-heatmaps} shows that Baseline and the
Complexity-/Depth-High recipes are generally weaker than their Mid counterparts,
consistent with the view that RL gains depend on headroom near the model's competence boundary~\citep{zhang2026on}. Our result extends this view from a single difficulty axis to a two-dimensional depth--complexity region.

\begin{figure*}[t]
\centering
\includegraphics[width=\textwidth]{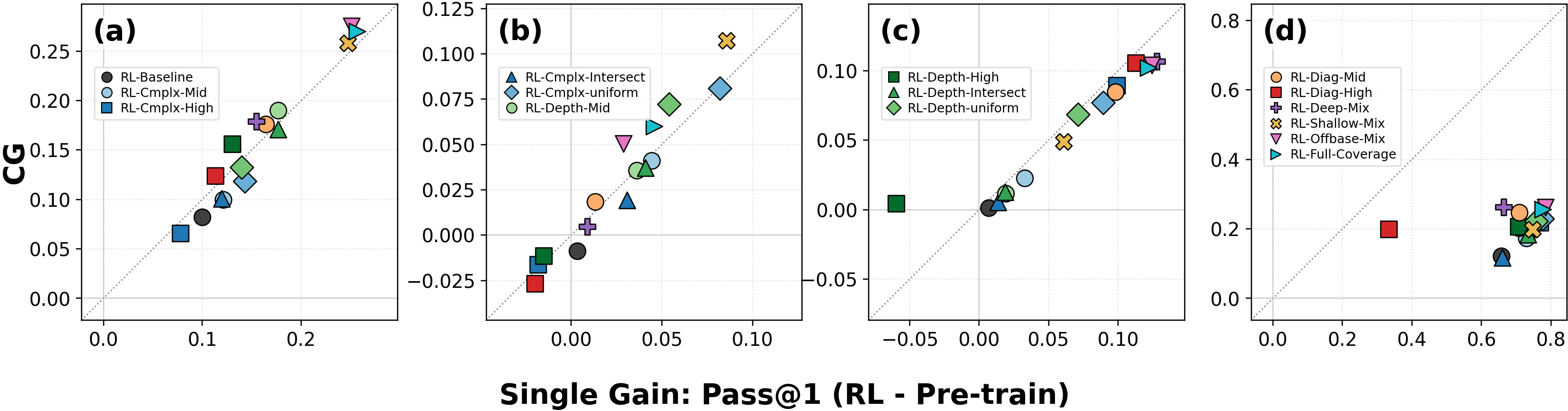}
\caption{SG/CG plane across task families. Each point is one RL recipe. Deductive, abductive, and inductive recipes lie near
the SG \(=\) CG diagonal, whereas analogy recipes lie in the high-SG, low-CG region.}
\label{fig:task-asymmetry-sg-cg}
\end{figure*}

\paragraph{Deductive is depth-sensitive, while abductive is more balanced.}
A deductive answer applies the next operation forward from the current state. Abductive
reasoning instead works backward: at each hop, the model must reconstruct adjacent
states, match them, and infer the missing operation from their difference. Environment
complexity makes this matching harder because distractor operations create many
competing post-operation states. Environment-complexity training is therefore more
necessary for abductive reasoning, whereas deductive reasoning mainly benefits from
depth-oriented training and uses complexity for robustness.

\subsection{Practical Guidance}
Post-training data should cover both reasoning depth and environment complexity. For
hard deductive targets, allocation can lean toward deeper cells, but it should still
retain complexity coverage rather than collapse to a depth-only recipe.

\section{Do All Reasoning Tasks Behave Uniformly Under Post-Training?}
\label{sec:task-asymmetry}

We next test whether the four reasoning families respond uniformly to post-training.

\subsection{Task Setting}
We evaluate out-of-recipe generalization, task-pair correlation, and
reasoning-capability growth.

\paragraph{Task 1: Out-of-Recipe Generalization}
For each task family, we compute CG for 15 post-training recipes over 60 cells,
yielding 900 recipe-cell units. Covered units are in-recipe; the rest are
out-of-recipe. Under the Baseline recipe, for example,
\(D=1, T=2\) is in-recipe, whereas \(D=1, T=3\) is out-of-recipe.

\paragraph{Task 2: Task Correlation}
For each recipe-cell unit, we obtain CG for all four task families and compute Pearson
correlation for the six task pairs over the corresponding 900 CG pairs. Appendix~\ref{appx:experiment-evaluation-metrics}
gives the formal definition.

\paragraph{Task 3: SG/CG Diagnostic}
For each task family, we compute SG and CG for each of the 15 recipes on the full
60-cell grid.

\subsection{Observations}
\paragraph{Task 1.}
Figure~\ref{fig:intro-main-findings}(d) shows positive out-of-recipe CG for deductive,
inductive, and analogy, about 50--70\% of their in-recipe gains, whereas abductive is
negative. Figure~\ref{fig:q1-marginal-heatmaps} shows the same abductive failure:
without overlap between post-training and pre-training cells, blue regions near the
pre-training block indicate degradation below the pre-trained model; pooled over
out-of-recipe cells, this degradation is significant for 6 of 10 tested recipes
(Appendix~\ref{appx:bootstrap}).

\begin{figure}[t]
\centering
\includegraphics[width=\columnwidth]{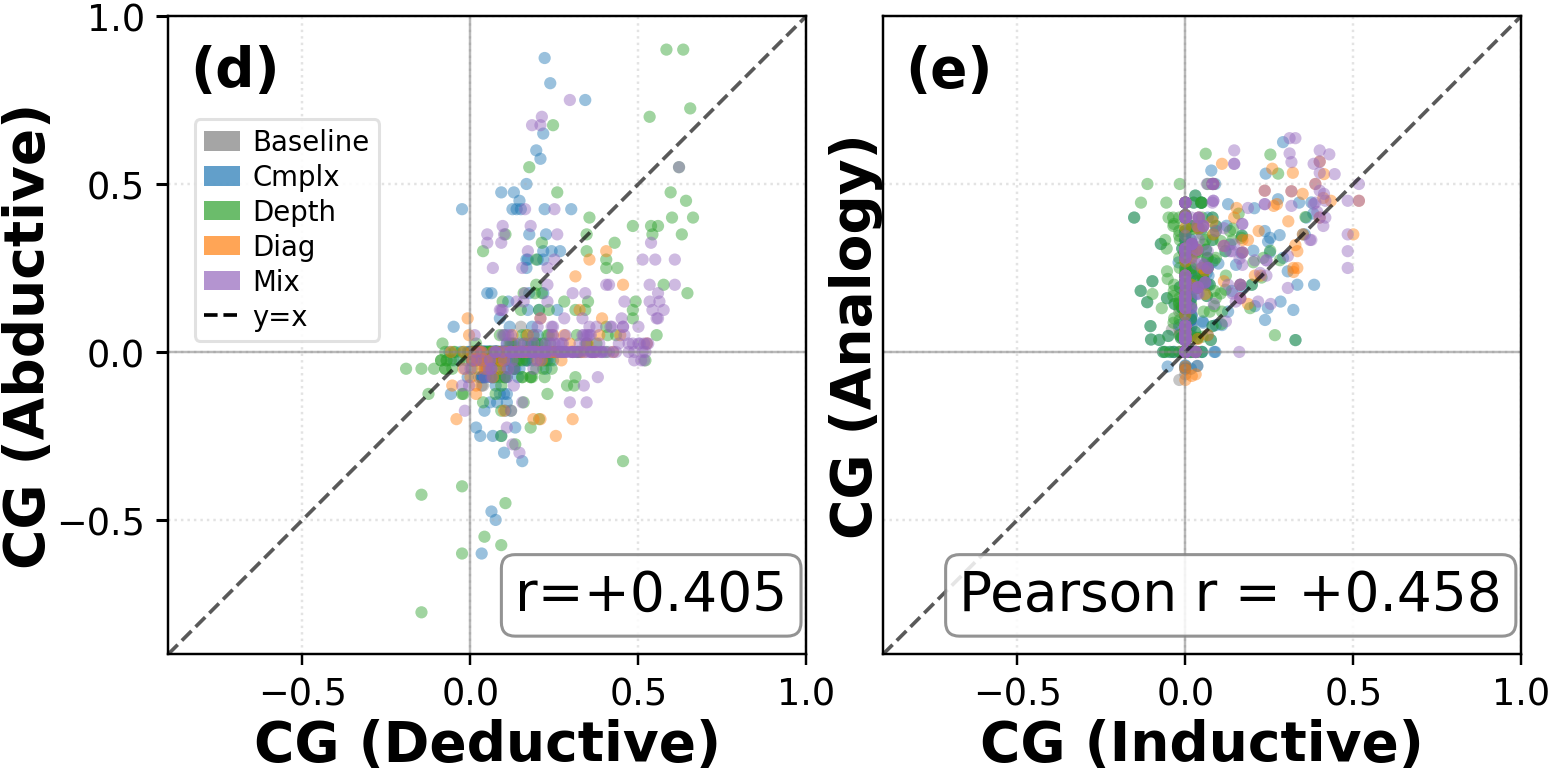}
\caption{Strong task-pair CG correlations across the 60-cell grid and 15 RL recipes.
Deductive--abductive (left) and inductive--analogy (right) are the strongest positive
pairs; panel letters follow Figure~\ref{fig:task-pair-correlation-remaining}.}
\label{fig:task-pair-correlation-strong}
\end{figure}

\paragraph{Task 2.}
Figures~\ref{fig:task-pair-correlation-strong} and
\ref{fig:task-pair-correlation-remaining} identify deductive--abductive and
inductive--analogy as the strongest positive pairs; deductive--analogy is moderate,
and the remaining three pairs are weak.

\paragraph{Task 3.}
Figure~\ref{fig:task-asymmetry-sg-cg} places deductive, abductive, and inductive near
the SG \(=\) CG diagonal, but analogy in the high-SG, low-CG region.

\subsection{Takeaways}
\paragraph{Abductive post-training shows almost no generalization beyond the data coverage, while the other three tasks generalize well.}

\paragraph{Task correlations concentrate in deductive--abductive and inductive--analogy.}

\paragraph{Deductive, abductive, and inductive exhibit healthy growth in reasoning ability during post-training, whereas analogy shows a mode-collapse phenomenon.}

\subsection{Discussion}
\paragraph{Abductive requires direct coverage.}
The weak out-of-recipe transfer of abductive reasoning is consistent with its reliance
on multi-hop reverse inference, as discussed in Section~\ref{sec:recipe-design}.
Prior work shows that reversal remains difficult for autoregressive language models
without sufficient training support~\citep{berglund2024reversal,lv-etal-2024-analysis,
zhu2024towards,golovneva2024reverse,zhu-etal-2026-memorization}; however, these studies
mainly consider one-hop or fact-level reversal. Our setting imposes a stronger
requirement: the model must reconstruct missing steps across multiple hops while
resolving competing states induced by high environment complexity. Abductive gains
therefore depend strongly on direct coverage of the target data distribution.

\paragraph{Task correlations reflect shared reasoning components.}
The deductive--abductive correlation arises from a shared dynamic transition system
and from the deductive subproblem embedded in abductive reasoning: predicting the state
after the previous operation is a forward state-prediction step. The
inductive--analogy correlation reflects shared reliance on kinship relations and rule
internalization, since solving an analogy instance requires extracting the relation
pattern from a source case and transferring it to a target case. The weaker
deductive--analogy coupling is consistent with their common entity/state-to-next-state
prediction structure.

\paragraph{Analogy shows mode collapse.}
Deductive, abductive, and inductive gains require improving the underlying reasoning
process, so pass@1 and the sampled candidate set tend to rise together. Analogy often
admits a compact mapping pattern; RL can sharpen this dominant pattern and improve
pass@1 without expanding the sampled candidate set, yielding high SG but low CG.

\subsection{Practical Guidance}
For abductive, directly cover the target evaluation region rather than assuming
transfer beyond data coverage. For multi-task post-training, first pair deductive with
abductive and inductive with analogy before testing broader mixtures.

\section{Does Staged Curriculum Improve Over Uniform Post-Training?}
\label{sec:curriculum}

We compare uniform mixing with staged curricula under a fixed post-training budget.

\paragraph{Task Setting}
All schedules use the same post-training budget and differ only in how target cells
are partitioned. We split the target region along either the depth or complexity axis
and train from easy to hard. For example, on the complexity axis, Cmplx-Uniform uses
all 24 target cells in one stage; 3-block uses
\(T1\)--\(T2 \rightarrow T3\)--\(T4 \rightarrow T5\)--\(T6\); and 6-block uses one
complexity level per stage.

\paragraph{Observations}
Figure~\ref{fig:intro-main-findings}(e) shows that, for both deductive and abductive
tasks and along both the depth and complexity axes, uniform mixing outperforms 3-block
E2H, which in turn outperforms 6-block E2H. Both uniform-versus-3-block gaps exceed
evaluation-sampling noise (Appendix~\ref{appx:bootstrap}).

\paragraph{Discussion}
\label{sec:curriculum-discussion}

\textbf{Uniform mixing is the strongest default under a fixed budget.}
Each curriculum transition introduces forgetting and adaptation costs: updates on a
new block can overwrite support for earlier blocks, and part of the budget is spent
adapting to the next sub-distribution before useful learning resumes. These costs can
be offset only when each block receives enough compute and later blocks build on
earlier scaffolding, which is not guaranteed under a fixed post-training budget.
Recent LLM-RL curricula, including VCRL \citep{jiang2025vcrl}, Re-Schedule
\citep{wang2026scheduling}, and the E2H Reasoner \citep{parashar2026curriculum}, report
gains over random or unscheduled RL baselines, but do not fix the total budget. Under
fixed compute, per-transition forgetting \citep{ibrahim2024simple} can dominate. This
is also consistent with curriculum theory \citep{bu2026provable}, whose
sample-complexity gains assume that per-block capacity is not budget-limited.

\paragraph{Practical Guidance}
Default to uniform mixing.

\section{Benchmarking Off-the-Shelf Reasoning Models}
\label{sec:benchmark-evaluation}

We benchmark recent general-purpose and reasoning-oriented language models on the
same 60-cell evaluation set to test whether the RL task-family asymmetries also
appear in off-the-shelf and frontier models. Each model uses task- and cell-matched
3-shot prompting: for a target task at depth \(D\) and environment complexity \(T\),
the prompt contains three random RL-data examples from the same task family and
\((D,T)\) cell. 

\begin{table}[t]
\centering
\scriptsize
\setlength{\tabcolsep}{2pt}
\resizebox{\columnwidth}{!}{%
\begin{tabular}{lccccc}
\toprule
Model & Overall & Ded. & Abd. & Ind. & Ana. \\
\midrule
Kimi-K2.6 & \textbf{.748} & \textbf{.831} & \textbf{.262} & \underline{.746} & \underline{.649} \\
GPT-5.4 & \textbf{.743} & \textbf{.827} & \textbf{.265} & .743 & .617 \\
GLM-5.1-FP8 & \textbf{.742} & \textbf{.805} & \textbf{.264} & \textbf{.797} & \underline{.700} \\
\textbf{Offbase-Mix} & \underline{.585} & .563 & .177 & \textbf{.822} & \textbf{.865} \\
\textbf{Shallow-Mix} & \underline{.575} & .559 & \underline{.235} & \underline{.759} & \underline{.823} \\
DeepSeek-R1 & \underline{.560} & \underline{.615} & .135 & .614 & .522 \\
Qwen3.6-35B-A3B & .558 & \underline{.651} & .089 & .526 & .402 \\
Mixtral-8x22B-Instruct & .550 & \underline{.607} & .099 & .581 & .585 \\
Kimi-K2-Thinking & .543 & .578 & .169 & .609 & .610 \\
Qwen3.5-397B-A17B-FP8 & .521 & .552 & .162 & .567 & .634 \\
\textbf{Cmplx-Uniform} & .510 & .453 & \underline{.227} & \textbf{.787} & \textbf{.850} \\
\textbf{Depth-Uniform} & .503 & .452 & \underline{.202} & \underline{.770} & \textbf{.835} \\
Qwen3.5-35B-A3B & .494 & .539 & .097 & .565 & .475 \\
QwQ-32B-Preview & .475 & .594 & .078 & .270 & .347 \\
Qwen2.5-32B-Instruct & .427 & .430 & .085 & .603 & .494 \\
Qwen3-8B & .344 & .391 & .051 & .357 & .284 \\
DS-R1-Distill-Qwen-32B & .301 & .256 & .078 & .535 & .544 \\
Qwen2.5-72B-Instruct & .247 & .145 & .092 & .689 & .477 \\
Qwen3.6-27B & .220 & .209 & .183 & .269 & .264 \\
Qwen2.5-7B-Instruct & .203 & .219 & .032 & .201 & .302 \\
Qwen3.5-122B-A10B-FP8 & .176 & .106 & .013 & .522 & .316 \\
DS-R1-Distill-Qwen-7B & .138 & .180 & .015 & .063 & .079 \\
\bottomrule
\end{tabular}%
}
\caption{Benchmark results on the 60-cell evaluation set with task- and cell-matched
3-shot prompting. Current LLMs remain weak on abductive tasks.}
\label{tab:benchmark-leaderboard-calibrated}
\end{table}

Table~\ref{tab:benchmark-leaderboard-calibrated} shows three patterns. Kimi-K2.6,
GPT-5.4, and GLM-5.1-FP8 are the strongest overall models, with very close scores.
Nearly all models are much stronger on deductive than abductive tasks; even the best
abductive scores remain low, suggesting weaker post-training pressure on reverse
inference. Despite their 107M scale, our RL checkpoints remain competitive:
Offbase-Mix and Shallow-Mix rank highly overall, while Cmplx-Uniform and
Depth-Uniform are especially strong on inductive and analogy tasks.

\section{Ablation Study}
\label{sec:ablation}

We ablate four factors held fixed in the main experiments: post-training
budget, GRPO rollout group size, model scale, and model backbone.

\subsection{Data Volume}
\label{sec:ablation-data-volume}
\begin{figure}[t]
\centering
\includegraphics[width=\columnwidth]{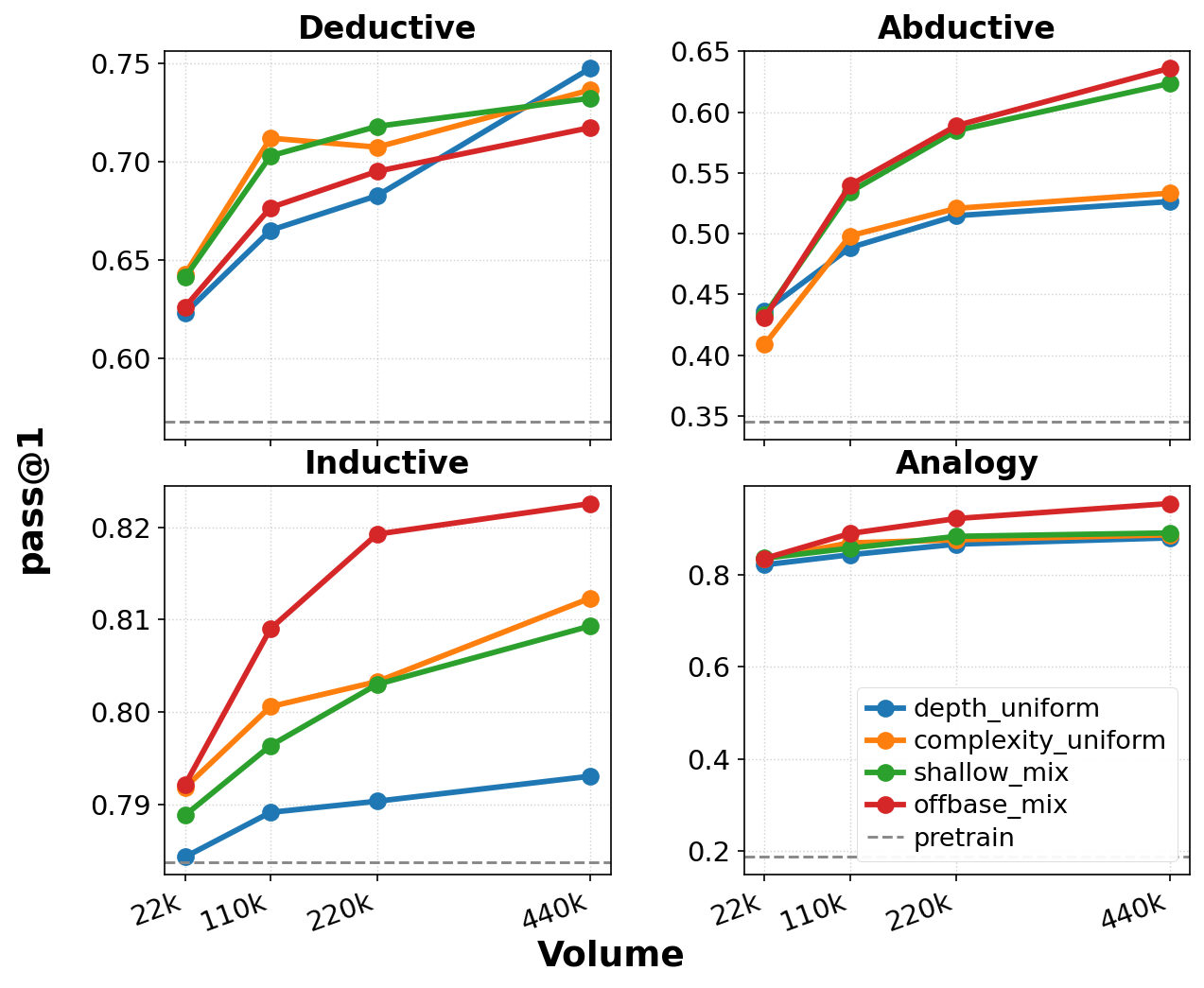}
\caption{Data-volume ablation: single-sample pass@1 per task family versus
post-training budget (22K--440K) for four recipes; dashed: pre-trained model.}
\label{fig:ablation-data-volume}
\end{figure}

Figure~\ref{fig:ablation-data-volume} varies the post-training budget from 22K
to 440K tasks for four recipe families. All task families improve with budget,
though not strictly monotonically and with task-dependent magnitudes. The 220K
budget captures most of the gain, while 440K still helps at a higher data
cost, so we use 220K as a practical middle point rather than a saturated
regime; all recipe comparisons in
Sections~\ref{sec:recipe-design}--\ref{sec:curriculum} assume this budget.

\subsection{Number of Rollouts}
\label{sec:ablation-rollouts}
\begin{figure}[t]
\centering
\includegraphics[width=\columnwidth]{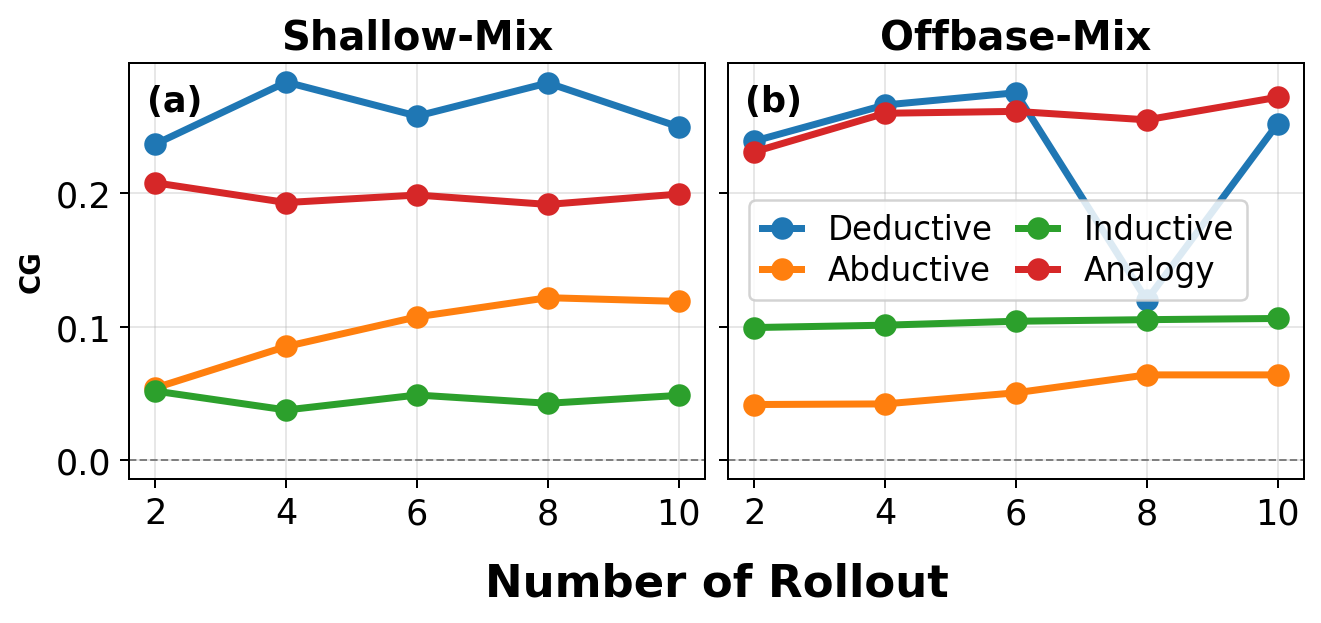}
\caption{Rollout-count ablation: CG versus GRPO rollout group size \(N\) for
Shallow-Mix and Offbase-Mix.}
\label{fig:ablation-rollout-count}
\end{figure}

Figure~\ref{fig:ablation-rollout-count} varies the GRPO rollout group size
\(N\in\{2,4,6,8,10\}\) for Shallow-Mix and Offbase-Mix. Larger groups do not
yield monotonic gains: Shallow-Mix peaks at \(N=8\), whereas Offbase-Mix is
less stable under larger groups and shows a sharp deductive drop at \(N=8\).

\subsection{Model Scale}
\label{sec:ablation-scale}
To test whether the per-family pattern is specific to 107M, we repeat the full
pipeline at 51M, 524M, and 1.02B with the same Qwen2.5-style architecture
(Appendix~\ref{app:scale-backbone-configs}). In Table~\ref{tab:scale-backbone}
(strict single-sample pass@1), deductive rises from 46.6 to 65.5 and abductive from
20.7 to 33.5 over the roughly \(20\times\) sweep, yet abductive stays far below
the other families at every scale: its weakness is not resolved by scale alone.
All families improve monotonically from 107M upward; the only exception is a
small analogy dip from 51M to 107M (85.4 to 82.3). Covering one recipe and
single-sample pass@1, these runs test the family-level pattern, not the recipe
rankings or the SG/CG diagnostic, which remain 107M results.

\begin{table}[t]
\centering
\footnotesize
\setlength{\tabcolsep}{5pt}
\begin{tabular*}{\columnwidth}{@{\extracolsep{\fill}}lcccc@{}}
\toprule
Model & Ded. & Abd. & Ind. & Ana. \\
\midrule
\multicolumn{5}{l}{\textit{Model scale (Qwen2.5-style)}} \\
51M & 46.6 & 20.7 & 75.6 & 85.4 \\
107M (main) & 55.9 & 23.5 & 75.9 & 82.3 \\
524M & 62.4 & 31.9 & 82.4 & 95.5 \\
1.02B & 65.5 & 33.5 & 87.6 & 96.2 \\
\midrule
\multicolumn{5}{l}{\textit{Model backbone (matched scale)}} \\
GPT-2-style 104M & 49.6 & 26.5 & 73.5 & 75.6 \\
Llama-style 105M & 53.4 & 21.8 & 80.8 & 81.7 \\
\bottomrule
\end{tabular*}
\caption{Model-scale and backbone ablations under the Shallow-Mix recipe:
strict single-sample pass@1 (\%) per task family on the 60-cell evaluation set. Each
row repeats pre-training from scratch and GRPO post-training; 107M is the
main-paper model.}
\label{tab:scale-backbone}
\end{table}

\subsection{Model Backbone}
\label{sec:ablation-backbone}
We also repeat the pipeline on two backbones at matched scale: a GPT-2-style
model (104M; 14 layers, multi-head attention) and a Llama-style model (105M)
structurally identical to our Qwen2.5-style model except for its architecture
class. In Table~\ref{tab:scale-backbone}, every family stays within a narrow
band across the three backbones (spread at most 7.3 points), and abductive is
the weakest for each, so the per-family conclusions are not an artifact of one
model family.

\section{Related Work}
\label{sec:related-work}

\paragraph{Controlled Experiments from Scratch}
Controlled training from scratch with synthetic data studies language models under
known data-generating processes, reducing opaque pre-training exposure and benchmark
contamination as attribution confounds~\citep{magar-schwartz-2022-data,
sainz-etal-2023-nlp,roberts2024to}. Recent controlled studies use synthetic reasoning
data to isolate the effects of pre-training, mid-training, RL, relational semantics,
and small-model emergence~\citep{zhang2026on,zhu-etal-2026-memorization,
eldan2023tinystories,allen-zhu2025physics}. 

\paragraph{Depth-Oriented RLVR and Difficulty Allocation}
Recent RLVR and post-training work often frames difficulty through depth-like signals:
longer chains, compositional depth, long horizons, competence boundaries, and
curriculum schedules~\citep{zhang2026on,lu2026rhorizon,yang2026depthbreadth,
huang2026on,wang2026can,jiang2025vcrl,wang2026scheduling,
parashar2026curriculum,bu2026provable,kung2026learning}. These studies clarify
when RL sharpens or extends reasoning, but usually treat difficulty as scalar or
depth-like. Controlled evaluations show that distractors, symbolic perturbations, and
complexity-OOD shifts also affect reasoning~\citep{mirzadeh2025gsmsymbolic,
samiei2026bridging,wang2026can}. We instead make environment complexity a
post-training allocation axis, not only an evaluation stressor.

\paragraph{Logical Reasoning Tasks and Benchmarks}
Logical reasoning work evaluates LLMs on deductive and first-order benchmarks,
proof-style tasks, and synthetic logic data~\citep{parmar-etal-2024-logicbench,han-etal-2024-folio,
tafjord-etal-2021-proofwriter,liu2025synlogic}. Other studies examine abductive reasoning and
reverse inference through abductive benchmarks~\citep{Bhagavatula2020Abductive,
cao-etal-2026-semeval,han2026molquest}, reversal analyses~\citep{berglund2024reversal,
lv-etal-2024-analysis,zhu2024towards}, or analogy transfer~\citep{webb2023emergent,musker2025semantic}. Most use these families as benchmarks or study one form at a time. We use our four-task dataset for both RLVR training and benchmark evaluation.

\section{Conclusion}
\label{sec:conclusion}

RLVR recipe design is a data-allocation problem over both difficulty axes and
reasoning forms. Using a controlled synthetic KG environment, we show that joint
depth--complexity coverage is more reliable than single-axis recipes, that task
families respond differently to the same post-training data, and that abductive
reasoning is especially fragile outside the training distribution. Task correlations separate into deductive--abductive and inductive--analogy pairs, while uniform mixing outperforms a staged curriculum under a fixed budget. Recent instruct/reasoning models show the same deductive-over-abductive gap, suggesting that RLVR post-training should optimize structured coverage rather than only aggregate benchmark gains.

\clearpage
\section{Limitations}

\paragraph{Synthetic KG setting.}
Our synthetic KG world provides strong control over pre-training, post-training, and
evaluation distributions, but it simplifies real language environments. The setting
does not fully cover semantic ambiguity, open-world knowledge, long-document context,
or the linguistic diversity of natural reasoning tasks.

\paragraph{Controlled but narrow task families.}
Deductive, abductive, inductive, and analogy tasks are defined through explicit KG
operations. These definitions make the task families comparable within the same
controlled framework, but they represent only one controlled instantiation of each
reasoning form and do not cover the full range of logical reasoning in the wild.

\paragraph{Finite depth--complexity grid.}
Our evaluation covers \(D\in\{1,\ldots,10\}\) and
\(T\in\{1,\ldots,6\}\). The observed trends may change under longer reasoning
horizons, denser distractor environments, or more complex structures beyond this
grid.

\paragraph{Diagnostic benchmark comparison.}
The Table~\ref{tab:benchmark-leaderboard-calibrated} comparison should be interpreted
as diagnostic rather than as a controlled head-to-head evaluation. Our four internal
models are only 107M parameters, but they receive domain-specific pre-training
and RL post-training on the synthetic KG distribution. In contrast, off-the-shelf
models are evaluated only with task- and cell-matched few-shot prompts that teach the
answer format and basic story setting. Therefore, the table mainly tests whether the
same deductive--abductive asymmetry appears in existing models, not whether our small
models are fairly comparable to frontier systems.

\paragraph{Single training run.}
All checkpoints come from one pre-training seed and one RL run per recipe. The
paired bootstrap intervals in Appendix~\ref{appx:bootstrap} capture
evaluation-sampling uncertainty only and do not bound run-to-run variance; gaps of
a few points should therefore not be read as robust orderings, and multi-seed
replication remains future work.

\paragraph{Scale and backbone coverage.}
The model-scale (51M--1.02B) and backbone ablations in Section~\ref{sec:ablation}
cover the Shallow-Mix recipe and single-sample pass@1 only, one run each. We did not
replicate the full recipe grid, the uniform-versus-curriculum comparison, or the
pass@128-based SG/CG diagnostic at other scales or backbones; the depth--complexity
coupling, the analogy mode collapse, and the curriculum finding are therefore
established at 107M only, and whether they hold at larger scales is untested.
Backbone choice alone shifts per-family levels by up to 7.3 points, comparable to
the recipe gaps we report.

\paragraph{Fixed task mixture and budget.}
All recipes share the 140K/60K/10K/10K task mixture and the 220K budget, so recipe
comparisons are internally controlled, but the conclusions are conditional on
both: the task-mix ablation (Appendix~\ref{appx:ablation-task-mix-subsec}) varies
only the deductive--abductive split, the inductive and analogy shares were not
varied, and overall recipe orderings can shift at larger budgets
(Section~\ref{sec:ablation-data-volume}).

\paragraph{Reward design.}
The outcome-gated reward is held fixed across all recipes and calibrated by the
process-ratio ablation (Appendix~\ref{appx:ablation-process-reward-ratio}), so it
cannot bias recipe comparisons, but it remains coarse: binary full-trace matching
gives no partial credit, the outcome gate discards process information when the
final answer is wrong, and one scheme serves all four families. Ungated, graded,
and per-family reward variants were not tested.

\section*{Acknowledgments}
This work was partially supported by JST SPRING JPMJSP2110 (YZ); JSPS KAKENHI JP22H05106, JP23K28045, JP26H02495, JST CREST JPMJCR21N3, and JST Moonshot JPMJMS263I (HS); and JSPS KAKENHI JP26K25543 (QL). It was also supported by the "Development Acceleration Use" program of ABCI 3.0, provided by AIST and AIST Solutions. We sincerely thank all reviewers for their valuable comments, constructive suggestions, and strong support during the rebuttal stage.

\section*{Ethics Statement}
This study complies with the \href{https://www.aclweb.org/portal/content/acl-code-ethics}{ACL Ethics Policy}.

\bibliography{references}

\clearpage

\appendix
\raggedbottom
\renewcommand{\topfraction}{0.95}
\renewcommand{\dbltopfraction}{0.95}
\renewcommand{\textfraction}{0.05}
\renewcommand{\floatpagefraction}{0.85}
\renewcommand{\dblfloatpagefraction}{0.85}
\section{Appendix}

{\footnotesize
\noindent\textbf{Appendix Contents}\par
\smallskip\hrule\smallskip\noindent
\begin{tabular}{@{}p{0.12\columnwidth}p{0.69\columnwidth}r@{}}
\ref{app:supplementary-material-arr} & Reproducibility and Release & \pageref{app:supplementary-material-arr} \\
\ref{app:data-generation-framework} & Data Generation Framework & \pageref{app:data-generation-framework} \\
\hspace*{1em} \ref{app:summary-of-notation} & Summary of Notation & \pageref{app:summary-of-notation} \\
\hspace*{1em} \ref{app:formal-graph-structure} & Formal Graph Structure & \pageref{app:formal-graph-structure} \\
\hspace*{1em} \ref{app:dynamic-event-operators} & Dynamic Event Semantics & \pageref{app:dynamic-event-operators} \\
\hspace*{1em} \ref{app:static-kinship-relations} & Static Component & \pageref{app:static-kinship-relations} \\
\hspace*{1em} \ref{app:depth-and-environment-complexity} & Reasoning Depth and Environment Complexity & \pageref{app:depth-and-environment-complexity} \\
\hspace*{1em} \ref{app:entity-sampling-and-rendering} & Entity Sampling and Surface Rendering & \pageref{app:entity-sampling-and-rendering} \\
\hspace*{1em} \ref{app:split-construction} & Split Construction & \pageref{app:split-construction} \\
\hspace*{1em} \ref{app:task-specific-supervision} & Task-Specific Supervision & \pageref{app:task-specific-supervision} \\
\hspace*{1em} \ref{app:worked-example} & Worked Example & \pageref{app:worked-example} \\
\ref{app:training-setup} & Training Setup & \pageref{app:training-setup} \\
\hspace*{1em} \ref{app:model-architecture-tokenizer} & Model Architecture and Tokenizer & \pageref{app:model-architecture-tokenizer} \\
\hspace*{1em} \ref{app:pretraining-objective-hyperparameters} & Pre-Training Objective and Hyperparameters & \pageref{app:pretraining-objective-hyperparameters} \\
\hspace*{1em} \ref{app:grpo-post-training-implementation} & GRPO Post-Training Implementation & \pageref{app:grpo-post-training-implementation} \\
\hspace*{1em} \ref{app:parsing-matching} & Generated-Content Parsing and Matching & \pageref{app:parsing-matching} \\
\hspace*{1em} \ref{app:rl-recipes-task-instantiation} & RL Recipes and Task Instantiation & \pageref{app:rl-recipes-task-instantiation} \\
\hspace*{1em} \ref{app:scale-backbone-configs} & Scale and Backbone Sweep Configurations & \pageref{app:scale-backbone-configs} \\
\ref{appx:experiment-evaluation-metrics} & Experiment Evaluation Metrics & \pageref{appx:experiment-evaluation-metrics} \\
\hspace*{1em} -- & Strict pass@k, SG, and CG & \pageref{appx:experiment-evaluation-metrics} \\
\hspace*{1em} -- & Global Mean Rank & \pageref{appx:experiment-evaluation-metrics} \\
\hspace*{1em} -- & Local Cell CG & \pageref{appx:experiment-evaluation-metrics} \\
\hspace*{1em} -- & Task-Pair Pearson Correlation & \pageref{appx:experiment-evaluation-metrics} \\
\ref{appx:bootstrap} & Statistical Significance of the Main Comparisons & \pageref{appx:bootstrap} \\
\ref{appx:ablation-studies} & Additional Ablation Studies & \pageref{appx:ablation-studies} \\
\hspace*{1em} \ref{appx:ablation-task-mix-subsec} & Task Mix & \pageref{appx:ablation-task-mix-subsec} \\
\hspace*{1em} \ref{appx:ablation-process-reward-ratio} & Process-Reward Ratio & \pageref{appx:ablation-process-reward-ratio} \\
\ref{appx:supplementary-experiments-results} & Supplementary Experiments and Results & \pageref{appx:supplementary-experiments-results} \\
\hspace*{1em} \ref{appx:marginal-cg-heatmaps} & Full Marginal CG Heatmaps & \pageref{appx:marginal-cg-heatmaps} \\
\hspace*{1em} \ref{appx:inductive-analogy-experiments} & Task-Family Supplementary Results & \pageref{appx:inductive-analogy-experiments} \\
\hspace*{1em} \ref{appx:passk-curves} & Pass@k Exploration Profile & \pageref{appx:passk-curves} \\
\hspace*{1em} \ref{appx:benchmark-evaluation} & Benchmark Supplementary Results & \pageref{appx:benchmark-evaluation} \\
\hspace*{1em} \ref{appx:best-recipe-per-region} & Best RL Recipe per Region & \pageref{appx:best-recipe-per-region} \\
\ref{appx:curriculum-supplementary-results} & Curriculum Supplementary Experiments & \pageref{appx:curriculum-supplementary-results} \\
\hspace*{1em} \ref{appx:curriculum-order-replay-setting} & Task Setting & \pageref{appx:curriculum-order-replay-setting} \\
\hspace*{1em} \ref{appx:curriculum-da-trajectories} & Observation & \pageref{appx:curriculum-da-trajectories} \\
\hspace*{1em} \ref{appx:curriculum-discussion-section} & Discussion & \pageref{appx:curriculum-discussion-section} \\
\hspace*{1em} \ref{appx:curriculum-full-grid} & Full-Grid and Task-Family Extensions & \pageref{appx:curriculum-full-grid} \\
\end{tabular}
\smallskip\hrule\bigskip
}

\subsection{Reproducibility and Release}
\label{app:supplementary-material-arr}

Unless otherwise stated, we report results from a single training run due to computational constraints (pre-training from scratch and post-training are expensive under our GPU budget); paired bootstrap confidence intervals for the main comparisons are reported in Appendix~\ref{appx:bootstrap}.

Moreover, all data are synthetically generated and do not contain personally identifying information. We also avoid generating offensive content by construction and manually spot-check a sample of the generated corpus.
Finally, all datasets in this work are synthetically generated for research and reproducibility purposes. The released artifacts (code, data, and model checkpoints) are intended for research use only.

Our experiments were facilitated by leveraging \href{https://pytorch.org}{PyTorch}, \href{https://huggingface.co}{Hugging Face}, and \href{https://numpy.org}{NumPy} as essential tools. Furthermore, we used \href{https://chat.openai.com/#}{ChatGPT}, \href{https://openai.com/index/introducing-codex/}{Codex}, and \href{https://docs.anthropic.com/en/docs/claude-code/overview}{Claude Code} in our paper writing and programming.

We release all code, the generated synthetic datasets, and the trained checkpoints under the MIT License at \url{https://github.com/YihuaZhu111/kg-rlvr-allocation}, sufficient to reproduce every table and figure.
We use standard open-source libraries and frameworks, including PyTorch, NumPy, Hugging Face Transformers/Datasets/Tokenizers, LLaMA-Factory, verl, vLLM, pandas, and Matplotlib, and comply with their respective licenses.

\subsection{Data Generation Framework}
\label{app:data-generation-framework}

\paragraph{Scope of This Appendix}
This appendix provides the formal details of the data generation framework introduced
in Section~\ref{sec:method}. The framework is organized around a single symbolic graph,
two reasoning components, and two orthogonal setup axes. The dynamic component supports
object-state reasoning over time, while the static component supports kinship-rule
reasoning over relation paths. The same graph can therefore produce deductive,
abductive, inductive, and analogy tasks with process-verifiable supervision.

\subsubsection{Summary of Notation}
\label{app:summary-of-notation}

\paragraph{Data Generation Notation}
Table~\ref{tab:a1-notation} summarizes the notation introduced in this appendix and used
in the main text.

\begin{table*}[t]
\centering
\footnotesize
\setlength{\tabcolsep}{4pt}
\begin{tabular}{p{0.12\linewidth}p{0.24\linewidth}p{0.15\linewidth}p{0.41\linewidth}}
\hline
Category & Symbol & Type & Meaning \\
\hline
Graph &
\(\mathcal{G}=(\mathcal{P},\mathcal{O},\mathcal{E}_{\mathrm{dyn}},
\mathcal{E}_{\mathrm{stat}},s_0)\) &
tuple & A reasoning instance \\
Graph & \(\mathcal{P}\), \(\mathcal{O}\) & sets & Person and object entities \\
Graph & \(s_0\) & state & Initial world state \\
Dynamic & \(\mathcal{C}_{\mathrm{dyn}}\) & component & Dynamic event-state component \\
Dynamic & \(\mathcal{E}_{\mathrm{dyn}}\) & sequence & Dynamic events on objects \\
Dynamic & \(\mathcal{A}_{\mathrm{init}}\) &
operator set & Initialization declarations: \(\mathrm{own}\) and \(\mathrm{create}\) \\
Dynamic & \(\mathcal{A}_{\mathrm{op}}\) &
operator set & Operation events counted by reasoning depth \\
Dynamic & \(\mathcal{S}\) & state space & Object-state mapping space \\
Dynamic & \(s_t(o)\) & tuple & Owner, possessor, and integrity of object \(o\) \\
Dynamic & \(\delta\) & function & Deterministic state transition \\
Dynamic & \(o^\star\) & object & Target object queried by a task \\
Dynamic & \(\mathrm{Obj}(e_i)\) & set & Objects affected by event \(e_i\) \\
Static & \(\mathcal{C}_{\mathrm{stat}}\) & component & Static kinship component \\
Static & \(\mathcal{E}_{\mathrm{stat}}\) & relation set & Static binary kinship facts \\
Static & \(\mathcal{W}\) & vocabulary & Basic kinship labels \\
Static & \(\mathcal{R}\) & rule set & Composite kinship rules \\
Static & \(r\) & rule & A composite kinship rule \\
Static & \(\gamma(r)\) & path & Basic-relation path realizing \(r\) \\
Static & \(\mathrm{hop}(r)\) & count & Length of the path \(\gamma(r)\) \\
Static & \(\hat{P}\) & person & Matched target person in an analogy task \\
Dynamic & \(T_{\mathrm{final}}\), \(T_q\), \(k_q\) &
indices & Final time, queried time, and target-chain state index \\
Complexity & \(D(\mathcal{G})\) & scalar & Target-chain reasoning depth \\
Complexity & \(T(\mathcal{G})\) & level index & Environment-complexity level \\
Complexity & \(\Theta(T)\) & parameter map & Structural complexity preset \\
Complexity & \(k_{\mathrm{chains}}\) & distribution & Complexity-specific chain-count
distribution \\
Complexity & \(n_{\mathrm{persons}}\) & range & Person-count range for entity sampling \\
Complexity & \(\rho_{\mathrm{crossover}}\) & probability & Cross-chain person-overlap rate \\
Complexity & \(\rho_{\mathrm{exchange}}\) & probability & Exchange-event rate \\
Rendering & \(\Phi\) & renderer & Symbolic graph to natural-language story \\
Supervision & \(x=(Q,\pi,a^\star)\) & tuple & Task instance \\
Supervision & \(Q\), \(\pi\), \(a^\star\) & task fields & Question, gold trace, and gold answer \\
Supervision &
\(\pi_{\mathrm{dyn}}^{\mathrm{ded}}\), \(\pi_{\mathrm{dyn}}^{\mathrm{abd}}\),
\(\pi_{\mathrm{stat}}^{\mathrm{ind}}\), \(\pi_{\mathrm{stat}}^{\mathrm{ana}}\) &
traces & Task-specific gold trace formats \\
Supervision &
\(a^{\star,\mathrm{ind}}\), \(a^{\star,\mathrm{ana}}\) &
answers & Task-specific gold answers for static tasks \\
Supervision & \(\hat{\pi}\), \(\hat{a}\) & model outputs & Predicted trace and predicted answer \\
Supervision & \(\mathrm{m}_{\mathrm{P}}\), \(\mathrm{m}_{\mathrm{A}}\) &
predicates & Process and answer matching criteria \\
Splits & \(n_{\mathrm{cell}}\) & count & Graphs per \((D,T)\) cell \\
\hline
\end{tabular}
\caption{Notation used in the data generation framework.}
\label{tab:a1-notation}
\end{table*}

\subsubsection{Formal Graph Structure}
\label{app:formal-graph-structure}

\paragraph{Generated Graph Instance}
Each instance is a five-tuple
\[
\mathcal{G} =
(\mathcal{P}, \mathcal{O}, \mathcal{E}_{\mathrm{dyn}},
\mathcal{E}_{\mathrm{stat}}, s_0),
\]
where \(\mathcal{P}\) is a finite set of person entities, \(\mathcal{O}\) is a finite
set of object entities, \(\mathcal{E}_{\mathrm{dyn}}\) is a time-ordered sequence of
object events, \(\mathcal{E}_{\mathrm{stat}}\) is a finite set of static binary kinship
relations, and \(s_0\) is the initial world state. The graph carries two components,
defined below.

\paragraph{Dynamic Object States}

\begin{table*}[t]
\centering
\small
\setlength{\tabcolsep}{4pt}
\begin{tabular}{p{0.20\linewidth}p{0.22\linewidth}p{0.50\linewidth}}
\hline
Event class & Included events & State update for object \(o\) \\
\hline
Initialization & own, create &
\(s_0(o)=(p,p,\mathrm{intact})\). These declarations initialize an object state and
are excluded from \(\mathcal{A}_{\mathrm{op}}\). \\
Ownership-possession transfer & gift, sale, exchange &
If \(s_{t-1}(o)=(p,p,g)\), then \(s_t(o)=(p',p',g)\). For exchange, \(p'\) is the new
holder of \(o\). \\
Possession-only transfer & loan, return &
If \(s_{t-1}(o)=(u,v,g)\), then \(s_t(o)=(u,v',g)\). The owner remains \(u\). \\
Integrity update & break, repair &
If \(s_{t-1}(o)=(u,v,g)\), then \(s_t(o)=(u,v,g')\), where
\(g,g' \in \{\mathrm{intact},\mathrm{broken}\}\) and \(g'\neq g\). \\
\hline
\end{tabular}
\caption{State-slot semantics of dynamic event classes.}
\label{tab:operator-effects}
\end{table*}

The dynamic event-state component is
\[
\mathcal{C}_{\mathrm{dyn}} =
(\mathcal{E}_{\mathrm{dyn}}, s_0, \delta,
\mathcal{A}_{\mathrm{init}}, \mathcal{A}_{\mathrm{op}}).
\]
It describes how object-centered states evolve. Object states live in the space
\[
\begin{aligned}
\mathcal{S}
= \{\,s \mid s &: \mathcal{O} \rightarrow
\mathcal{P} \times \mathcal{P} \\
&\quad \times \{\mathrm{intact},\mathrm{broken}\}\,\}.
\end{aligned}
\]
At time \(t\), the projection for object \(o\) is
\[
\begin{aligned}
s_t(o) =
(&\mathrm{owner}(o,t), \mathrm{possessor}(o,t),\\
&\mathrm{integrity}(o,t)).
\end{aligned}
\]
Initial declarations set the initial state and are collected in
\(\mathcal{A}_{\mathrm{init}}=\{\mathrm{own},\mathrm{create}\}\). They are not counted
as operation events for reasoning depth. Subsequent operation events are collected in
\(\mathcal{A}_{\mathrm{op}}\) and update object states through a deterministic
transition function
\[
\delta : \mathcal{S} \times \mathcal{A}_{\mathrm{op}} \rightarrow \mathcal{S},
\]
so that \(s_t = \delta(s_{t-1}, e_t)\) for \(e_t \in \mathcal{A}_{\mathrm{op}}\).
Events whose preconditions are not satisfied are rejected during generation, which keeps
the symbolic trace well defined.

\paragraph{Static Kinship Rules}
The static kinship component is
\[
\mathcal{C}_{\mathrm{stat}} =
(\mathcal{E}_{\mathrm{stat}}, \mathcal{W}, \mathcal{R}).
\]
It describes fixed person-person relations. The relation set is
\[
\mathcal{E}_{\mathrm{stat}} \subseteq \mathcal{P} \times \mathcal{W} \times \mathcal{P},
\]
where \(\mathcal{W}\) is a basic kinship vocabulary. Each fact \((p,w,p')\) is closed
under inverse, so the generator also inserts \((p',w^{-1},p)\) when required. A
composite kinship rule \(r \in \mathcal{R}\) is a named relation realized by a path
\(\gamma(r) \in \mathcal{W}^{\mathrm{hop}(r)}\), with
\(\mathrm{hop}(r) \in \{2,3\}\). The static component is sampled independently of the
dynamic transition function.

\subsubsection{Dynamic Event Semantics}
\label{app:dynamic-event-operators}

\paragraph{Dynamic Event Classes}
The dynamic component groups events by which slots of \(s_t(o)\) they change.
Table~\ref{tab:operator-effects} gives the resulting state-slot semantics. This
presentation abstracts away from surface forms, while preserving the distinction between
ownership transfer, possession-only transfer, and integrity change. Exchange events are
included with ownership-possession transfers because, for each exchanged object, both
the owner and possessor slots move to the new holder.

\paragraph{Interleaved Object Chains}
A graph may contain several dynamic chains, one per object, and these chains are
interleaved into a shared timeline. Entity crossover allows the same person to appear in
multiple chains. Distractor-chain events are rendered in the story even when they do not
affect the target object's answer. Thus the surface story may contain explicit but
irrelevant events, while the gold trace only follows the target object's causal chain.

\paragraph{Dynamic Task Traces}
Let \(e_1,\ldots,e_D\) be the \(D\) target-object operation events on object
\(o^\star\), in causal order. Let \(s_k(o^\star)\) be the target-object state after the
\(k\)-th such event. For a deductive task, the gold dynamic trace is
\[
\pi_{\mathrm{dyn}}^{\mathrm{ded}} =
\left\langle s_0(o^\star), (e_k, s_k(o^\star))_{k=1}^{D} \right\rangle,
\]
and the gold answer is the queried slot of the relevant state. For an abductive task,
one event or initial fact is hidden. Let \(m\) denote the mask mode, \(\tilde e_k\)
denote the visible target-object events, and \(\tilde s_k^{-}\) and \(\tilde s_k^{+}\)
denote the states immediately before and after a visible event. The gold trace is
\[
\begin{aligned}
\pi_{\mathrm{dyn}}^{\mathrm{abd}} =
\left\langle
m,\ s_0^{(m)}(o^\star),\right.&
(\tilde e_k,\tilde s_k^{-}(o^\star),
\tilde s_k^{+}(o^\star))_{k=1}^{\tilde D},\\
&\left.h^\star \right\rangle.
\end{aligned}
\]
Here \(s_0^{(m)}\) is the visible initial state after masking, \(\tilde D\) is the
number of visible target-object events, and \(h^\star\) is the hidden event or initial
fact. The prediction is correct only when both the reconstructed process and the final
answer match the gold structure.

\subsubsection{Static Component}
\label{app:static-kinship-relations}

\paragraph{Basic and Composite Kinship}
The static component uses the basic vocabulary
\[
\begin{aligned}
\mathcal{W}=\{&
\mathrm{father}, \mathrm{mother}, \mathrm{son}, \mathrm{daughter},\\
&\mathrm{husband}, \mathrm{wife}, \mathrm{brother}, \mathrm{sister},\\
&\mathrm{friend}\}.
\end{aligned}
\]
A composite rule \(r\) names a relation realized by a path \(\gamma(r)\) of length two
or three through \(\mathcal{W}\). For example, \(\mathrm{father}(x,y)\) and
\(\mathrm{father}(y,z)\) imply \(\mathrm{grandfather}(x,z)\), so
\(\gamma(\mathrm{grandfather})=(\mathrm{father},\mathrm{father})\).
Table~\ref{tab:composite-rules} lists the 23 composite rules used by the generator.

\begin{table}[t]
\centering
\footnotesize
\setlength{\tabcolsep}{2pt}
\begin{tabular}{p{0.12\linewidth}p{0.36\linewidth}p{0.42\linewidth}}
\hline
Hop & Rule & Path \(\gamma(r)\) \\
\hline
2 & grandmother & (mother, father) \\
2 & grandfather & (father, father) \\
2 & aunt & (sister, father) \\
2 & uncle & (brother, father) \\
2 & mother-in-law & (mother, husband) \\
2 & father-in-law & (father, husband) \\
2 & brother-in-law & (brother, husband) \\
2 & sister-in-law & (sister, husband) \\
2 & grandson & (son, son) \\
2 & granddaughter & (daughter, son) \\
2 & nephew & (son, brother) \\
2 & niece & (daughter, brother) \\
2 & son-in-law & (husband, daughter) \\
2 & daughter-in-law & (wife, son) \\
3 & great-grandfather & (father, father, father) \\
3 & great-grandmother & (mother, father, father) \\
3 & great-grandson & (son, son, son) \\
3 & great-granddaughter & (daughter, son, son) \\
3 & great-uncle & (brother, father, father) \\
3 & great-aunt & (sister, father, father) \\
3 & grand-nephew & (son, son, brother) \\
3 & grand-niece & (daughter, son, brother) \\
3 & cousin & (son, brother, father) \\
\hline
\end{tabular}
\caption{Composite kinship rules and canonical composition paths.}
\label{tab:composite-rules}
\end{table}

\paragraph{Static Task Traces}
For an inductive task, the generator plants support chains that realize a selected rule
\(r\), adds noise kinship facts, and asks the model to recover the composite rule. The
gold static trace records the support sentences, the path \(\gamma(r)\), and the rule
name:
\[
\pi_{\mathrm{stat}}^{\mathrm{ind}} =
\left\langle \{(\sigma_j,\rho_j)\}_{j=1}^{k}, \gamma(r), \hat{r}=r \right\rangle.
\]
\(\sigma_j\) denotes the \(j\)-th support example, and \(\rho_j\) is the corresponding
basic-relation path expressed in \(\mathcal{W}\). The tuple records the support
evidence, the shared composition path \(\gamma(r)\), and the gold composite rule
\(\hat r=r\).
For an analogy task, the generator selects two person pairs that instantiate the same
path. The gold trace records the shared path and the matched target person:
\[
\begin{aligned}
\pi_{\mathrm{stat}}^{\mathrm{ana}} =
\big\langle
&\gamma(r),\ (A \xrightarrow{\gamma(r)} B),\\
&(C \xrightarrow{\gamma(r)} \hat{P}),\ \hat{P}
\big\rangle.
\end{aligned}
\]
The first pair \((A,B)\) provides the relation pattern, and the second pair
\((C,\hat P)\) applies the same path to a new source entity. The final element
\(\hat P\) is the gold target entity.

\subsubsection{Reasoning Depth and Environment Complexity}
\label{app:depth-and-environment-complexity}

\paragraph{Reasoning Depth Definition}
Difficulty is controlled by reasoning depth \(D\) and environment complexity \(T\). Depth
counts only operation events on the target object's causal chain:
\[
D(\mathcal{G}) =
\left|\{e_i \in \mathcal{E}_{\mathrm{dyn}} :
e_i \in \mathcal{A}_{\mathrm{op}},\
o^\star \in \mathrm{Obj}(e_i)\}\right|.
\]
Here \(\mathrm{Obj}(e_i)\) is the set of objects affected by event \(e_i\). For ordinary
single-object events, this set has one element. For exchange events, it contains the two
exchanged objects, so the event contributes to the depth of a target chain only when
\(o^\star\) is one of those objects. Initialization declarations in
\(\mathcal{A}_{\mathrm{init}}\) set \(s_0\) and are not counted in \(D(\mathcal{G})\).
Pre-training uses \(D \in \{1,\ldots,4\}\), while RL and evaluation use
\(D \in \{1,\ldots,10\}\).

\paragraph{Environment Complexity Presets}
The environment complexity \(T(\mathcal{G}) \in \{1,\ldots,6\}\) indexes a
structural parameter map
\[
\begin{aligned}
\Theta(T) &=
(k_{\mathrm{chains}}, n_{\mathrm{persons}}, \rho_{\mathrm{crossover}},
\rho_{\mathrm{exchange}}).
\end{aligned}
\]
\(k_{\mathrm{chains}}\) specifies the complexity-specific distribution over the number of
parallel object chains. \(n_{\mathrm{persons}}\) specifies the person-count range used
when sampling entities. \(\rho_{\mathrm{crossover}}\) controls the probability of
cross-chain person overlap, and \(\rho_{\mathrm{exchange}}\) controls the probability
of exchange events between chains.
Table~\ref{tab:complexity-presets} lists the presets.

\begin{table}[t]
\centering
\scriptsize
\setlength{\tabcolsep}{2pt}
\begin{tabular}{p{0.18\linewidth}p{0.27\linewidth}p{0.13\linewidth}
p{0.14\linewidth}p{0.14\linewidth}}
\hline
Complexity & Chain distribution & Persons & Crossover & Exchange \\
\hline
T1 & \(\{1:1.0\}\) & 5--8 & 0.0 & 0.0 \\
T2 & \(\{1:0.5,2:0.5\}\) & 8--12 & 0.2 & 0.0 \\
T3 & \(\{2:1.0\}\) & 10--15 & 0.3 & 0.2 \\
T4 & \(\{3:1.0\}\) & 12--18 & 0.5 & 0.5 \\
T5 & \(\{4:1.0\}\) & 15--22 & 0.8 & 0.8 \\
T6 & \(\{5:1.0\}\) & 18--25 & 1.0 & 1.0 \\
\hline
\end{tabular}
\caption{Structural complexity presets.}
\label{tab:complexity-presets}
\end{table}

\paragraph{Depth--Complexity Cell Targeting}
Because \(D\) and \(T\) are set independently, the generator can target any cell in the
60-cell depth--complexity grid. This design allows the experiments to separate long-chain
reasoning from reasoning under distractor structure.

\subsubsection{Entity Sampling and Surface Rendering}
\label{app:entity-sampling-and-rendering}

\paragraph{Entity and Object Sampling}
After \(T\) fixes the chain distribution, person-count range, crossover rate, and
exchange rate, person entities are sampled without replacement from synthetic full-name
pools. Object entities are then sampled to instantiate the chain objects; any extra
objects are rendered as distractors. Names are kept distinct across splits to reduce
direct memorization.

\paragraph{Natural-Language Rendering}
A renderer \(\Phi\) maps the symbolic graph and a surface grammar to a natural-language
story:
\[
\Phi : (\mathcal{G}, \mathrm{surface\_grammar}) \rightarrow \mathrm{story}.
\]
The renderer uses a small set of surface variants for each canonical operator.
Table~\ref{tab:surface-variants} illustrates variants for
\(\mathrm{transfer\_gift}(\mathrm{Alice}, \mathrm{Bob}, \mathrm{red\_ball})\).

\begin{table}[t]
\centering
\small
\setlength{\tabcolsep}{3pt}
\begin{tabular}{p{0.30\linewidth}p{0.60\linewidth}}
\hline
Variant & Example surface form \\
\hline
Active & Alice gives the red ball to Bob. \\
Passive & The red ball is given to Bob by Alice. \\
Cleft / role-based & It is Alice who gives the red ball to Bob. \\
Epistemic-marked & Alice decides to give the red ball to Bob. \\
\hline
\end{tabular}
\caption{Surface-form variants used by the renderer.}
\label{tab:surface-variants}
\end{table}

\paragraph{Interleaving Story Streams}
Dynamic and static streams are rendered separately and then interleaved into one story.
This creates surface-level interference between event-state facts and kinship facts
while preserving their symbolic separation. The rendered dynamic events can be divided
into explicit events, which appear in the story, and implicit events or states, which
are used by the simulator or gold trace but are not always stated directly.

\subsubsection{Split Construction}
\label{app:split-construction}

\paragraph{Split Size Construction}
The generator enumerates the requested \((D,T)\) cells and produces a fixed number of
graphs per cell:
\[
\left|\mathrm{Data}_{\mathrm{split}}\right| =
\left|D\right| \cdot \left|T\right| \cdot n_{\mathrm{cell}}.
\]
This prevents the empirical distribution from drifting toward medium-complexity regions.
The pre-training split contains \(D1\)--\(D4\) and \(T1\)--\(T2\), giving 8 cells with
400,000 graphs per cell and 3.2M graphs in total. The shared RL pool contains all
\(D1\)--\(D10\) and \(T1\)--\(T6\) cells, with 20,000 graphs per cell and 1.2M graphs in
total. The evaluation split uses the same 60-cell grid with 80 graphs per cell, for
4,800 graphs.

\paragraph{Sampling Within Each Cell}
Within a cell, the generator samples one target object, a length-\(D\) sequence of
target-chain operations, and additional operations on parallel chains. Cross-chain
exchange events are inserted according to \(\rho_{\mathrm{exchange}}\). The static
component is generated over the same person pool, closed under inverse relations, and
augmented
with noise kinship facts. RL variants are then sampled without replacement from the
shared RL pool, bucket-wise within each \((D,T)\) cell. Thus the RL recipes differ in
data distribution while sharing the same underlying generator.

\subsubsection{Task-Specific Supervision}
\label{app:task-specific-supervision}

\paragraph{Process-Verified Task Instance}
Each task instance is a triple
\begin{equation}
x = (Q, \pi, a^\star),
\end{equation}
where \(Q\) is the question, \(\pi\) is the gold solution trace, and \(a^\star\) is the
gold answer. Under process-verified evaluation, a model prediction
\((\hat{\pi},\hat{a})\) is correct only if both the process and the answer match the
gold structure:
\[
\mathrm{m}_{\mathrm{P}}(\hat{\pi},\pi)
\wedge
\mathrm{m}_{\mathrm{A}}(\hat{a},a^\star).
\]
The framework supplies two distinct trace formats, chosen by the task family.

\paragraph{Deductive and Abductive Supervision}
Deductive and abductive tasks use the dynamic gold solution traces defined in
Section~\ref{app:dynamic-event-operators}. For deductive tasks, the generator
instantiates up to three \texttt{deduction\_full\_info} queries per graph, one for each
state slot \texttt{owner}, \texttt{possessor}, or \texttt{integrity}, queried at the
final time \(T_{\mathrm{final}}\). It also instantiates up to three
\texttt{deduction\_hard} queries, where the same three slots are queried at an
intermediate time \(T_q<T_{\mathrm{final}}\). The trace is
\(\pi=\pi_{\mathrm{dyn}}^{\mathrm{ded}}\), and the gold answer is the projection of
\(s_{k_q}(o^\star)\) onto the queried slot. For abductive tasks, the generator chooses
\texttt{mask\_one\_event} with probability \(0.75\) and
\texttt{mask\_one\_initial\_fact} with probability \(0.25\). The chosen element is
removed from the rendered story and recorded as the gold answer. The trace is
\(\pi=\pi_{\mathrm{dyn}}^{\mathrm{abd}}\), and the gold answer is the natural-language
realization of the masked element. If a target object would end in the broken state in a
way that makes the masked inference ambiguous, an automatic repair event is appended to
preserve solvability.

\paragraph{Inductive and Analogy Supervision}
Inductive and analogy tasks use the static gold solution traces defined in
Section~\ref{app:static-kinship-relations}. They sample the planted composite rule \(r\)
from a six-group frequency stratification of the 23 composite rules in \(\mathcal{R}\).
Pre-training uses the long-tailed distribution in
Table~\ref{tab:kinship-frequency-groups}; evaluation uses uniform coverage over the
same groups; and RL uses uniform sampling over the non-holdout groups. Inductive tasks
draw \(r\) from groups A--E, giving 19 rules in total. Analogy tasks draw \(r\) from
groups A--C only, giving 12 rules. Group F is excluded from pre-training and RL to
preserve a clean evaluation axis for unseen rule compositions.

\begin{table*}[t]
\centering
\small
\setlength{\tabcolsep}{4pt}
\begin{tabular}{p{0.08\linewidth}p{0.10\linewidth}p{0.76\linewidth}}
\hline
Group & Weight & Relations \\
\hline
A & 0.50 & grandmother, grandfather, great-grandfather, great-grandmother \\
B & 0.30 & aunt, uncle, cousin, grand-nephew \\
C & 0.15 & mother-in-law, brother-in-law, grand-niece, great-uncle \\
D & 0.04 & grandson, granddaughter, great-grandson, great-granddaughter \\
E & 0.01 & nephew, niece, great-aunt \\
F & 0.00 & son-in-law, daughter-in-law, sister-in-law, father-in-law \\
\hline
\end{tabular}
\caption{Kinship rule frequency groups in the inductive dataset.}
\label{tab:kinship-frequency-groups}
\end{table*}

\paragraph{Static Query Construction}
For inductive tasks, the generator builds two or three support sentences, each realizing
the planted rule \(r\) on a distinct person triple drawn from
\(\mathcal{E}_{\mathrm{stat}}\). It then adds one to three noise kinship facts and
fabricates a fresh trio of synthetic persons for the query path. The trace is
\(\pi=\pi_{\mathrm{stat}}^{\mathrm{ind}}\), and the gold answer is
\(a^{\star,\mathrm{ind}}=r\). For analogy tasks, the generator selects two paths in
\(\mathcal{E}_{\mathrm{stat}}\) that share the same composition \(\gamma(r)\) and
rephrases the question as ``\(A\) is to \(B\) as \(C\) is to \(?\)''. The trace is
\(\pi=\pi_{\mathrm{stat}}^{\mathrm{ana}}\), and the gold answer is the matched person
\(a^{\star,\mathrm{ana}}=\hat{P}\).

\paragraph{Question-Form Inventory}
The four reasoning families differ not only in skill but also in the number of
distinct question forms produced by the generator. Deductive supervision yields
six forms, corresponding to the Cartesian product of three state slots
(\texttt{owner}, \texttt{possessor}, \texttt{integrity}) and two temporal modes
(\texttt{full\_info} at the final time \(T_{\mathrm{final}}\) versus
\texttt{hard} at an intermediate time \(T_q<T_{\mathrm{final}}\)). Abductive
supervision yields two forms that share the same surface template but differ in
the masking target: \texttt{mask\_one\_event} hides a single operation, while
\texttt{mask\_one\_initial\_fact} hides a single initial-state fact. Inductive
and analogy supervision each yield one question form, drawn at sampling time
from four interchangeable phrasing variants to reduce surface-level
memorization. Table~\ref{tab:question-forms} summarizes the inventory.

\begin{table*}[t]
\centering
\small
\setlength{\tabcolsep}{4pt}
\renewcommand{\arraystretch}{1.15}
\begin{tabular}{p{0.08\linewidth}p{0.03\linewidth}p{0.20\linewidth}p{0.60\linewidth}}
\hline
Task & \# & Form name & Canonical question template \\
\hline
\multirow{6}{*}{Deductive}
  & 1 & Final possessor    & ``After all events, who possesses \{the object\}?'' \\
  & 2 & Final owner        & ``After all events, who owns \{the object\}?'' \\
  & 3 & Final integrity    & ``Is \{the object\} broken or intact after all events?'' \\
  & 4 & Kth-step possessor & ``After the \(k\)th operation involving \{the object\}, who possesses \{the object\}?'' \\
  & 5 & Kth-step owner     & ``After the \(k\)th operation involving \{the object\}, who owns \{the object\}?'' \\
  & 6 & Kth-step integrity & ``After the \(k\)th operation involving \{the object\}, is \{the object\} broken or intact?'' \\
\hline
\multirow{2}{*}{Abductive}
  & 1 & Mask one event        & ``Some information is missing from the story. Please infer the missing event or information involving \{the object\}.'' (a single event in the temporal sequence is hidden) \\
  & 2 & Mask one initial fact & Same surface template, but a single initial-state fact is hidden instead \\
\hline
Inductive & 1 & Rule composition & ``What is \{subj\} to \{obj\}?'' --- compose the planted kinship rule from support chains. Four interchangeable paraphrases. \\
\hline
Analogy & 1 & A:B::C:?         & ``\{A\} is to \{B\} as \{C\} is to ?'' --- complete the analogy with the missing entity. Four interchangeable paraphrases. \\
\hline
\end{tabular}
\caption{Question-form inventory across the four reasoning task families.
Deductive yields six forms from three query slots (\texttt{possess},
\texttt{own}, \texttt{integrity}) crossed with two temporal modes
(\texttt{full\_info} at \(T_{\mathrm{final}}\) versus \texttt{hard} at
\(T_q<T_{\mathrm{final}}\)). Abductive yields two forms that share a single
surface template but differ in the masking target. Inductive and analogy each
use one question form realized through four interchangeable phrasing variants.}
\label{tab:question-forms}
\end{table*}

\paragraph{Missing Task Families}
For a given graph, some task families may be empty if the structural sampler does not
produce the required prerequisites; for example, the static component may be sampled
with too few persons to support a two-hop rule. The worked example in
Section~\ref{app:worked-example} shows one such graph, in which deductive and inductive
tasks are populated by the generator while abductive and analogy tasks are constructed
manually for pedagogy.

\subsubsection{Worked Example}
\label{app:worked-example}

\paragraph{Concrete Example Instance}
Table~\ref{tab:worked-example} shows one concrete graph instance rendered as a full
story and instantiated as four reasoning tasks. The example is chosen because its target
object timeline contains diverse dynamic operations rather than only break/repair events.
For pedagogy, the abductive and analogy rows are constructed from the same symbolic graph
by masking one event and reusing the planted static relation path.

\begin{table*}[t]
\centering
\captionsetup{skip=0pt,font=scriptsize}
\tiny
\setlength{\tabcolsep}{2pt}
\renewcommand{\arraystretch}{0.86}
\begin{tabular}{>{\raggedright\arraybackslash}p{0.16\textwidth}
>{\raggedright\arraybackslash}p{0.78\textwidth}}
\hline
Field & Content \\
\hline
Structural facts &
Target-object reasoning depth \(D=10\), generated from two chains with per-chain depth
7, complexity level \(T=3\), entity crossover enabled, and exchange enabled. The target
object is the ivory chess piece; the second object is the ebony cane. The planted static
rule is \texttt{great-aunt}, with path
\((\texttt{sister},\texttt{father},\texttt{mother})\). \\
Target timeline &
The target-object timeline is
\(\texttt{sale}\rightarrow\texttt{loan}\rightarrow\texttt{return}\rightarrow
\texttt{break}\rightarrow\texttt{exchange}\rightarrow\texttt{repair}\rightarrow
\texttt{break}\rightarrow\texttt{gift}\rightarrow\texttt{repair}\rightarrow
\texttt{sale}\). \\
Rendered story &
Karen Baylor Sawyer currently own the ivory chess piece. The ebony cane is owned by
Esther Delvin Murdoch. Agnes Drummond Fielding is the sister of Charity Lark Amberley.
Karen Baylor Sawyer sell Charity Lark Amberley the ivory chess piece. Charity Lark
Amberley holds the relation of sister to Agnes Drummond Fielding. The ivory chess piece
is lent to Dustin Keith Ravenscroft by Charity Lark Amberley. Charity Lark Amberley acts
in the role of father to Jacqueline Coby Pettigrew. The ivory chess piece is returned to
Charity Lark Amberley by Dustin Keith Ravenscroft. Jacqueline Coby Pettigrew holds the
relation of son to Charity Lark Amberley. It is Esther Delvin Murdoch who break the
ebony cane. Jacqueline Coby Pettigrew holds the relation of mother to Andrea Bryson
McCaffrey. Charity Lark Amberley break the ivory chess piece. Andrea Bryson McCaffrey
serves as Jacqueline Coby Pettigrew's daughter. Charity Lark Amberley exchange the ivory
chess piece with Esther Delvin Murdoch for the ebony cane. Edward Delvin Clements acts
in the role of sister to Cyril Amias Devonshire. Charity Lark Amberley repair the ebony
cane. Cyril Amias Devonshire is the sister of Edward Delvin Clements. Charity Lark
Amberley accidentally break the ebony cane. Cyril Amias Devonshire serves as Bella
Aether Ainsworth's father. Esther Delvin Murdoch decide to repair the ivory chess piece.
Bella Aether Ainsworth is the son of Cyril Amias Devonshire. It is Esther Delvin Murdoch
who break the ivory chess piece. Bella Aether Ainsworth acts in the role of mother to
Andrew Essex Northcott. Esther Delvin Murdoch give the ivory chess piece to Karen Baylor
Sawyer. Andrew Essex Northcott holds the relation of daughter to Bella Aether Ainsworth.
Karen Baylor Sawyer repair the ivory chess piece. Denise Hanley Glendale is the son of
Dustin Keith Ravenscroft. Karen Baylor Sawyer decide to sell the ivory chess piece to
Eric Aston Fairbanks. Dustin Keith Ravenscroft holds the relation of father to Denise
Hanley Glendale. Andrea Bryson McCaffrey holds the relation of brother to Denise Hanley
Glendale. Denise Hanley Glendale serves as Andrea Bryson McCaffrey's brother. Edward
Delvin Clements acts in the role of daughter to Andrea Bryson McCaffrey. Andrea Bryson
McCaffrey is the mother of Edward Delvin Clements. \\
\end{tabular}

\vspace{0pt}
\begin{tabular}{>{\raggedright\arraybackslash}p{0.06\textwidth}
>{\raggedright\arraybackslash}p{0.10\textwidth}
>{\raggedright\arraybackslash}p{0.71\textwidth}
>{\raggedright\arraybackslash}p{0.07\textwidth}}
\specialrule{0.8pt}{0pt}{1pt}
\multicolumn{4}{l}{\textbf{Dynamic Reasoning Tasks}} \\
\hline
Task & Question & Solution & Answer \\
\hline
Deductive &
After all events, who possess the ivory chess piece? &
\begin{minipage}[t]{\linewidth}
Initial state: Karen Baylor Sawyer own and possess the ivory chess piece (intact).

Step 1: Karen Baylor Sawyer sells the ivory chess piece to Charity Lark Amberley.
State: Charity Lark Amberley own and possess the ivory chess piece (intact).

Step 2: Charity Lark Amberley lends the ivory chess piece to Dustin Keith Ravenscroft.
State: Charity Lark Amberley own the ivory chess piece, Dustin Keith Ravenscroft
possess the ivory chess piece (intact).

Step 3: Dustin Keith Ravenscroft returns the ivory chess piece to Charity Lark
Amberley. State: Charity Lark Amberley own and possess the ivory chess piece (intact).

Step 4: Charity Lark Amberley breaks the ivory chess piece. State: Charity Lark
Amberley own and possess the ivory chess piece (broken).

Step 5: Charity Lark Amberley exchanges the ivory chess piece with Esther Delvin
Murdoch for the ebony cane. State: Esther Delvin Murdoch own and possess the ivory
chess piece (broken).

Step 6: Esther Delvin Murdoch repairs the ivory chess piece. State: Esther Delvin
Murdoch own and possess the ivory chess piece (intact).

Step 7: Esther Delvin Murdoch breaks the ivory chess piece. State: Esther Delvin
Murdoch own and possess the ivory chess piece (broken).

Step 8: Esther Delvin Murdoch gives the ivory chess piece to Karen Baylor Sawyer.
State: Karen Baylor Sawyer own and possess the ivory chess piece (broken).

Step 9: Karen Baylor Sawyer repairs the ivory chess piece. State: Karen Baylor Sawyer
own and possess the ivory chess piece (intact).

Step 10: Karen Baylor Sawyer sells the ivory chess piece to Eric Aston Fairbanks.
State: Eric Aston Fairbanks own and possess the ivory chess piece (intact).

Therefore, the answer is Eric Aston Fairbanks.
\end{minipage} &
Eric Aston Fairbanks \\
\hline
Abductive &
Some information is missing from the story. Please infer the missing event or
information involving the ivory chess piece. &
\begin{minipage}[t]{\linewidth}
Tracing backwards.

Last step: Karen Baylor Sawyer sells the ivory chess piece to Eric Aston Fairbanks.
Before: Karen Baylor Sawyer own and possess the ivory chess piece (intact). After:
Eric Aston Fairbanks own and possess the ivory chess piece (intact).

Previous step: Karen Baylor Sawyer repairs the ivory chess piece. Before: Karen Baylor
Sawyer own and possess the ivory chess piece (broken). After: Karen Baylor Sawyer own
and possess the ivory chess piece (intact). After matches before of the next step, no
gap.

Previous step: Esther Delvin Murdoch gives the ivory chess piece to Karen Baylor
Sawyer. Before: Esther Delvin Murdoch own and possess the ivory chess piece (broken).
After: Karen Baylor Sawyer own and possess the ivory chess piece (broken). After
matches before of the next step, no gap.

Previous step: Esther Delvin Murdoch breaks the ivory chess piece. Before: Esther
Delvin Murdoch own and possess the ivory chess piece (intact). After: Esther Delvin
Murdoch own and possess the ivory chess piece (broken). After matches before of the
next step, no gap.

Previous step: Esther Delvin Murdoch repairs the ivory chess piece. Before: Esther
Delvin Murdoch own and possess the ivory chess piece (broken). After: Esther Delvin
Murdoch own and possess the ivory chess piece (intact). After matches before of the
next step, no gap.

Previous step: Charity Lark Amberley breaks the ivory chess piece. Before: Charity
Lark Amberley own and possess the ivory chess piece (intact). After: Charity Lark
Amberley own and possess the ivory chess piece (broken). After does not match before
of the next step, gap found.

Before the missing step: Charity Lark Amberley own and possess the ivory chess piece
(broken). After the missing step: Esther Delvin Murdoch own and possess the ivory chess
piece (broken).

Therefore, the missing information is: Charity Lark Amberley exchanges the ivory chess
piece with Esther Delvin Murdoch for the ebony cane.
\end{minipage} &
Charity Lark Amberley exchanges the ivory chess piece with Esther Delvin Murdoch for the
ebony cane. \\
\end{tabular}

\vspace{0pt}
\begin{tabular}{>{\raggedright\arraybackslash}p{0.06\textwidth}
>{\raggedright\arraybackslash}p{0.28\textwidth}
>{\raggedright\arraybackslash}p{0.50\textwidth}
>{\raggedright\arraybackslash}p{0.08\textwidth}}
\specialrule{0.8pt}{0pt}{1pt}
\multicolumn{4}{l}{\textbf{Static Reasoning Tasks}} \\
\hline
Task & Question & Solution & Answer \\
\hline
Inductive &
Andrea Bryson McCaffrey serves as Edward Delvin Clements's mother. Agnes Drummond
Fielding holds the great-aunt relation to Andrea Bryson McCaffrey. Dustin Keith
Ravenscroft holds the father relation to Denise Hanley Glendale. Edward Delvin Clements
acts as the great-aunt of Andrew Essex Northcott. Andrea Bryson McCaffrey holds the
daughter relation to Jacqueline Coby Pettigrew. Bonnie Ansel Barncastle acts in the
role of sister to Pauline Beckett Brookhaven, and Pauline Beckett Brookhaven serves as
Phyllis Euan Ames's father, and Phyllis Euan Ames holds the relation of mother to
Antonio Beck Harmon. What is the relationship between Bonnie Ansel Barncastle and
Antonio Beck Harmon? &
\begin{minipage}[t]{\linewidth}
Example: Agnes Drummond Fielding is the sister of Charity Lark Amberley, and Charity
Lark Amberley is the father of Jacqueline Coby Pettigrew, and Jacqueline Coby Pettigrew
is the mother of Andrea Bryson McCaffrey, so Agnes Drummond Fielding is the great-aunt
of Andrea Bryson McCaffrey.

Example: Edward Delvin Clements is the sister of Cyril Amias Devonshire, and Cyril
Amias Devonshire is the father of Bella Aether Ainsworth, and Bella Aether Ainsworth is
the mother of Andrew Essex Northcott, so Edward Delvin Clements is the great-aunt of
Andrew Essex Northcott.

The pattern is sister then father then mother gives great-aunt.

Applying to query: Bonnie Ansel Barncastle is the sister of Pauline Beckett Brookhaven,
and Pauline Beckett Brookhaven is the father of Phyllis Euan Ames, and Phyllis Euan
Ames is the mother of Antonio Beck Harmon. By the same pattern, Bonnie Ansel Barncastle
is the great-aunt of Antonio Beck Harmon.

Therefore, the answer is great-aunt.
\end{minipage} &
great-aunt \\
\hline
Analogy &
Agnes Drummond Fielding is to Andrea Bryson McCaffrey as Edward Delvin Clements is to
\(?\). &
\begin{minipage}[t]{\linewidth}
From the story, Agnes Drummond Fielding is the sister of Charity Lark Amberley, and
Charity Lark Amberley is the father of Jacqueline Coby Pettigrew, and Jacqueline Coby
Pettigrew is the mother of Andrea Bryson McCaffrey. Through sister then father then
mother, Agnes Drummond Fielding is the great-aunt of Andrea Bryson McCaffrey.

Similarly, Edward Delvin Clements is the sister of Cyril Amias Devonshire, and Cyril
Amias Devonshire is the father of Bella Aether Ainsworth, and Bella Aether Ainsworth is
the mother of Andrew Essex Northcott. By the same pattern, Edward Delvin Clements is the
great-aunt of Andrew Essex Northcott.

Therefore, the answer is Andrew Essex Northcott.
\end{minipage} &
Andrew Essex Northcott \\
\hline
\end{tabular}
\caption{Worked example. Deductive and inductive are generated task instances.
Abductive is constructed by masking this operation: Charity Lark Amberley exchanges the
ivory chess piece with Esther Delvin Murdoch for the ebony cane. Analogy reuses the
planted \texttt{great-aunt} path.}
\label{tab:worked-example}
\end{table*}

\subsection{Training Setup}
\label{app:training-setup}

This appendix gives the implementation details for the training setup described in
Section~\ref{sec:train-setup}. The reward definition is given in the main text; here we
summarize the model, pre-training configuration, GRPO implementation, parsing and
matching protocol, and recipe coverage.
All training runs were conducted on a single node with 8 NVIDIA H200 GPUs. One
pre-training epoch took approximately 1.5 hours, and one GRPO post-training epoch
took approximately 5 hours.

\subsubsection{Model Architecture and Tokenizer}
\label{app:model-architecture-tokenizer}

All experiments use the same 107M-parameter Qwen2.5 decoder-only transformer. The
model is initialized from scratch with the domain tokenizer and then trained on the KG
pre-training corpus. Table~\ref{tab:model-architecture} lists the architectural
settings used by the model.

\begin{table*}[t]
\centering
\small
\setlength{\tabcolsep}{5pt}
\begin{tabular}{p{0.22\linewidth}p{0.20\linewidth}p{0.50\linewidth}}
\hline
Component & Value & Description \\
\hline
Architecture & Qwen2.5 decoder & Causal decoder-only language model \\
Parameters & 107M & Model scale used in all experiments \\
Hidden size & 768 & Transformer hidden dimension \\
Layers & 12 & Number of decoder blocks \\
Attention heads & 12 & Query attention heads \\
KV heads & 2 & Grouped-query key/value heads \\
FFN size & 3072 & Intermediate dimension in the feed-forward block \\
Vocabulary size & 4096 & KG-domain tokenizer vocabulary \\
Context length & 2048 & Maximum model context window \\
Weight tying & enabled & Input and output embeddings are tied \\
Precision & bfloat16 & Training and rollout precision where supported \\
\hline
\end{tabular}
\caption{Model architecture and tokenizer settings.}
\label{tab:model-architecture}
\end{table*}

\subsubsection{Pre-Training Objective and Hyperparameters}
\label{app:pretraining-objective-hyperparameters}

Pre-training uses full-parameter causal language modeling from scratch rather than
instruction tuning. The corpus covers the pre-training block
\(D\in\{1,\ldots,4\}\) and \(T\in\{1,2\}\). The final pre-trained checkpoint is used as
the shared initialization for all RL recipes. Table~\ref{tab:pretraining-hyperparameters}
lists the pre-training hyperparameters.

\begin{table*}[t]
\centering
\small
\setlength{\tabcolsep}{5pt}
\begin{tabular}{p{0.24\linewidth}p{0.22\linewidth}p{0.46\linewidth}}
\hline
Setting & Value & Notes \\
\hline
Objective & causal LM & Full-parameter pre-training from scratch \\
Epochs & 3 & Final checkpoint used for RL initialization \\
Learning rate & \(2\times 10^{-4}\) & Peak learning rate \\
Scheduler & cosine with min LR & Learning-rate schedule \\
Warmup ratio & 0.05 & Fraction of steps used for LR warmup \\
Weight decay & 0.1 & AdamW regularization \\
Per-device batch & 16 & Training batch size per GPU \\
Gradient accumulation & 2 & Effective global batch uses 8 GPUs \\
Optimizer & AdamW & Betas \((0.9,0.999)\), epsilon \(10^{-8}\) \\
Seed & 42 & Pre-training seed \\
\hline
\end{tabular}
\caption{Pre-training hyperparameters.}
\label{tab:pretraining-hyperparameters}
\end{table*}

\subsubsection{GRPO Post-Training Implementation}
\label{app:grpo-post-training-implementation}

RL post-training starts from the final pre-trained checkpoint and uses GRPO with a
rule-based reward computed from parsed answers and traces. We do not train a separate
learned reward model. Table~\ref{tab:grpo-hyperparameters} gives the main GRPO
settings.

\begin{table*}[t]
\centering
\small
\setlength{\tabcolsep}{5pt}
\begin{tabular}{p{0.24\linewidth}p{0.22\linewidth}p{0.46\linewidth}}
\hline
Setting & Value & Notes \\
\hline
Implementation & distributed GRPO & Post-training implementation \\
Initial checkpoint & final pre-trained checkpoint & Shared start point for all recipes \\
Rollouts per prompt & 6 & Group size for GRPO advantage estimation \\
Train batch size & 1024 & Prompt batch size \\
Prompt length & 1024 & Maximum prompt tokens \\
Response length & 1024 & Maximum generated tokens \\
Actor learning rate & \(10^{-6}\) & Actor optimizer LR \\
KL loss coefficient & 0.001 & Fixed KL regularization coefficient \\
GPUs & 8 & Single-node multi-GPU training \\
Reward model & none & Rule-based verifier only \\
\hline
\end{tabular}
\caption{GRPO hyperparameters.}
\label{tab:grpo-hyperparameters}
\end{table*}

\subsubsection{Generated-Content Parsing and Matching}
\label{app:parsing-matching}

The verifier converts each generated response into two objects: a predicted answer
\(\hat{a}\) and a predicted reasoning trace \(\hat{\pi}\). It first extracts the
solution and answer fields for the target task family. If the solution field is missing
or cannot be parsed into the expected trace structure, the process indicator is set to
\(\mathrm{m}_{\mathrm{P}}=0\). The answer indicator
\(\mathrm{m}_{\mathrm{A}}\) is computed by matching the extracted answer against the
gold answer after light normalization; for abductive tasks, equivalent descriptions of
the same missing event or initial fact are also accepted.

Process matching is task-specific. Dynamic tasks are parsed into object-state traces,
where each comparable state records owner, possessor, and integrity. Static tasks are
parsed into relation paths and composite-rule applications.
Table~\ref{tab:parsing-matching-protocol} summarizes the protocol used to compute the
binary process and answer indicators.

\begin{table*}[t]
\centering
\small
\setlength{\tabcolsep}{4pt}
\begin{tabular}{p{0.12\linewidth}p{0.28\linewidth}p{0.34\linewidth}p{0.18\linewidth}}
\hline
Task family & Parsed process & Process match \(\mathrm{m}_{\mathrm{P}}\) &
Answer match \(\mathrm{m}_{\mathrm{A}}\) \\
\hline
Deductive &
Initial object state and ordered post-operation states for the target object &
\(\mathrm{m}_{\mathrm{P}}=1\) only if the full owner, possessor, and integrity
state sequence matches the gold trace with no missing or extra step &
Normalized predicted answer equals the gold state value \\
Abductive &
Backward visible steps, each step's before/after states, and the required initial
state or missing-step gap &
\(\mathrm{m}_{\mathrm{P}}=1\) only if all visible states and the missing-condition
states match the gold backward trace &
Normalized predicted answer matches any valid description of the same missing
event or initial fact \\
Inductive &
Support-example relation path, inferred composite rule, and query application &
\(\mathrm{m}_{\mathrm{P}}=1\) only if the basic-relation path, inferred rule, and
query-side rule application all match the gold structure &
Normalized predicted rule name equals the gold composite rule \\
Analogy &
Source-pair relation pattern and target-pair mapped answer &
\(\mathrm{m}_{\mathrm{P}}=1\) only if the source path/rule and the target mapping
both match the gold relation pattern &
Normalized predicted entity equals the gold mapped entity \\
\hline
\end{tabular}
\caption{Parsing and matching protocol for generated solutions.}
\label{tab:parsing-matching-protocol}
\end{table*}

Tables~\ref{tab:dynamic-parsing-examples} and~\ref{tab:static-parsing-examples} separate the generated text from the structured view produced by
the same parser schema used during evaluation.

\begin{table*}[t]
\centering
\scriptsize
\setlength{\tabcolsep}{2pt}
\begin{tabular}{
>{\raggedright\arraybackslash}p{0.08\linewidth}
>{\raggedright\arraybackslash}p{0.49\linewidth}
>{\raggedright\arraybackslash}p{0.37\linewidth}}
\hline
Task & Generated solution & Parsed content from parser schema \\
\hline
Deductive &
Initial state: Arthur Brolin Bellingham own and possess the copper coin (intact).
Step 1: Arthur Brolin Bellingham gives the copper coin to Mason Kent Glendale.
State: Mason Kent Glendale own and possess the copper coin (intact). Step 2:
Mason Kent Glendale lends the copper coin to Helen Rufus Oaksworth. State: Mason
Kent Glendale own the copper coin, Helen Rufus Oaksworth possess the copper coin
(intact). Step 3: Helen Rufus Oaksworth returns the copper coin to Mason Kent
Glendale. State: Mason Kent Glendale own and possess the copper coin (intact).
Step 4: Mason Kent Glendale sells the copper coin to Amber Blair Houghton. State:
Amber Blair Houghton own and possess the copper coin (intact). Therefore, the
answer is Amber Blair Houghton. &
\texttt{object\_name}: the copper coin.
\texttt{initial\_state}: (Arthur Brolin Bellingham, Arthur Brolin Bellingham,
intact).
\texttt{operations}: 1. gives; before (Arthur, Arthur, intact), after (Mason,
Mason, intact). 2. lends; before (Mason, Mason, intact), after (Mason, Helen,
intact). 3. returns; before (Mason, Helen, intact), after (Mason, Mason, intact).
4. sells; before (Mason, Mason, intact), after (Amber, Amber, intact).
\texttt{answer}: Amber Blair Houghton. \\
Abductive &
Tracing backwards. Last step: Connor Arbor Wolcott repairs the leather bag.
Before: Ariana Reef Nelson own the leather bag, Connor Arbor Wolcott possess the
leather bag (broken). After: Ariana Reef Nelson own the leather bag, Connor Arbor
Wolcott possess the leather bag (intact). Previous step: Ariana Reef Nelson lends
the leather bag to Connor Arbor Wolcott. Before: Ariana Reef Nelson own and
possess the leather bag (broken). After: Ariana Reef Nelson own the leather bag,
Connor Arbor Wolcott possess the leather bag (broken). After matches before of
the next step, no gap. Previous step: Connor Arbor Wolcott returns the leather
bag to Ariana Reef Nelson. Before: Ariana Reef Nelson own the leather bag, Connor
Arbor Wolcott possess the leather bag (broken). After: Ariana Reef Nelson own and
possess the leather bag (broken). After matches before of the next step, no gap.
Previous step: Ariana Reef Nelson lends the leather bag to Connor Arbor Wolcott.
Before: Ariana Reef Nelson own and possess the leather bag (broken). After:
Ariana Reef Nelson own the leather bag, Connor Arbor Wolcott possess the leather
bag (broken). After matches before of the next step, no gap. Previous step:
Ariana Reef Nelson breaks the leather bag. Before: Ariana Reef Nelson own and
possess the leather bag (intact). After: Ariana Reef Nelson own and possess the
leather bag (broken). After matches before of the next step, no gap. Required
initial state: Ariana Reef Nelson own and possess the leather bag (intact). The
story does not state this. Therefore, the missing information is: Ariana Reef
Nelson own the leather bag. &
\texttt{object\_name}: the leather bag.
\texttt{operations} are parsed in backward order: 1. repairs; before (Ariana,
Connor, broken), after (Ariana, Connor, intact). 2. lends; before (Ariana,
Ariana, broken), after (Ariana, Connor, broken). 3. returns; before (Ariana,
Connor, broken), after (Ariana, Ariana, broken). 4. lends; before (Ariana,
Ariana, broken), after (Ariana, Connor, broken). 5. breaks; before (Ariana,
Ariana, intact), after (Ariana, Ariana, broken).
\texttt{initial\_state}: (Ariana Reef Nelson, Ariana Reef Nelson, intact).
\texttt{missing\_step\_before}: null.
\texttt{missing\_step\_after}: null.
\texttt{answer}: Ariana Reef Nelson own the leather bag. \\
\hline
\end{tabular}
\caption{Concrete parsing examples for dynamic reasoning tasks. Generated
solutions are copied verbatim from model outputs; parsed content is
reconstructed with the same parser schema used during evaluation. Dataset identifiers
are omitted. State tuples are
\((owner, possessor, integrity)\), and both examples contain at least four
target-object operations.}
\label{tab:dynamic-parsing-examples}
\end{table*}

\begin{table*}[t]
\centering
\scriptsize
\setlength{\tabcolsep}{2pt}
\begin{tabular}{
>{\raggedright\arraybackslash}p{0.08\linewidth}
>{\raggedright\arraybackslash}p{0.55\linewidth}
>{\raggedright\arraybackslash}p{0.31\linewidth}}
\hline
Task & Generated solution & Parsed content from parser schema \\
\hline
Inductive &
Example: Hope Baden Paddington is the sister of Asher Fischer Bellamy, and Asher
Fischer Bellamy is the mother of Anita Cole Clarkson, so Hope Baden Paddington is
the aunt of Anita Cole Clarkson. Example: Oscar Rayner Frazier is the sister of
Lena Ciel Stanford, and Lena Ciel Stanford is the mother of Liam Carlyle
Merriweather, so Oscar Rayner Frazier is the aunt of Liam Carlyle Merriweather.
The pattern is sister then mother gives aunt. Applying to query: Jean Digby
Murdoch is the sister of Annika Ridge Overton, and Annika Ridge Overton is the
mother of Oliver Ivar Grayson. By the same pattern, Jean Digby Murdoch is the aunt
of Oliver Ivar Grayson. Therefore, the answer is aunt. &
\texttt{pattern\_path}: [sister, mother].
\texttt{pattern\_rule}: aunt.
\texttt{query\_rule}: aunt.
\texttt{answer}: aunt. \\
Analogy &
From the story, Kimberly Chandos Fitzgerald is the brother of Anthony Bryden
Emerson, and Anthony Bryden Emerson is the mother of Jared Remington Berwick.
Through brother then mother, Kimberly Chandos Fitzgerald is the uncle of Jared
Remington Berwick. Similarly, Emma Adair Cromwell is the brother of Dorothy
Lester Blanchard, and Dorothy Lester Blanchard is the mother of Julia Thane
Burleigh. By the same pattern, Emma Adair Cromwell is the uncle of Julia Thane
Burleigh. Therefore, the answer is Julia Thane Burleigh. &
\texttt{source\_path}: [brother, mother].
\texttt{source\_rule}: uncle.
\texttt{target\_rule}: uncle.
\texttt{target\_D}: Julia Thane Burleigh.
\texttt{answer}: Julia Thane Burleigh. \\
\hline
\end{tabular}
\caption{Concrete parsing examples for static reasoning tasks. Generated
solutions are copied verbatim from model outputs; parsed content is
reconstructed with the same parser schema used during evaluation. Dataset identifiers
are omitted.}
\label{tab:static-parsing-examples}
\end{table*}

The binary indicators used in Section~\ref{sec:train-setup} are therefore
\(A=\mathrm{m}_{\mathrm{A}}(\hat{a},a^\star)\) and
\(P=\mathrm{m}_{\mathrm{P}}(\hat{\pi},\pi)\). Intermediate parser diagnostics are
used only for analysis; reward and strict correctness use the binary indicators.

\subsubsection{RL Recipes and Task Instantiation}
\label{app:rl-recipes-task-instantiation}

Each recipe selects a subset of the \((D,T)\) grid and allocates the same task budget
reported in Section~\ref{sec:train-setup}. The main sampler recipes are drawn by cell;
uniform and diagonal recipes are produced by the corresponding curriculum or data-volume
schedules. Table~\ref{tab:rl-recipes} summarizes the 15 recipe specifications.

The Baseline recipe is a pre-training-region control. The Cmplx recipes, Cmplx-Mid,
Cmplx-High, Cmplx-Intersect, and Cmplx-Uniform, keep depth shallow and vary
environment complexity. The Depth recipes, Depth-Mid, Depth-High, Depth-Intersect,
and Depth-Uniform, keep environment complexity low and vary the depth axis. The Diag
recipes, Diag-Mid and Diag-High, move both axes together. The Mix recipes, Deep-Mix,
Shallow-Mix, Offbase-Mix, and Full-Coverage, compare broad-coverage distributions
that include or exclude the pre-training block while expanding depth and complexity
coverage.

\begin{table*}[t]
\centering
\footnotesize
\setlength{\tabcolsep}{3pt}
\begin{tabular}{p{0.13\linewidth}p{0.16\linewidth}p{0.27\linewidth}p{0.08\linewidth}p{0.28\linewidth}}
\hline
Family & Recipe & Coverage region & Cells & Source \\
\hline
Baseline & Baseline & \(D1\)--\(D4\), \(T1\)--\(T2\) & 8 & Main recipe sampler \\
Cmplx & Cmplx-Mid & \(D1\)--\(D4\), \(T3\)--\(T4\) & 8 & Main recipe sampler \\
Cmplx & Cmplx-High & \(D1\)--\(D4\), \(T5\)--\(T6\) & 8 & Main recipe sampler \\
Cmplx & Cmplx-Intersect & \(D1\)--\(D4\), \(T2\)--\(T3\) & 8 & Main recipe sampler \\
Cmplx & Cmplx-Uniform & \(D1\)--\(D4\), \(T1\)--\(T6\) & 24 & Cmplx-Uniform schedule \\
Depth & Depth-Mid & \(D5\)--\(D7\), \(T1\)--\(T2\) & 6 & Main recipe sampler \\
Depth & Depth-High & \(D8\)--\(D10\), \(T1\)--\(T2\) & 6 & Main recipe sampler \\
Depth & Depth-Intersect & \(D4\)--\(D6\), \(T1\)--\(T2\) & 6 & Main recipe sampler \\
Depth & Depth-Uniform & \(D1\)--\(D10\), \(T1\)--\(T2\) & 20 & Depth-Uniform schedule \\
Diag & Diag-Mid & \(D5\)--\(D7\), \(T3\)--\(T4\) & 6 & Diagonal middle block \\
Diag & Diag-High & \(D8\)--\(D10\), \(T5\)--\(T6\) & 6 & Diagonal hard block \\
Mix & Deep-Mix & \(D5\)--\(D10\), \(T3\)--\(T6\) & 24 & Main recipe sampler \\
Mix & Shallow-Mix & \(D1\)--\(D4\), \(T1\)--\(T6\); \(D5\)--\(D10\), \(T1\)--\(T2\) & 36 & Main recipe sampler \\
Mix & Offbase-Mix & all cells except \(D1\)--\(D4\), \(T1\)--\(T2\) & 52 & Main recipe sampler \\
Mix & Full-Coverage & \(D1\)--\(D10\), \(T1\)--\(T6\) & 60 & Main recipe sampler \\
\hline
\end{tabular}
\caption{RL training recipes over the depth--complexity grid.}
\label{tab:rl-recipes}
\end{table*}

For each recipe, tasks are distributed uniformly over its selected cells. The total task
budget is fixed at 220K tasks, with 140K deductive, 60K abductive, 10K inductive, and
10K analogy tasks. This fixed mixture makes recipe comparisons mainly reflect grid
coverage rather than changes in task-family proportion.

\subsubsection{Scale and Backbone Sweep Configurations}
\label{app:scale-backbone-configs}
Table~\ref{tab:scale-backbone-configs} lists the architectures used in the
model-scale and backbone ablations of Section~\ref{sec:ablation}. ``Qwen2'' is
the Hugging Face architecture class used by Qwen2.5-style models; the 107M row
is the main-paper model. Every sweep run repeats the full pipeline:
pre-training from scratch on the same KG corpus
(Section~\ref{app:pretraining-objective-hyperparameters}), followed by GRPO
post-training with the Shallow-Mix recipe
(Section~\ref{app:grpo-post-training-implementation}), and evaluation on the
same 60-cell evaluation set with the strict process-verified criterion.

\begin{table*}[t]
\centering
\footnotesize
\setlength{\tabcolsep}{5pt}
\begin{tabular}{llrrrrr}
\toprule
Model & Architecture & Layers & Hidden & Heads & KV heads & Intermediate \\
\midrule
Qwen2.5-style 51M & Qwen2 & 10 & 576 & 9 & 3 & 2304 \\
Qwen2.5-style 107M (main) & Qwen2 & 12 & 768 & 12 & 2 & 3072 \\
Qwen2.5-style 524M & Qwen2 & 22 & 1280 & 20 & 4 & 5120 \\
Qwen2.5-style 1.02B & Qwen2 & 30 & 1536 & 24 & 4 & 6144 \\
GPT-2-style 104M & GPT-2 & 14 & 768 & 12 & --- (MHA) & 3072 \\
Llama-style 105M & Llama & 12 & 768 & 12 & 2 & 3072 \\
\bottomrule
\end{tabular}
\caption{Model configurations for the scale and backbone ablations
(Section~\ref{sec:ablation}). Hidden and Intermediate denote the hidden size
and the feed-forward intermediate size (\texttt{n\_embd} and
\texttt{n\_inner} for GPT-2); GPT-2 uses multi-head attention without grouped
KV heads.}
\label{tab:scale-backbone-configs}
\end{table*}

\subsection{Experiment Evaluation Metrics}
\label{appx:experiment-evaluation-metrics}

This appendix gives the formal definitions used by the experiment sections. The main
text uses these metrics only at a high level, while the formulas below specify how the
reported recipe ranks, heatmaps, in/out-of-recipe gains, and task-pair correlations
are computed.

\paragraph{Strict pass@k, SG, and CG}
For recipe \(r\), task family \(q\), and evaluation subset \(S\), let
\(p_{r,q}^{(k)}(S)\) denote strict process-verified \(\mathrm{pass@}k\). The
pre-trained model is denoted by \(\mathrm{pre}\). Single Gain (SG) and Ceiling
Gain (CG) are
\[
\begin{aligned}
\mathrm{SG}_{r,q}(S)
&= p_{r,q}^{(1)}(S)-p_{\mathrm{pre},q}^{(1)}(S),\\
\mathrm{CG}_{r,q}(S)
&= p_{r,q}^{(128)}(S)-p_{\mathrm{pre},q}^{(128)}(S).
\end{aligned}
\]
SG measures the single-sample (pass@1) improvement, whereas CG measures the change in the
sampled reasoning ceiling under 128 attempts. As a worked example, the
Shallow-Mix recipe on deductive at \((D{=}5,T{=}3)\) yields
\(p^{(1)}_{\mathrm{Shallow\text{-}Mix},\mathrm{Ded}}\!=\!0.58\) versus
\(p^{(1)}_{\mathrm{pre},\mathrm{Ded}}\!=\!0.42\), giving
\(\mathrm{SG}\!=\!+0.16\); the same recipe-cell pair gives
\(p^{(128)}_{\mathrm{Shallow\text{-}Mix},\mathrm{Ded}}\!=\!0.79\) versus
\(p^{(128)}_{\mathrm{pre},\mathrm{Ded}}\!=\!0.64\), so
\(\mathrm{CG}\!=\!+0.15\). When the two gains disagree, the (SG, CG)
plane separates capability extension (both positive) from mode collapse
(positive SG with near-zero or negative CG).

\paragraph{Global Mean Rank}
Let \(C_{q,r}(d,t)=\mathrm{CG}_{r,q}(S_{d,t})\) be the cell-level ceiling gain of
recipe \(r\) on task family \(q\), where \(S_{d,t}\) is the evaluation subset at depth
\(d\) and complexity level \(t\). Let \(\Omega_q\) be the valid evaluated cells for task \(q\).
For a complexity level \(t\), define \(\Omega_q(t)=\{d:(d,t)\in\Omega_q\}\); for a depth \(d\),
define \(\Omega_q(d)=\{t:(d,t)\in\Omega_q\}\). The complexity-axis mean rank is
\[
\begin{aligned}
\bar C^{T}_{q,r}(t)
&=
\frac{1}{|\Omega_q(t)|}
\sum_{d:(d,t)\in\Omega_q} C_{q,r}(d,t),\\
\mathrm{MR}^{T}_{q,r}
&=
\frac{1}{|\mathcal{T}_q|}
\sum_{t\in\mathcal{T}_q}
\operatorname{rank}^{\downarrow}_{r}
\!\left(\bar C^{T}_{q,r}(t)\right),
\end{aligned}
\]
where \(\mathcal{T}_q\) is the set of evaluated complexity levels. The depth-axis mean rank is
defined symmetrically:
\[
\begin{aligned}
\bar C^{D}_{q,r}(d)
&=
\frac{1}{|\Omega_q(d)|}
\sum_{t:(d,t)\in\Omega_q} C_{q,r}(d,t),\\
\mathrm{MR}^{D}_{q,r}
&=
\frac{1}{|\mathcal{D}_q|}
\sum_{d\in\mathcal{D}_q}
\operatorname{rank}^{\downarrow}_{r}
\!\left(\bar C^{D}_{q,r}(d)\right),
\end{aligned}
\]
where \(\mathcal{D}_q\) is the set of evaluated depths. The operator
\(\operatorname{rank}^{\downarrow}_{r}\) ranks recipes by mean CG in descending order
within a slice, so rank 1 is best and lower mean rank is better.

\paragraph{Local Cell CG}
For each valid cell \((d,t)\in\Omega_q\), the local cell-level gain is
\[
\mathrm{CG}_{q,r}(d,t)
=
p_{r,q}^{(128)}(S_{d,t})
-
p_{\mathrm{pre},q}^{(128)}(S_{d,t}).
\]
The marginal heatmaps in the main text and appendix visualize these values directly.
Positive cells indicate sampled-ceiling expansion relative to the pre-trained model,
whereas negative cells indicate suppression. The in-recipe/out-of-recipe
split used in Section~\ref{sec:task-asymmetry} is defined by partitioning
\(\Omega_q\) into the cells that overlap recipe \(r\)'s post-training
distribution and the cells that do not; the per-cell formula above is
unchanged; only the aggregation set differs.

\paragraph{Task-Pair Pearson Correlation}
For task-pair analysis, we compare the CG patterns of two task families
\((q_a,q_b)\). Let \(u=(r,d,t)\) denote one recipe-cell observation and let
\(\mathcal{U}_{q_a,q_b}\) be the set of observations that are valid for both tasks.
In the full grid, \(|\mathcal{U}_{q_a,q_b}|=15\times60=900\). If a task pair has
invalid cells, we use only the common valid observations. For compactness, write
\(C_{q,u}=C_{q,r}(d,t)\). The global mean for task \(q\) over this pooled set is
\[
\bar C_q =
\frac{1}{|\mathcal{U}_{q_a,q_b}|}
\sum_{u\in\mathcal{U}_{q_a,q_b}} C_{q,u}.
\]
Let \(\mathcal{U}=\mathcal{U}_{q_a,q_b}\). The Pearson correlation is
\[
\begin{aligned}
N_{ab}
&=
\sum_{u\in\mathcal{U}}
(C_{q_a,u}-\bar C_{q_a})\\
&\quad
\times (C_{q_b,u}-\bar C_{q_b}),\\
Z_a
&=
\sqrt{\sum_{u\in\mathcal{U}}(C_{q_a,u}-\bar C_{q_a})^2},\\
Z_b
&=
\sqrt{\sum_{u\in\mathcal{U}}(C_{q_b,u}-\bar C_{q_b})^2},\\
r(q_a,q_b)
&=
\frac{N_{ab}}{Z_a Z_b}.
\end{aligned}
\]
We compute this correlation once over the merged recipe-cell observations, rather
than computing one correlation per recipe and averaging those values.
The same formula is applied to subsets of \(\mathcal{U}\) to obtain the
family- and region-conditional correlations reported in
Appendix~\ref{appx:inductive-analogy-experiments}; for example,
restricting \(\mathcal{U}\) to the four Depth-family recipes
(\(|\mathcal{U}|\!=\!4\times54\!=\!216\) valid observations) yields the
within-family Pearson \(r\!=\!0.62\) cited for the deductive--abductive
pair.

\FloatBarrier
\subsection{Statistical Significance of the Main Comparisons}
\label{appx:bootstrap}

All checkpoints in this paper come from a single training run
(Appendix~\ref{app:supplementary-material-arr}). To quantify whether the
headline gaps exceed evaluation noise, we compute paired bootstrap 95\%
confidence intervals (CIs) over evaluation items for the three main
comparisons: joint coverage versus single-axis recipes (Finding~1), abductive
out-of-recipe degradation (Finding~2), and uniform mixing versus staged
curriculum (Finding~3). For each comparison we resample evaluation items with
replacement 10{,}000 times, recompute the paired difference in strict
process-verified pass@1 (single sample) or pass@128 (sampled ceiling) on every
resample, and report the 2.5th and 97.5th percentiles. An interval that
excludes zero (``Excl.~0 = yes'') indicates that the gap exceeds
evaluation-sampling uncertainty. These intervals capture evaluation-sampling
uncertainty only; they do not bound run-to-run variance across training seeds.

\paragraph{Finding 1: Shallow-Mix versus single-axis recipes.}
Table~\ref{tab:bootstrap-finding1} compares Shallow-Mix with Cmplx-Uniform and
Depth-Uniform. The two uniform checkpoints were evaluated on a partially
overlapping evaluation sample, so comparisons use the \(N=11{,}700\) items
shared by the three checkpoints. The overall advantage of Shallow-Mix exceeds
evaluation noise on both comparisons (\(+0.055\), 95\% CI \([+0.045, +0.064]\),
and \(+0.072\), 95\% CI \([+0.063, +0.082]\), strict pass@1), as do the
deductive-slice gains, and the advantage also holds for the sampled ceiling
(strict pass@128: \(+0.076\) and \(+0.079\)). At the slice level, abductive is
at parity (intervals include zero), and the single-axis uniforms retain small
but significant edges on inductive and analogy (at most \(0.049\)).

\paragraph{Finding 2: Abductive out-of-recipe degradation.}
At the cell level, 131 of 396 out-of-recipe abductive cells have a negative CG
point estimate, but no individual cell reaches significance after
multiple-comparison correction (40 abductive items per cell). We therefore pool
each recipe's out-of-recipe abductive cells (240--1{,}920 items, 40 per cell) and
bootstrap the pooled CG (Table~\ref{tab:bootstrap-abductive}). The pooled CG is
significantly below zero for the four single-axis Mid/High recipes, Baseline,
and Offbase-Mix, whereas Shallow-Mix, Deep-Mix, and the two Intersect recipes
include zero; no recipe is significantly above zero. Full-Coverage has no
out-of-recipe cells.

\paragraph{Finding 3: Uniform mixing versus 3-block curriculum.}
Table~\ref{tab:bootstrap-finding3} compares the final checkpoints of the
uniform and 3-block easy-to-hard (E2H) runs on each axis, fully paired over
\(N=21{,}360\) items. Uniform mixing significantly outperforms the 3-block
curriculum on both axes (\(+0.037\), 95\% CI \([+0.033, +0.040]\), on the
complexity axis and \(+0.061\), 95\% CI \([+0.057, +0.065]\), on the depth
axis, strict pass@1), and the gaps persist for the sampled ceiling (strict
pass@128: \(+0.032\) and \(+0.068\)). The hard-to-easy (H2E) orderings are
likewise significantly below uniform mixing (\(+0.020\) and \(+0.043\), strict
pass@1). Slice-level deviations are small (at most \(0.014\), on inductive and
analogy).

\begin{table*}[t]
\centering
\footnotesize
\setlength{\tabcolsep}{4pt}
\resizebox{\textwidth}{!}{%
\begin{tabular}{llrccrcc}
\toprule
Comparison & Slice & \(N\) & \(\Delta\) pass@1 [95\% CI] & Excl.~0 & & \(\Delta\) pass@128 [95\% CI] & Excl.~0 \\
\midrule
\multirow{5}{*}{Shallow-Mix \(-\) Cmplx-Uniform}
 & Overall   & 11700 & \(+0.0550\) \([+0.0453, +0.0641]\) & yes & & \(+0.0755\) \([+0.0659, +0.0853]\) & yes \\
 & Deductive &  7824 & \(+0.0922\) \([+0.0808, +0.1037]\) & yes & & \(+0.1213\) \([+0.1093, +0.1331]\) & yes \\
 & Abductive &  1064 & \(-0.0028\) \([-0.0254, +0.0197]\) & no  & & \(+0.0169\) \([-0.0066, +0.0404]\) & no  \\
 & Inductive &  2359 & \(-0.0250\) \([-0.0479, -0.0017]\) & yes & & \(-0.0267\) \([-0.0492, -0.0038]\) & yes \\
 & Analogy   &   453 & \(-0.0353\) \([-0.0662, -0.0044]\) & yes & & \(-0.0464\) \([-0.0773, -0.0155]\) & yes \\
\midrule
\multirow{5}{*}{Shallow-Mix \(-\) Depth-Uniform}
 & Overall   & 11700 & \(+0.0724\) \([+0.0625, +0.0824]\) & yes & & \(+0.0791\) \([+0.0688, +0.0893]\) & yes \\
 & Deductive &  7824 & \(+0.1093\) \([+0.0966, +0.1221]\) & yes & & \(+0.1237\) \([+0.1111, +0.1362]\) & yes \\
 & Abductive &  1064 & \(+0.0179\) \([-0.0066, +0.0423]\) & no  & & \(+0.0188\) \([-0.0085, +0.0461]\) & no  \\
 & Inductive &  2359 & \(-0.0059\) \([-0.0297, +0.0174]\) & no  & & \(-0.0174\) \([-0.0398, +0.0055]\) & no  \\
 & Analogy   &   453 & \(-0.0287\) \([-0.0618, +0.0022]\) & no  & & \(-0.0486\) \([-0.0795, -0.0199]\) & yes \\
\bottomrule
\end{tabular}%
}
\caption{Finding 1: paired bootstrap 95\% CIs (10{,}000 resamples over
evaluation items) for Shallow-Mix minus each single-axis recipe, in strict
process-verified pass@1 (single sample) and pass@128 (sampled ceiling). Positive
\(\Delta\) favors Shallow-Mix. Comparisons use the \(N=11{,}700\) items shared
by the three checkpoints. ``Excl.~0'' marks intervals that exclude zero.}
\label{tab:bootstrap-finding1}
\end{table*}

\begin{table*}[t]
\centering
\footnotesize
\setlength{\tabcolsep}{4pt}
\begin{tabular}{lrrcc}
\toprule
Recipe & Cells & \(N\) & \(\Delta\)CG [95\% CI] & Excl.~0 \\
\midrule
Baseline        & 48 & 1920 & \(-0.0073\) \([-0.0130, -0.0016]\) & yes \\
Cmplx-Mid       & 48 & 1920 & \(-0.0120\) \([-0.0214, -0.0031]\) & yes \\
Cmplx-High      & 48 & 1920 & \(-0.0479\) \([-0.0630, -0.0333]\) & yes \\
Depth-Mid       & 48 & 1920 & \(-0.0328\) \([-0.0464, -0.0187]\) & yes \\
Depth-High      & 48 & 1920 & \(-0.0484\) \([-0.0646, -0.0328]\) & yes \\
Deep-Mix        & 30 & 1200 & \(+0.0000\) \([-0.0142, +0.0142]\) & no  \\
Shallow-Mix     & 24 &  960 & \(-0.0021\) \([-0.0146, +0.0094]\) & no  \\
Offbase-Mix     &  6 &  240 & \(-0.1792\) \([-0.2292, -0.1333]\) & yes \\
Cmplx-Intersect & 48 & 1920 & \(+0.0021\) \([-0.0073, +0.0115]\) & no  \\
Depth-Intersect & 48 & 1920 & \(+0.0005\) \([-0.0104, +0.0115]\) & no  \\
\bottomrule
\end{tabular}
\caption{Finding 2: abductive CG (strict pass@128) pooled over each recipe's
out-of-recipe cells, with paired bootstrap 95\% CIs over items (10{,}000
resamples). ``Cells'' is the number of out-of-recipe abductive cells and
\(N\) the number of pooled items. A negative \(\Delta\)CG with ``Excl.~0 =
yes'' means the post-training ceiling is significantly below that of the pre-trained
model.}
\label{tab:bootstrap-abductive}
\end{table*}

\begin{table*}[t]
\centering
\footnotesize
\setlength{\tabcolsep}{4pt}
\resizebox{\textwidth}{!}{%
\begin{tabular}{llrccrcc}
\toprule
Comparison & Slice & \(N\) & \(\Delta\) pass@1 [95\% CI] & Excl.~0 & & \(\Delta\) pass@128 [95\% CI] & Excl.~0 \\
\midrule
\multirow{5}{*}{Cmplx-Uniform \(-\) Cmplx 3-block (E2H)}
 & Overall   & 21360 & \(+0.0369\) \([+0.0333, +0.0404]\) & yes & & \(+0.0318\) \([+0.0283, +0.0352]\) & yes \\
 & Deductive & 14400 & \(+0.0492\) \([+0.0444, +0.0537]\) & yes & & \(+0.0422\) \([+0.0376, +0.0466]\) & yes \\
 & Abductive &  2160 & \(+0.0532\) \([+0.0417, +0.0653]\) & yes & & \(+0.0509\) \([+0.0398, +0.0625]\) & yes \\
 & Inductive &  3325 & \(-0.0069\) \([-0.0114, -0.0027]\) & yes & & \(-0.0051\) \([-0.0087, -0.0015]\) & yes \\
 & Analogy   &  1475 & \(-0.0081\) \([-0.0203, +0.0041]\) & no  & & \(-0.0142\) \([-0.0258, -0.0034]\) & yes \\
\midrule
\multirow{5}{*}{Depth-Uniform \(-\) Depth 3-block (E2H)}
 & Overall   & 21360 & \(+0.0612\) \([+0.0571, +0.0653]\) & yes & & \(+0.0676\) \([+0.0635, +0.0716]\) & yes \\
 & Deductive & 14400 & \(+0.0751\) \([+0.0695, +0.0806]\) & yes & & \(+0.0858\) \([+0.0803, +0.0912]\) & yes \\
 & Abductive &  2160 & \(+0.1005\) \([+0.0866, +0.1144]\) & yes & & \(+0.1037\) \([+0.0884, +0.1194]\) & yes \\
 & Inductive &  3325 & \(-0.0012\) \([-0.0048, +0.0024]\) & no  & & \(-0.0018\) \([-0.0048, +0.0009]\) & no  \\
 & Analogy   &  1475 & \(+0.0081\) \([-0.0034, +0.0197]\) & no  & & \(-0.0068\) \([-0.0169, +0.0027]\) & no  \\
\bottomrule
\end{tabular}%
}
\caption{Finding 3: paired bootstrap 95\% CIs for uniform mixing minus the
3-block easy-to-hard curriculum on each axis (final checkpoints, fully paired,
\(N=21{,}360\)), in strict process-verified pass@1 and pass@128. Positive
\(\Delta\) favors uniform mixing.}
\label{tab:bootstrap-finding3}
\end{table*}

\subsection{Additional Ablation Studies}
\label{appx:ablation-studies}

This appendix collects two further ablations of implementation choices that are
fixed in the main experiments; the data-volume, rollout-count, model-scale, and
backbone ablations are reported in Section~\ref{sec:ablation}. These studies
are not separate research questions; they check whether the main recipe-level
conclusions depend on the deductive--abductive task composition or on the
process-reward ratio.

\subsubsection{Task Mix}
\label{appx:ablation-task-mix-subsec}
\begin{figure*}[!t]
\centering
\includegraphics[width=\textwidth]{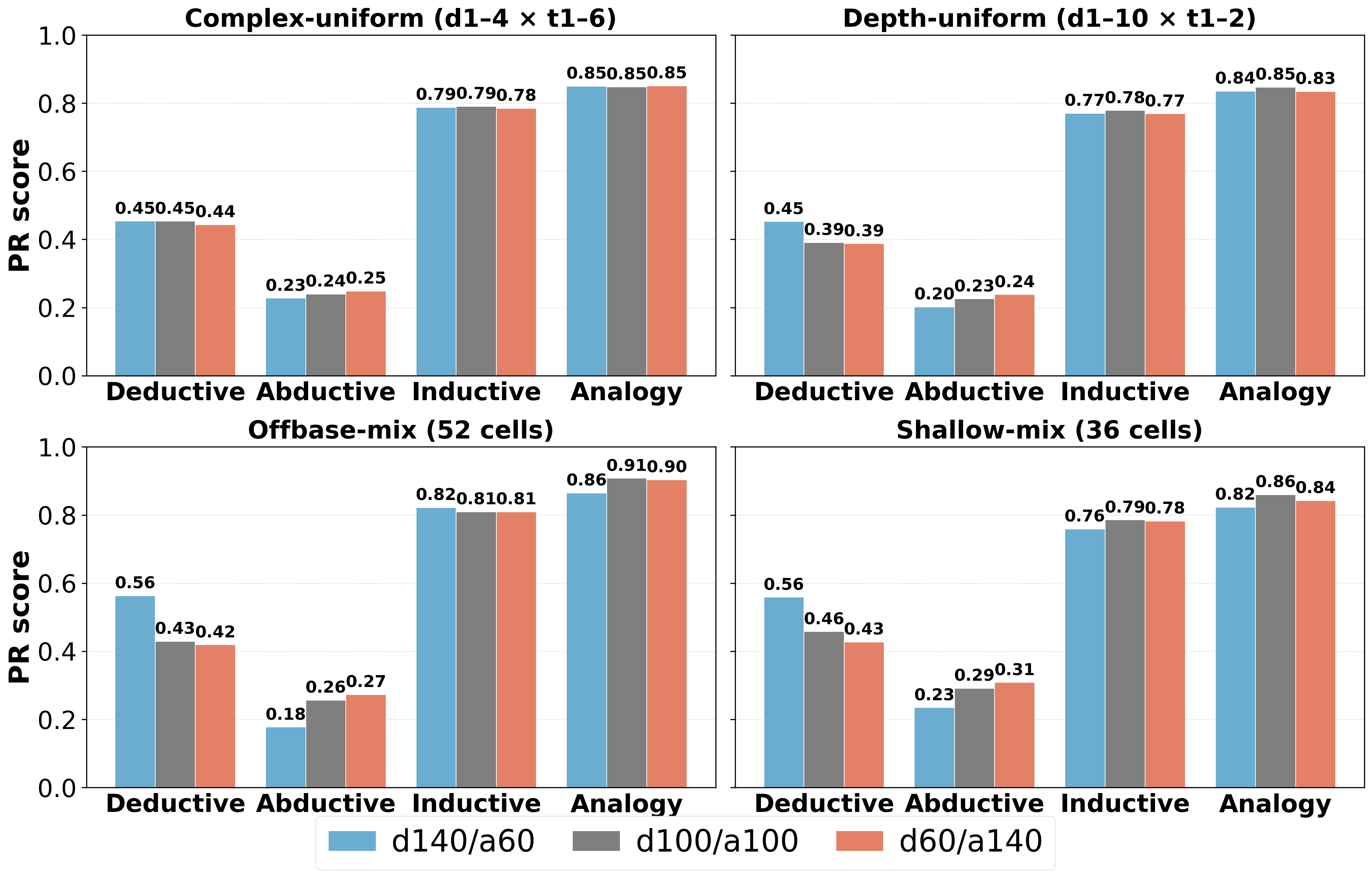}
\caption{Task-mix ablation under a fixed 220K budget across all four recipe
families (Cmplx-Uniform, Depth-Uniform, Offbase-Mix, Shallow-Mix). The three
mixed-task settings (d140/a60, d100/a100, d60/a140) vary the
deductive--abductive ratio while keeping the total budget approximately fixed.}
\label{fig:ablation-task-mix}
\end{figure*}

This ablation checks whether the main deductive--abductive training mixture is
important under a fixed post-training budget. We keep the total budget near 220K and
vary the internal deductive--abductive composition. Figure~\ref{fig:ablation-task-mix}
compares the three mixed-task settings across all four recipe families. Mixed
compositions are more stable than pure single-task training: d140/a60 (140K deductive
and 60K abductive tasks) gives the
strongest overall result, while increasing the abductive share improves abductive
performance but can trade off against other task families.

\begin{figure*}[!t]
\centering
\includegraphics[width=\textwidth]{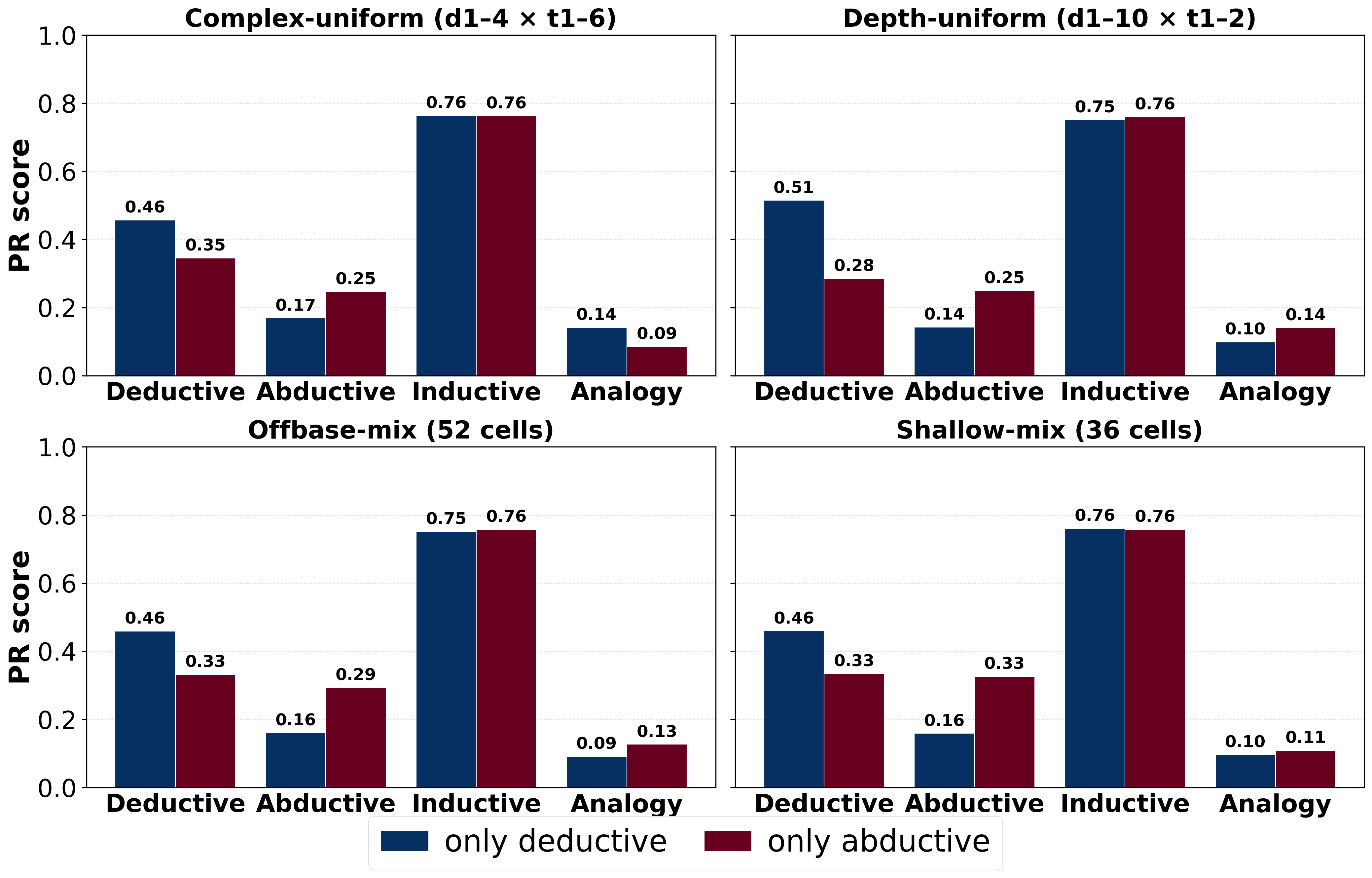}
\caption{Pure-task task-mix ablation under a fixed 220K budget. The figure
compares pure deductive and pure abductive post-training across the four recipe
families.}
\label{appx:ablation-task-mix-pure}
\end{figure*}

Figure~\ref{appx:ablation-task-mix-pure} provides the complementary pure-task
comparison. Pure deductive and pure abductive post-training are less stable than
mixed training across recipe families, which supports using a mixed dynamic-task
budget in the main experiments.

\subsubsection{Process-Reward Ratio}
\label{appx:ablation-process-reward-ratio}
\begin{figure*}[!t]
\centering
\begin{subfigure}{0.32\textwidth}
\centering
\includegraphics[width=\textwidth]{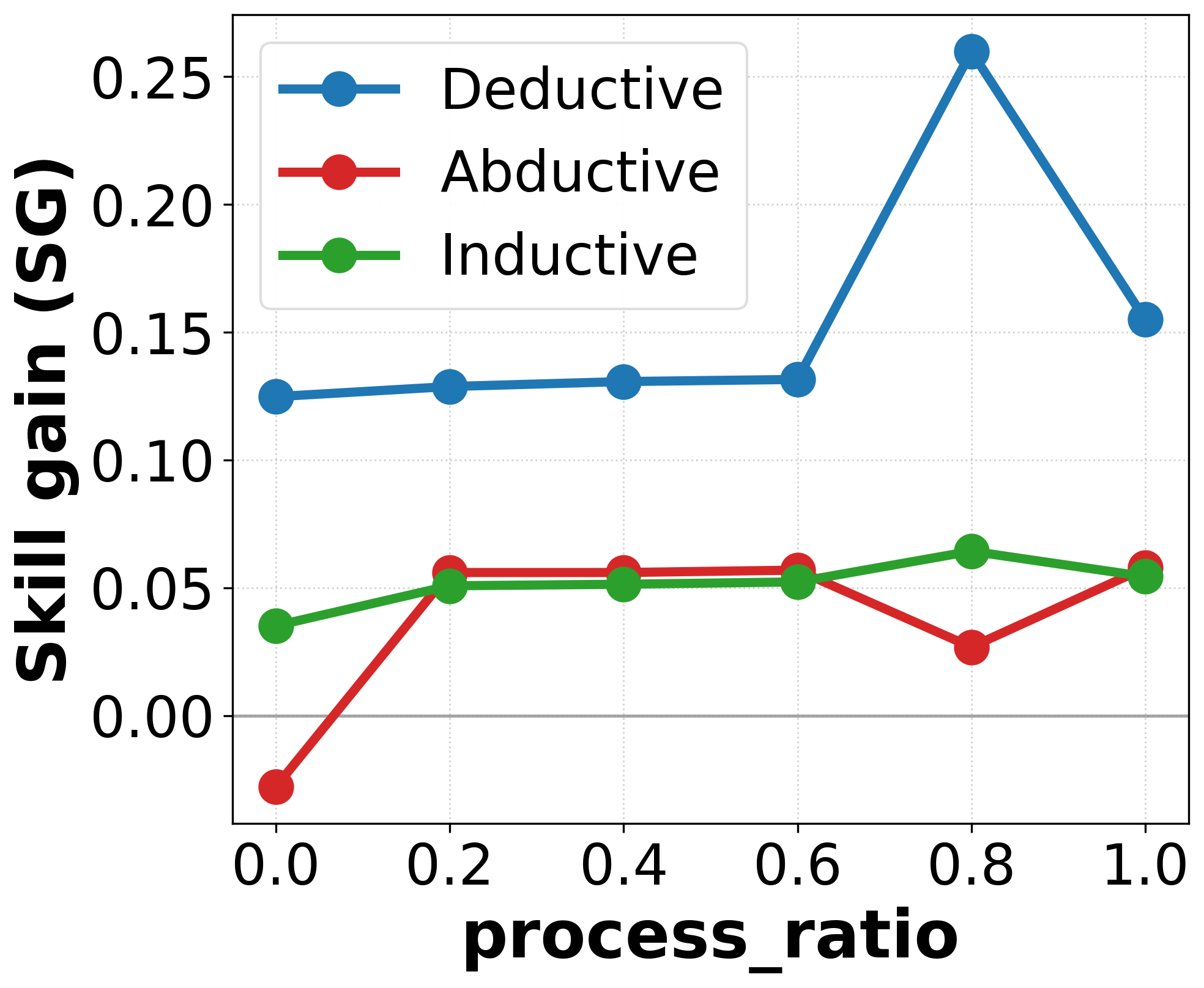}
\caption{Offbase-Mix.}
\label{fig:ablation-process-reward-ratio-offbase-mix}
\end{subfigure}
\hfill
\begin{subfigure}{0.32\textwidth}
\centering
\includegraphics[width=\textwidth]{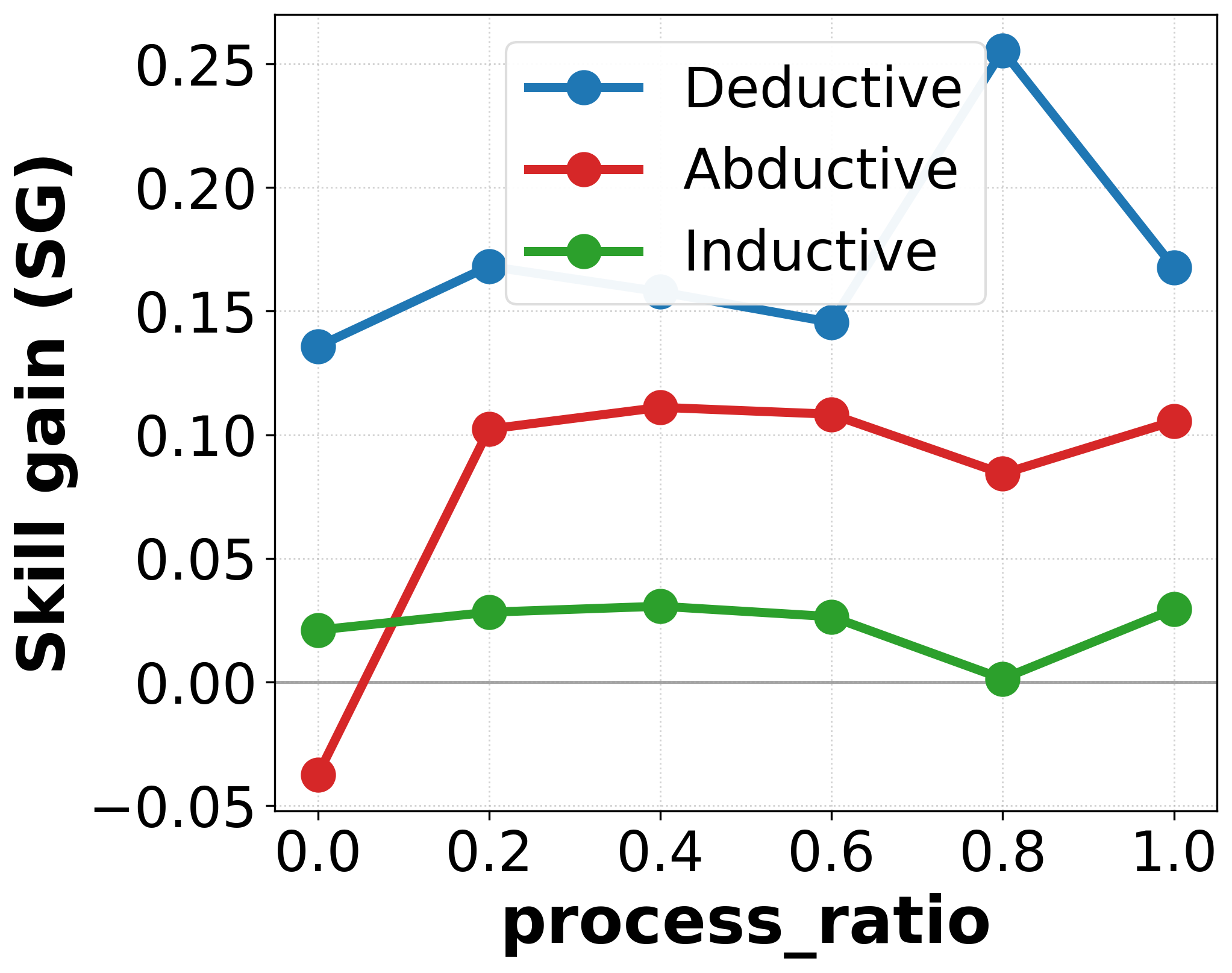}
\caption{Shallow-Mix.}
\label{fig:ablation-process-reward-ratio-shallow-mix}
\end{subfigure}
\hfill
\begin{subfigure}{0.32\textwidth}
\centering
\includegraphics[width=\textwidth]{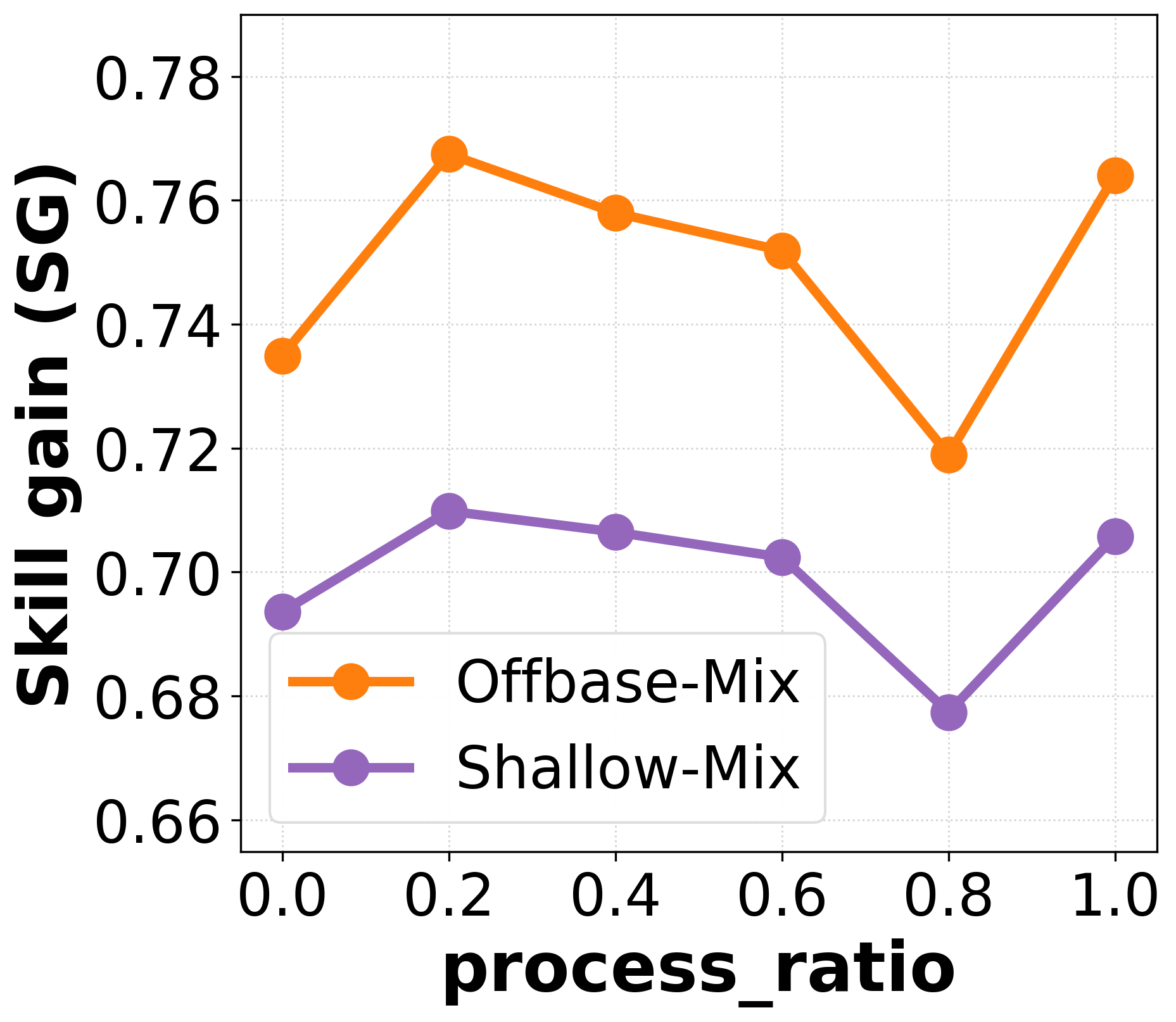}
\caption{Analogy across both recipes.}
\label{fig:ablation-process-reward-ratio-analogy}
\end{subfigure}
\caption{Process-reward-ratio ablation. Single Gain (SG, the pass@1 gain over
the pre-trained model; labeled ``Skill gain'' in the figure) as a function of the
process weight \(\lambda\) (\texttt{process\_ratio} in the figure). Panels (a) Offbase-Mix and (b) Shallow-Mix show the
three core task families (deductive, abductive, inductive); panel (c)
compares analogy across the two recipes on a separate axis because analogy
SG is much larger than the other families. Pure outcome reward
(\(\lambda = 0.0\)) hurts abductive in both recipes.}
\label{fig:ablation-process-reward-ratio}
\end{figure*}

This ablation examines how the trade-off between process- and outcome-based
reward signals affects RL post-training. We sweep the process weight
\(\lambda \in \{0.0, 0.2, 0.4, 0.6, 0.8, 1.0\}\) on two representative RL
recipes, Offbase-Mix and Shallow-Mix, keeping the total post-training budget
near 220K. Here \(\lambda\) is the trace-match weight in the outcome-gated
reward of Section~\ref{sec:train-setup}, with the outcome (final-answer) weight
fixed at \(1-\lambda\); \(\lambda = 0.8\) is the default used in the main
experiments. Figure~\ref{fig:ablation-process-reward-ratio} reports Single Gain
(SG, the pass@1 gain over the pre-trained model) as a function of \(\lambda\)
for the four task families. Pure outcome reward (\(\lambda = 0.0\)) degrades
abductive performance below the pre-trained baseline in both recipes;
introducing even a small process-reward component (\(\lambda = 0.2\)) recovers
and stabilizes it. Inductive and analogy gains remain positive but fluctuate
above \(\lambda = 0.2\), while deductive gains peak at the default
\(\lambda = 0.8\), supporting the choice used in the main experiments.

\subsection{Supplementary Experiments and Results}
\label{appx:supplementary-experiments-results}

This appendix collects supplementary figures that extend the main experimental
sections without changing their core claims. Each group below states which main-text
result it supports, what the figure measures, and what pattern should be read from
the figure.

\subsubsection{Full Marginal CG Heatmaps}
\label{appx:marginal-cg-heatmaps}

\begin{figure*}[!htbp]
\centering
\includegraphics[width=\textwidth]{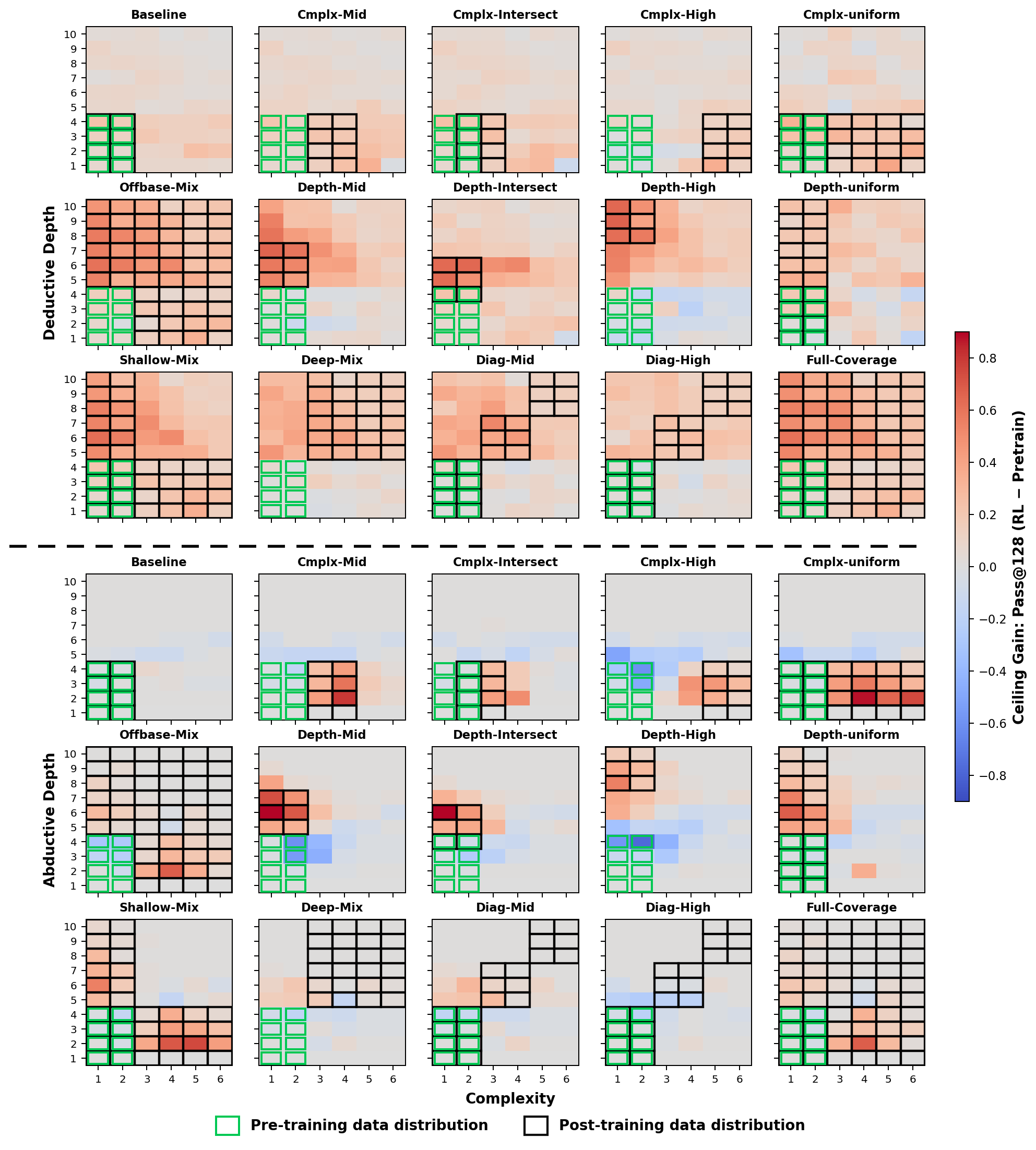}
\caption{Full 15-recipe per-cell marginal CG heatmaps on the 60-cell \((D,T)\) grid
for deductive (top three rows) and abductive (bottom three rows). Each task block
shares the same column order: top row is the baseline plus the four Cmplx
single-axis variants, middle row is Offbase-Mix plus the four Depth single-axis
variants, and bottom row groups the joint-coverage, diagonal, and full-coverage
recipes. Black borders mark each recipe's post-training cells, green borders mark
the pre-training cells \((D{\le}4, T{\le}2)\), and gray cells were not evaluated. The
color scale is shared with the inductive / analogy figure below.}
\label{fig:appx-marginal-heatmaps-da}
\end{figure*}

\begin{figure*}[!htbp]
\centering
\includegraphics[width=\textwidth]{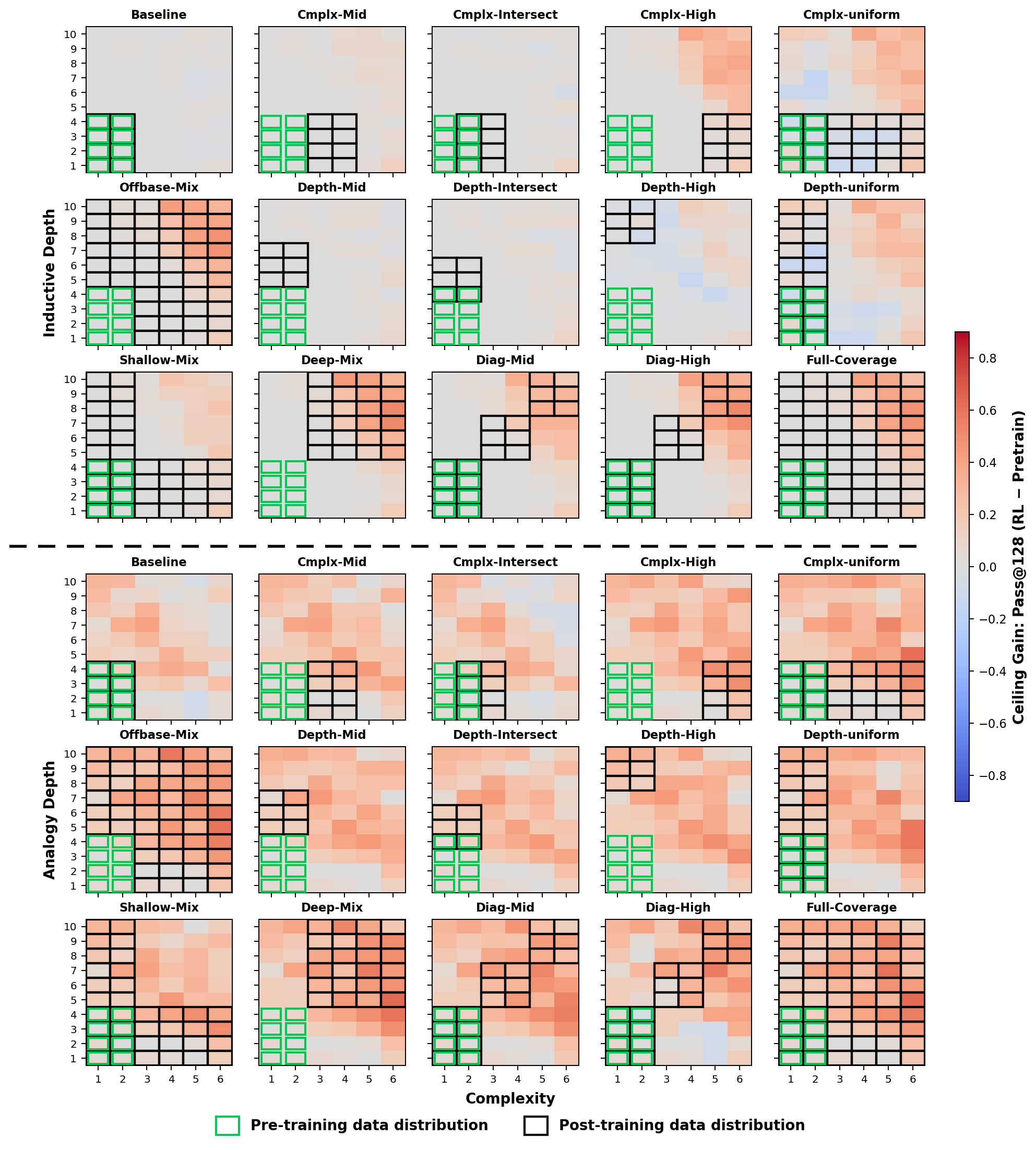}
\caption{Full 15-recipe per-cell marginal CG heatmaps on the 60-cell \((D,T)\) grid
for inductive (top three rows) and analogy (bottom three rows). The layout and the
color scale match the deductive / abductive figure above, so cells can be compared
across all four task families.}
\label{fig:appx-marginal-heatmaps-ia}
\end{figure*}

These heatmaps complement the data-distribution and task-family analyses in
Sections~\ref{sec:recipe-design} and~\ref{sec:task-asymmetry}. The main text shows
representative marginal CG heatmaps for deductive and abductive tasks; here we show
the full 15-recipe grid for all four task families
(Figures~\ref{fig:appx-marginal-heatmaps-da} and~\ref{fig:appx-marginal-heatmaps-ia}). The figures measure cell-level
CG on the 60-cell \((D,T)\) grid, so they reveal where each post-training recipe
expands or suppresses the sampled reasoning ceiling relative to the pre-trained
model. The main visible pattern is that joint-coverage recipes produce broader gains
than single-axis recipes, while abductive heatmaps contain more negative regions when
the post-training region does not support the relevant inverse-reasoning behavior.

These supplementary heatmaps also place our recipe design in context. Related RLVR
work asks whether RL mainly reweights behaviors already supported by the base model
or expands the model's reasoning boundary~\citep{ni2025grpo,yue2025does,liu2025prorl}. Our setting adds a distributional view of that
question: coverage determines which parts of the depth--complexity grid give RL
enough reward signal to use existing capability or extend it. Controlled reasoning
benchmarks vary difficulty axes for evaluation stress tests~\citep{mirzadeh2025gsmsymbolic,
wang2026can,samiei2026bridging}; here the same kind of
axis structure defines the post-training recipe itself, making the heatmaps a direct
view of where each recipe transfers, saturates, or suppresses capability.

Reading the heatmaps row by row, three patterns are immediately visible.
First, the in-recipe cells (black borders) are the most reliably positive
for every family except abductive, confirming that RL gains are densest
inside a recipe's own coverage and that out-of-recipe transfer is the
loose end. Second, joint-coverage recipes (Shallow-Mix, Offbase-Mix,
Full-Coverage) paint a much greener band across the depth axis for
deductive and abductive than the single-axis recipes, matching the
joint-coverage takeaway of Section~\ref{sec:recipe-design}. Third, the
abductive block shows visible \emph{negative} (red) cells outside the
pre-training region \((D{\le}4,T{\le}2)\), and these red cells are
concentrated in the same single-axis recipes that omit complexity
training; this localizes the abductive distribution sensitivity reported
in Section~\ref{sec:task-asymmetry} to specific axis-omission patterns.

\subsubsection{Task-Family Supplementary Results}
\label{appx:inductive-analogy-experiments}

\begin{figure}[!htbp]
\centering
\includegraphics[width=\columnwidth]{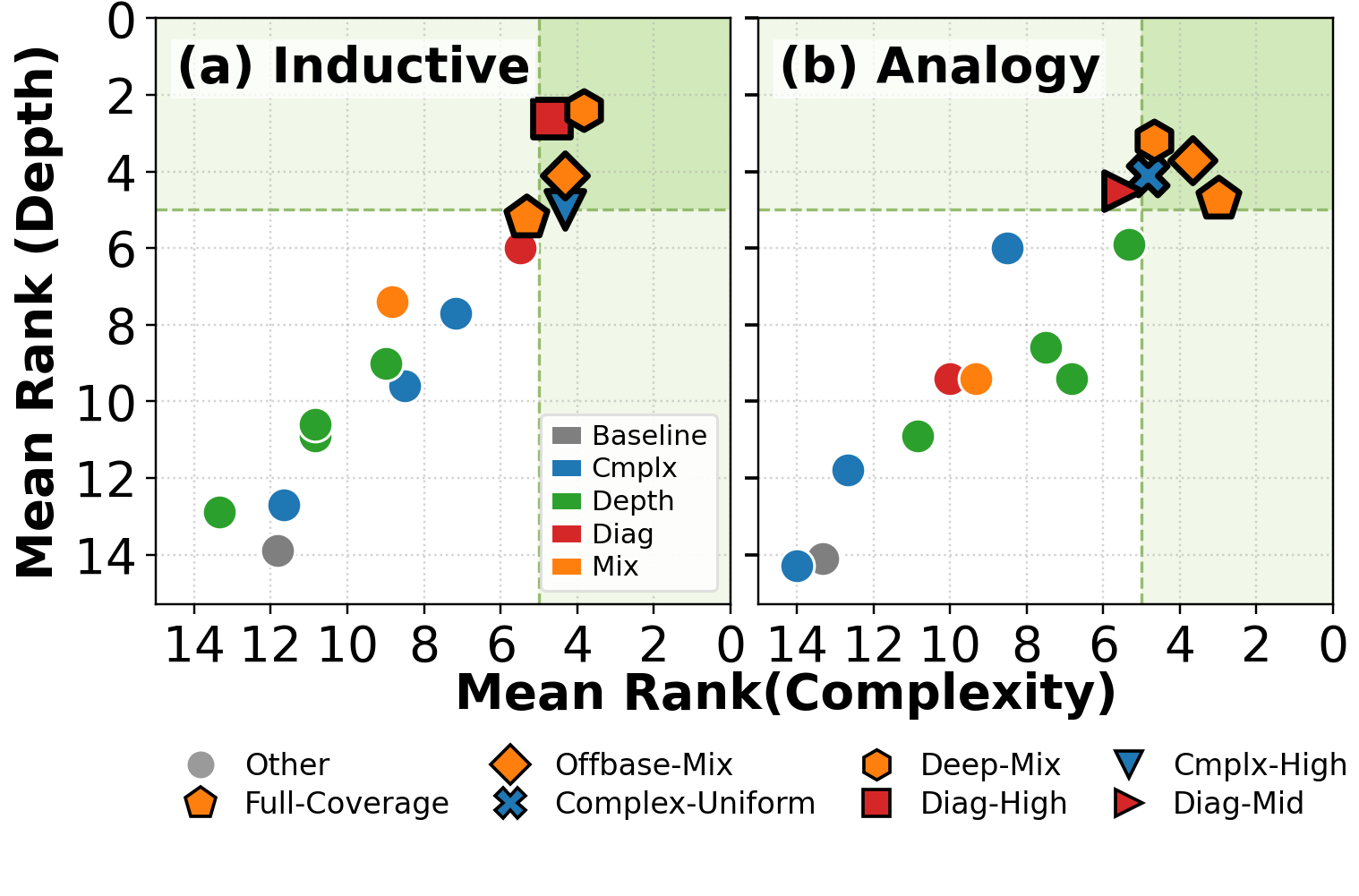}
\caption{Mean depth-rank versus mean complexity-rank for inductive and analogy. Each
point corresponds to one RL recipe. Lower values are better on both axes.
Joint-coverage recipes remain at least competitive with the best single-axis recipes.}
\label{fig:q1-overall-rank-ia}
\end{figure}

\begin{figure*}[!htbp]
\centering
\includegraphics[width=\textwidth]{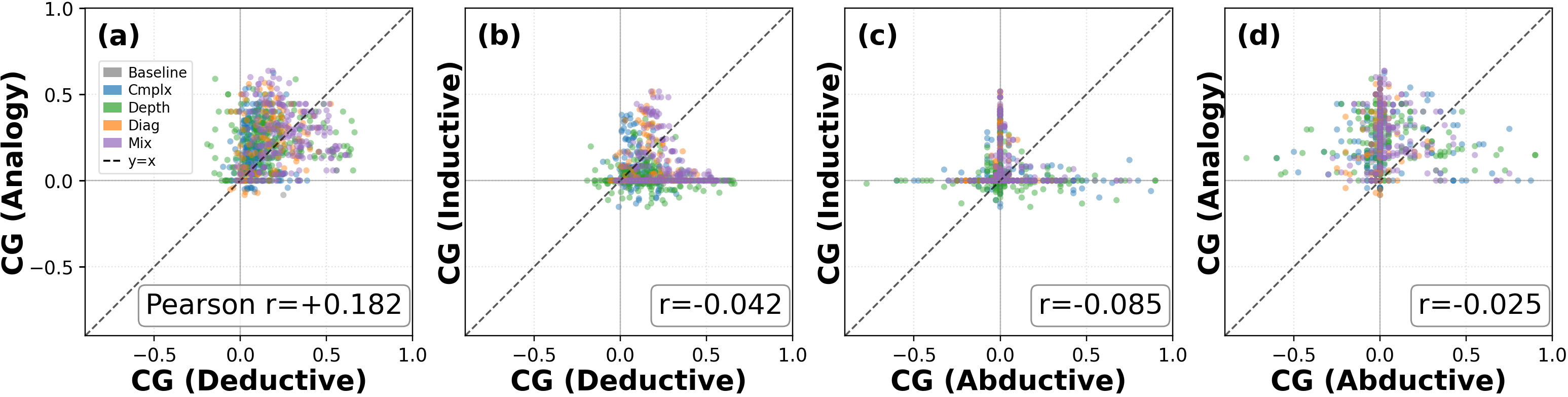}
\caption{Remaining task-pair CG correlations across the 60-cell grid and 15 RL
recipes. The deductive--analogy pair is moderately coupled, while the
deductive--inductive, abductive--inductive, and abductive--analogy pairs are weakly
coupled. Both axes report CG. The two strongest task pairs
(deductive--abductive and inductive--analogy) are shown in the main text in
Figure~\ref{fig:task-pair-correlation-strong}.}
\label{fig:task-pair-correlation-remaining}
\end{figure*}

These figures complement the main task-family analysis in
Section~\ref{sec:task-asymmetry}. The mean-rank plot in
Figure~\ref{fig:q1-overall-rank-ia} extends the recipe-ranking view from deductive
and abductive tasks to inductive and analogy tasks. It shows that joint-coverage
recipes remain at least competitive with the best single-axis recipes on both axes.
The remaining correlation plots in Figure~\ref{fig:task-pair-correlation-remaining}
complete the task-pair correlation analysis: deductive--analogy is moderately
coupled, while deductive--inductive, abductive--inductive, and abductive--analogy
are weakly coupled. This supports the main claim that reasoning task families do not
form one uniform RL regime.

\begin{figure}[!htbp]
\centering
\includegraphics[width=\columnwidth]{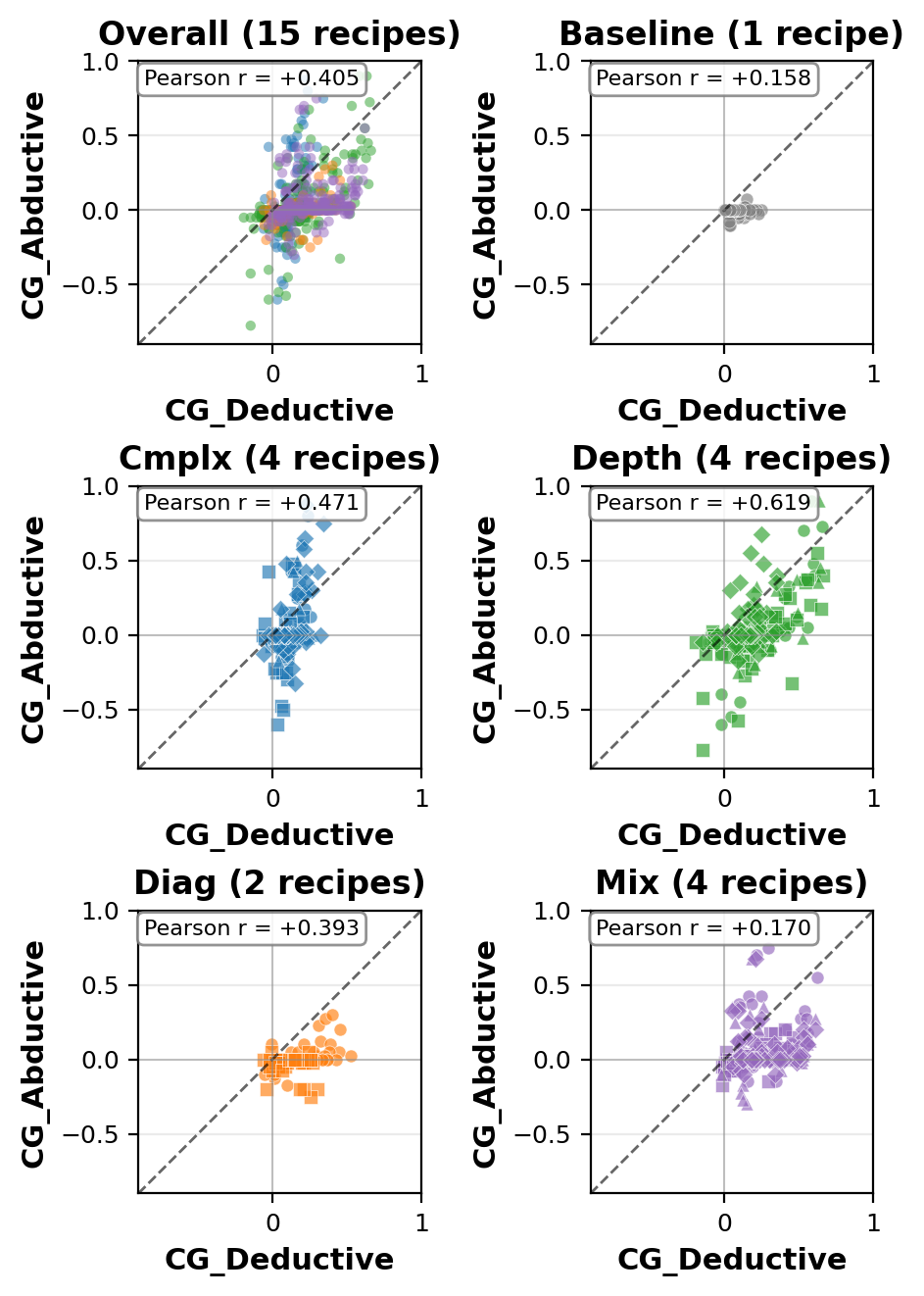}
\caption{Deductive--abductive correlation drill-down by recipe family.
``Overall'' uses all 15 recipes; each subsequent panel restricts to one
recipe family and recomputes the Pearson correlation within that family.
The Depth family (\(r=0.62\)) shows the strongest within-family coupling,
substantially above the overall \(r=0.40\); the Mix family (\(r=0.17\)) and
Baseline (\(r=0.16\)) are weaker.}
\label{fig:task-pair-drilldown-de-vs-ab-family}
\end{figure}

\begin{figure}[!htbp]
\centering
\includegraphics[width=\columnwidth]{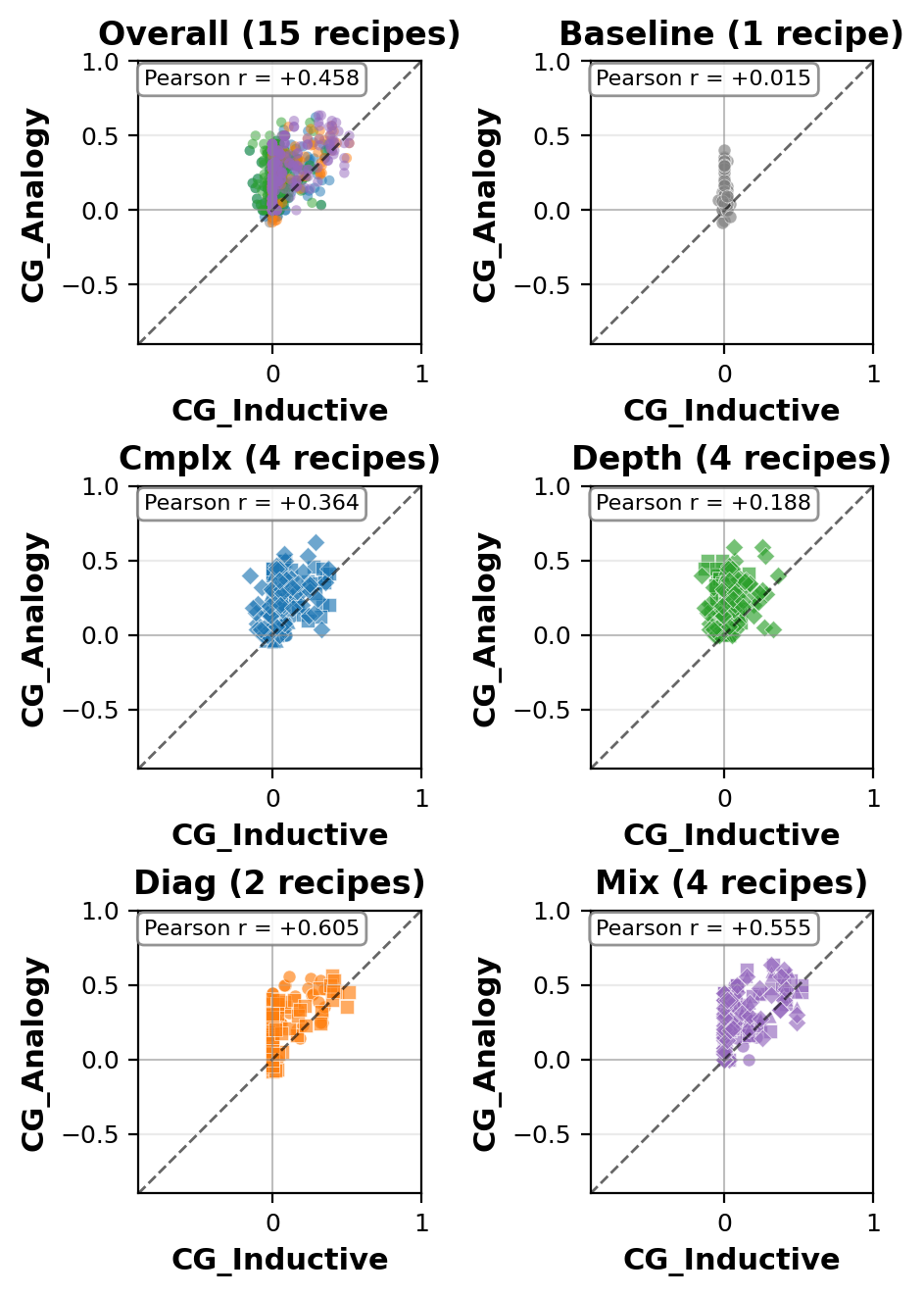}
\caption{Inductive--analogy correlation drill-down by recipe family. Same
layout and convention as Figure~\ref{fig:task-pair-drilldown-de-vs-ab-family}.
The within-family pattern differs from deductive--abductive: Diag
(\(r=0.61\)) and Mix (\(r=0.56\)) families show the strongest coupling,
while the Depth family is comparatively weak (\(r=0.19\)).}
\label{fig:task-pair-drilldown-in-vs-an-family}
\end{figure}

\paragraph{Task-pair correlation drill-downs by recipe family.}
The overall Pearson correlations reported in
Figures~\ref{fig:task-pair-correlation-strong} and
\ref{fig:task-pair-correlation-remaining} aggregate across all 15 recipes and
60 cells. To localize where the deductive--abductive and inductive--analogy
couplings come from, we drill the two strong pairs down by recipe family
(Baseline, Cmplx, Depth, Diag, Mix). Figures~\ref{fig:task-pair-drilldown-de-vs-ab-family}
and \ref{fig:task-pair-drilldown-in-vs-an-family} show the family drill-downs.
Within-family correlations are heterogeneous rather than uniformly stronger:
for the deductive--abductive pair the overall Pearson \(r=0.40\) rises to
\(r=0.62\) in the Depth family and \(r=0.47\) in the Cmplx family, but drops
to \(r=0.17\) in the Mix family and \(r=0.16\) for the Baseline. For
inductive--analogy the strongest within-family couplings are in Diag
(\(r=0.61\)) and Mix (\(r=0.56\)), while the Depth family is weakest
(\(r=0.19\)). The overall positive correlations therefore reflect
family-specific mechanisms rather than a uniform global property: depth-axis
training keeps deductive and abductive coupled, while diagonal or balanced
mixing keeps inductive and analogy coupled.

\subsubsection{Pass@k Exploration Profile}
\label{appx:passk-curves}

\begin{figure*}[!htbp]
\centering
\includegraphics[width=\textwidth]{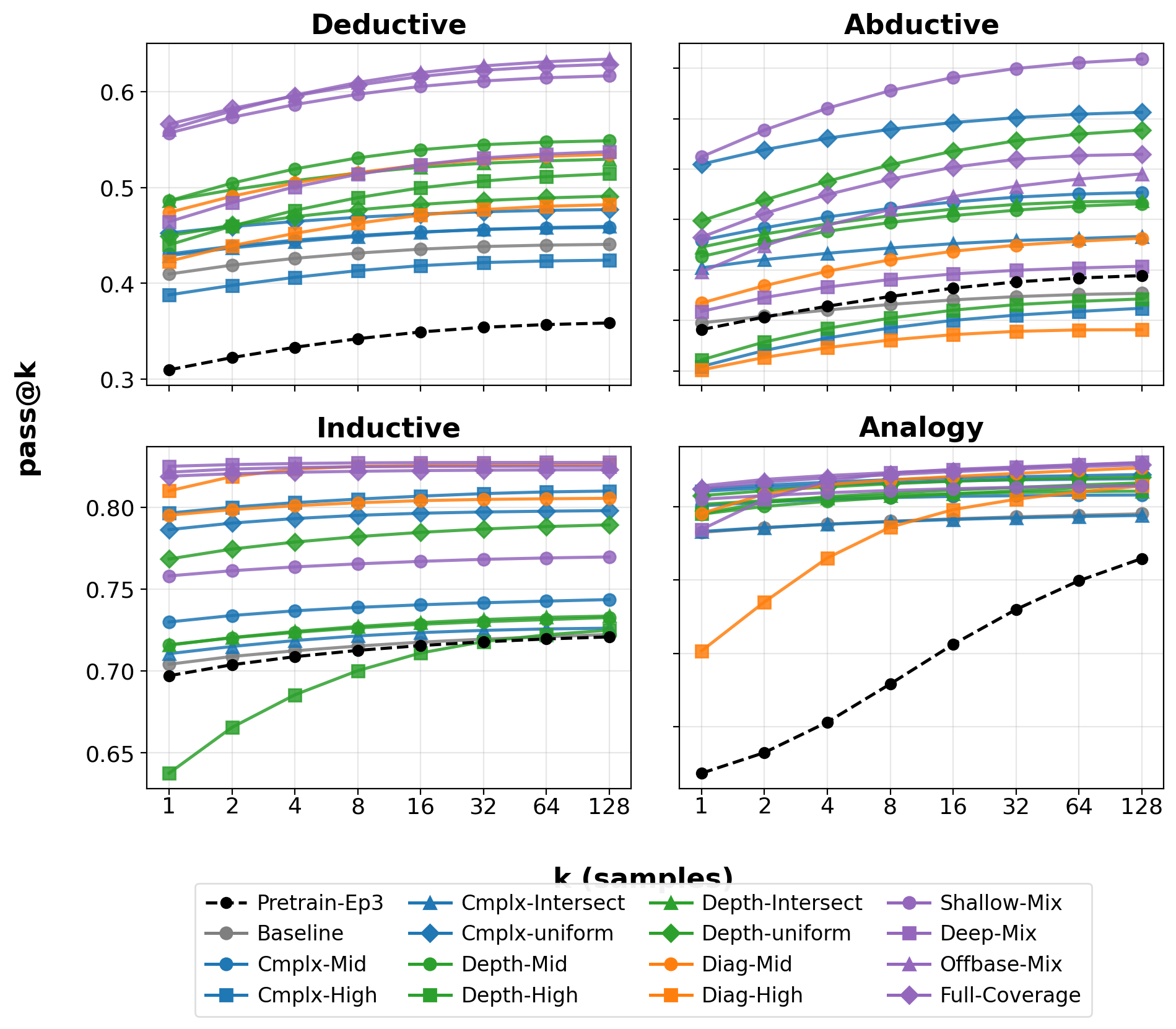}
\caption{Pass@k curves per task family for the 15 RL recipes and the
pre-trained model (black dashed; labeled Pretrain-Ep3 in the legend). The \(x\)-axis is the sample budget
\(k\) on a log-2 scale. RL recipes are colored by family (gray Baseline,
blue Cmplx, green Depth, orange Diag, purple Mix), with marker variation
within each family. Deductive shows consistent RL gains across \(k\), while
abductive gains are smaller and recipe-dependent; inductive shows shrinking RL
advantage as \(k\) grows; analogy
shows the clearest mode-collapse pattern, with most RL recipes flat near
pass@\(k\)\,\(\approx 0.85\) while the baseline's pass@k rises sharply.}
\label{fig:appx-passk-curves-per-task}
\end{figure*}

These pass@k curves complement the SG/CG analysis in
Section~\ref{sec:task-asymmetry} by showing the full exploration profile
behind a model's pass@1 (single sample) and pass@128 (ceiling) summary statistics.
Figure~\ref{fig:appx-passk-curves-per-task} plots pass@k for
\(k\in\{1,2,4,8,16,32,64,128\}\) on a log-\(k\) axis for the 15 RL recipes
together with the pre-trained model, separately per task family.

The four task families show qualitatively different exploration profiles.
For \emph{deductive}, every RL recipe clearly dominates the baseline across
the whole \(k\) range; joint-coverage Mix-family recipes occupy the top
envelope at every \(k\). For \emph{abductive}, gains are smaller and
recipe-dependent: several RL curves track the baseline closely, while the best
recipes shift the curve upward without changing its overall shape. For
\emph{inductive}, the baseline's pass@k climbs steeply with
\(k\) and approaches the RL recipes by \(k=128\); the RL advantage is
concentrated at small \(k\) and shrinks with sampling. For \emph{analogy},
the baseline shows the largest exploration headroom (pass@1
\(\approx 0.10\) climbing to pass@128 \(\approx 0.65\)), while almost every
RL recipe sits at pass@\(k\)\,\(\approx 0.85\) and is essentially flat
across \(k\)---the canonical signature of mode collapse, consistent with
the analogy pass@128-vs-pass@1 collapse highlighted in
Section~\ref{sec:task-asymmetry} and the (SG, CG) plane of
Figure~\ref{fig:task-asymmetry-sg-cg}. The single non-collapsed analogy
exception (Diag-High, orange triangle) shows a normal upward sweep
(pass@1 \(\approx 0.55\) to pass@128 \(\approx 0.85\)) and visually
confirms that mode collapse is recipe-conditional rather than intrinsic to
the task.

\subsubsection{Benchmark Supplementary Results}
\label{appx:benchmark-evaluation}

This appendix complements the benchmark evaluation in
Section~\ref{sec:benchmark-evaluation}. Table~\ref{tab:benchmark-model-metadata}
records the model metadata behind the leaderboard. Public models are cited from
official pages, Hugging Face model cards, arXiv papers, or official documentation.
Offbase-Mix, Shallow-Mix, Cmplx-Uniform, and Depth-Uniform are internal 107M
Qwen2.5-style checkpoints from this work. For FP8 releases, the table reports the
architecture scale rather than Hugging Face tensor-artifact totals; GPT-5.4 does
not disclose its parameter count.

\begin{table*}[t]
\centering
\scriptsize
\setlength{\tabcolsep}{3pt}
\begin{tabular}{@{}>{\raggedright\arraybackslash}p{0.25\textwidth}
                >{\raggedright\arraybackslash}p{0.18\textwidth}
                >{\raggedright\arraybackslash}p{0.20\textwidth}
                >{\raggedright\arraybackslash}p{0.29\textwidth}@{}}
\toprule
Model & Release / source & Scale & Notes \\
\midrule
Kimi-K2.6~\citep{moonshot2026kimik26}
& Apr. 2026, Moonshot / Hugging Face
& MoE, about 1T total / 32B activated
& Open-weight Kimi K2.6 model card. \\
GPT-5.4~\citep{openai2026gpt54}
& Mar. 2026, OpenAI
& Undisclosed
& Official OpenAI release. \\
GLM-5.1-FP8~\citep{zai2026glm51fp8}
& Apr. 2026, Z.ai / Hugging Face
& MoE, about 754B artifact scale, FP8
& FP8 release of GLM-5.1; model-family citation points to the GLM-5 technical report. \\
\textbf{Offbase-Mix}
& This work
& 107M Qwen2.5-style decoder
& Internal RL-trained checkpoint. \\
\textbf{Shallow-Mix}
& This work
& 107M Qwen2.5-style decoder
& Internal RL-trained checkpoint. \\
DeepSeek-R1~\citep{guo2025deepseek}
& Sep. 2025, DeepSeek / Nature (arXiv Jan. 2025)
& MoE, 671B total / 37B activated
& Open reasoning model trained with large-scale RL. \\
Qwen3.6-35B-A3B~\citep{qwen2026qwen36a3b}
& Apr. 2026, Qwen / Hugging Face
& MoE, 35B total / 3B activated
& Open-weight Qwen3.6 model. \\
Mixtral-8x22B-Instruct-v0.1~\citep{mistral2024mixtral8x22b}
& Apr. 2024, Mistral AI
& MoE, 141B total / 39B activated
& Official instruct release of Mixtral 8x22B. \\
Kimi-K2-Thinking~\citep{moonshot2025kimik2thinking}
& Nov. 2025, Moonshot / Hugging Face
& MoE, 1T total / 32B activated
& Thinking-agent version of Kimi K2 with long-context tool use. \\
Qwen3.5-397B-A17B-FP8~\citep{qwen2026qwen35_397b}
& Feb. 2026, Qwen / Hugging Face
& MoE, 397B total / 17B activated, FP8
& FP8 quantized release; table reports architecture scale. \\
\textbf{Cmplx-Uniform}
& This work
& 107M Qwen2.5-style decoder
& Internal RL-trained checkpoint. \\
\textbf{Depth-Uniform}
& This work
& 107M Qwen2.5-style decoder
& Internal RL-trained checkpoint. \\
Qwen3.5-35B-A3B~\citep{qwen2026qwen35_35b}
& Feb. 2026, Qwen / Hugging Face
& MoE, 35B total / 3B activated
& Qwen3.5 open-weight model. \\
QwQ-32B-Preview~\citep{qwen2024qwqpreview}
& Nov. 2024, Qwen / Hugging Face
& Dense, 32B
& Preview reasoning model based on Qwen2.5-32B. \\
Qwen2.5-32B-Instruct~\citep{qwen2025qwen25technicalreport}
& Sept. 2024, Qwen / Hugging Face
& Dense, 32.8B
& Instruction-tuned Qwen2.5 model. \\
Qwen3-8B~\citep{yang2025qwen3}
& Apr. 2025, Qwen / Hugging Face
& Dense, 8.2B
& Qwen3 dense open-weight model. \\
DeepSeek-R1-Distill-Qwen-32B~\citep{guo2025deepseek}
& Jan. 2025, DeepSeek / Hugging Face
& Dense, 32.8B
& Distilled from DeepSeek-R1 using the Qwen2.5-32B base. \\
Qwen2.5-72B-Instruct~\citep{qwen2025qwen25technicalreport}
& Sept. 2024, Qwen / Hugging Face
& Dense, 72.7B
& Instruction-tuned Qwen2.5 flagship model. \\
Qwen3.6-27B~\citep{qwen2026qwen3627b}
& Apr. 2026, Qwen / Hugging Face
& Dense, 27B
& Dense Qwen3.6 open-weight model. \\
Qwen2.5-7B-Instruct~\citep{qwen2025qwen25technicalreport}
& Sept. 2024, Qwen / Hugging Face
& Dense, 7.6B
& Instruction-tuned Qwen2.5 model. \\
Qwen3.5-122B-A10B-FP8~\citep{qwen2026qwen35_122b}
& Feb. 2026, Qwen / Hugging Face
& MoE, 122B total / 10B activated, FP8
& FP8 quantized release; table reports architecture scale. \\
DeepSeek-R1-Distill-Qwen-7B~\citep{guo2025deepseek}
& Jan. 2025, DeepSeek / Hugging Face
& Dense, 7.6B
& Distilled from DeepSeek-R1 using the Qwen2.5-7B base. \\
\bottomrule
\end{tabular}
\caption{Metadata for the benchmark models in Table~\ref{tab:benchmark-leaderboard-calibrated}.}
\label{tab:benchmark-model-metadata}
\end{table*}

\begin{figure*}[!htbp]
\centering
\includegraphics[width=0.62\textwidth]{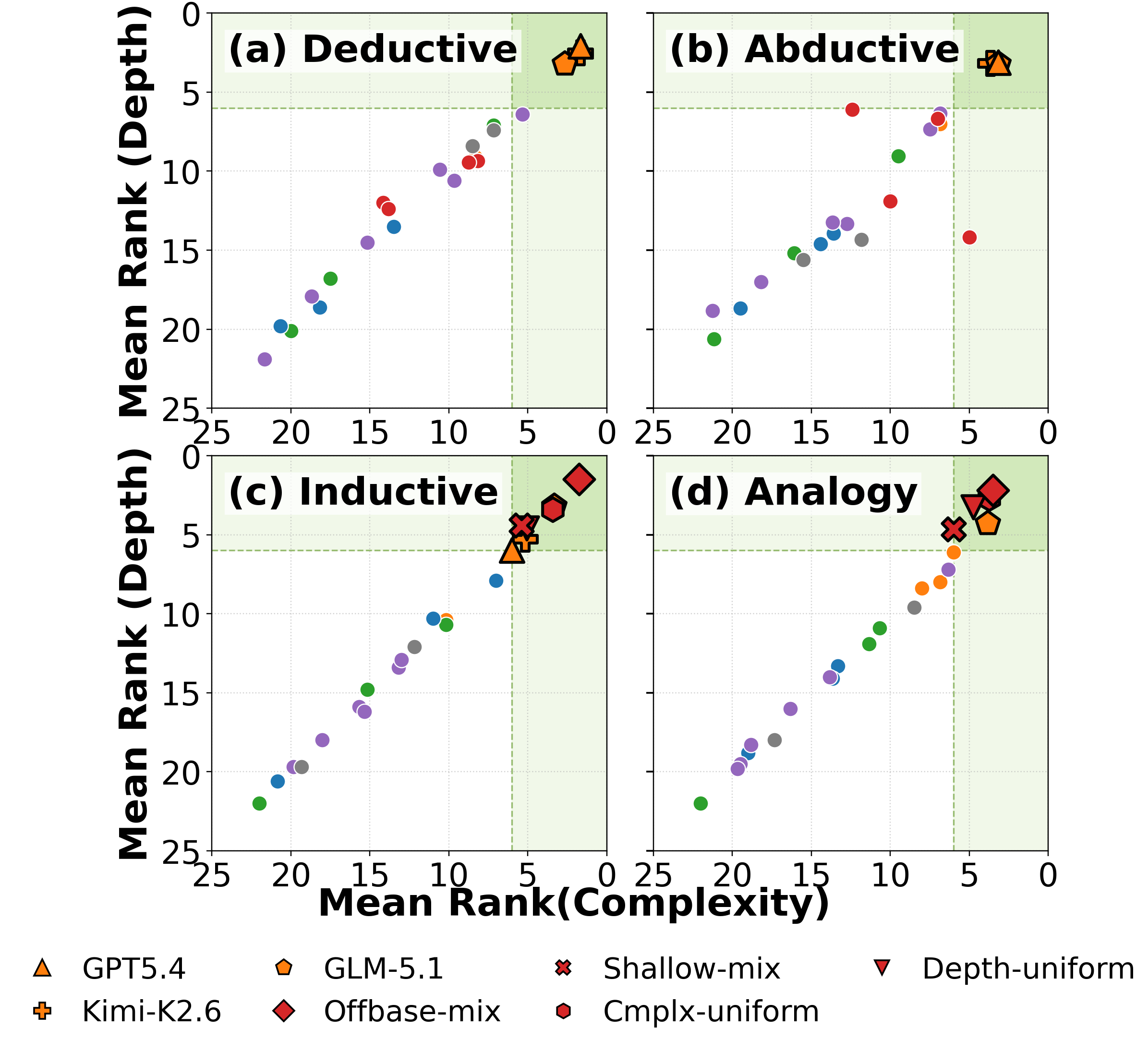}
\caption{Cross-axis mean-rank benchmark over 22 models. Each panel ranks models
by their mean performance across depth and complexity tiers for one reasoning
family. Labels are shown only for models in the top-right dual-winner region.}
\label{fig:benchmark-cross-axis-rank}
\end{figure*}

The main text reports the benchmark leaderboard, while
Figure~\ref{fig:benchmark-cross-axis-rank} shows how models rank across the depth and
complexity axes within each reasoning family. The benchmark pattern supports the
main task-family asymmetry: strong models can do well on deductive tasks while
remaining much weaker on abductive tasks.

\subsubsection{Best RL Recipe per Region}
\label{appx:best-recipe-per-region}

\begin{figure*}[!htbp]
\centering
\includegraphics[width=0.88\textwidth]{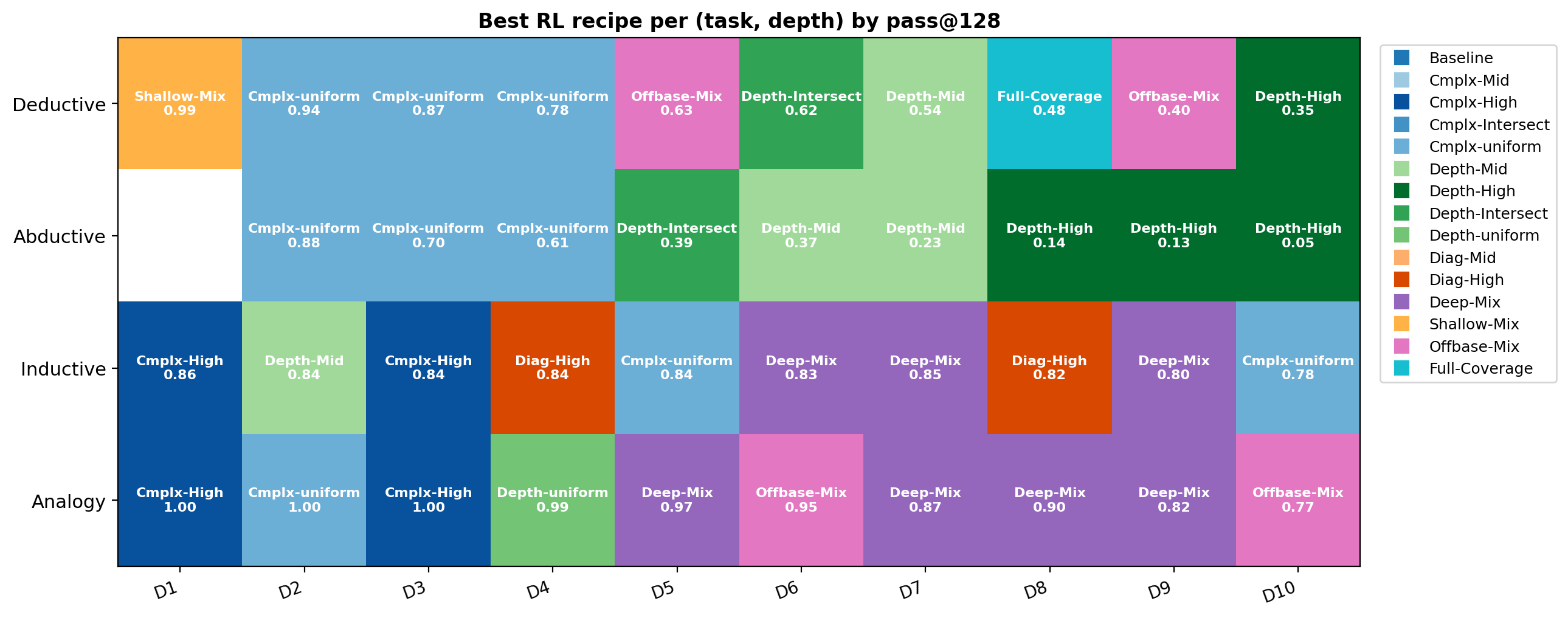}
\caption{Best RL recipe per (task, depth) cell by pass@128. Rows are task
families; columns are depth values \(D=1,\ldots,10\). Each cell is colored
by the winning recipe and labeled with the recipe name plus its pass@128
score. Deductive and abductive winners shift from Cmplx-\(\ast\) at shallow
depth to Depth- or Mix-family recipes at larger depth, with absolute pass@128
dropping sharply; inductive and analogy stay in a narrow high band across
depths despite changes in the winning recipe.}
\label{fig:appx-recipe-winner-depth}
\end{figure*}

\begin{figure*}[!htbp]
\centering
\includegraphics[width=0.88\textwidth]{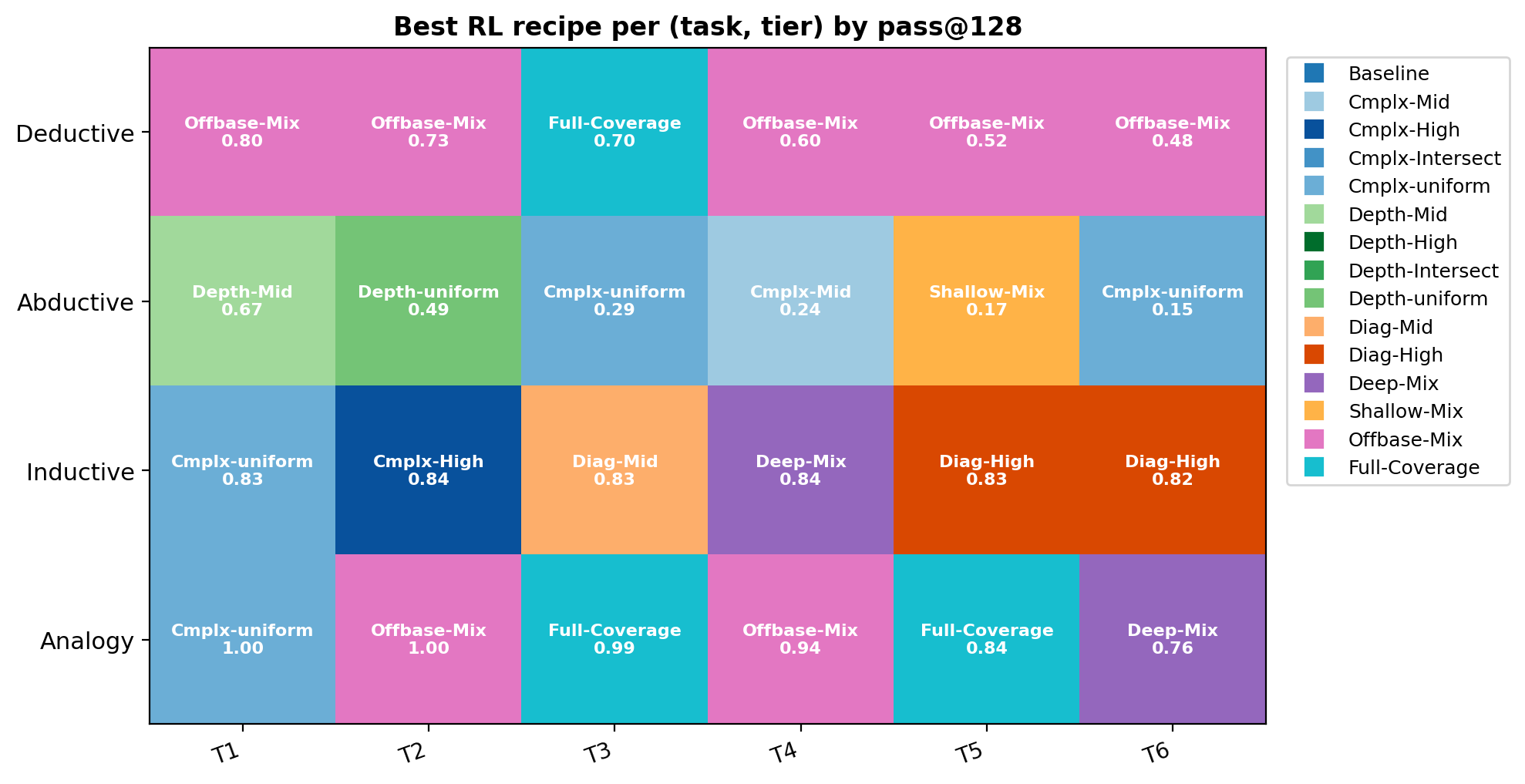}
\caption{Best RL recipe per (task, complexity tier) cell by pass@128. Same
convention as Figure~\ref{fig:appx-recipe-winner-depth}, but the
\(x\)-axis is the complexity tier \(T=1,\ldots,6\). Cmplx-family recipes
appear in several low-complexity columns, while higher-complexity columns mix
Depth-, Mix-, and Diag-family winners depending on the task.}
\label{fig:appx-recipe-winner-tier}
\end{figure*}

\begin{figure*}[!htbp]
\centering
\includegraphics[width=0.88\textwidth]{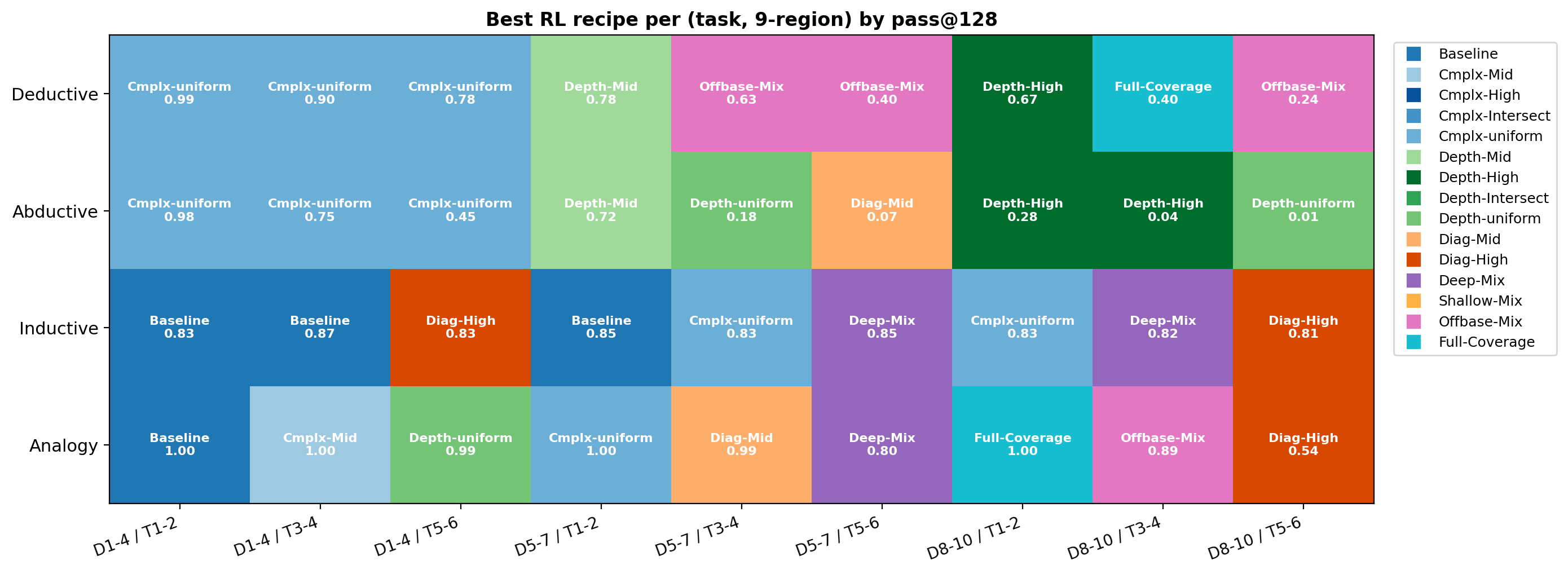}
\caption{Best RL recipe per (task, region) cell by pass@128. The nine
regions partition the 60-cell grid into depth bands
(\(D\in[1,4],[5,7],[8,10]\)) crossed with complexity bands
(\(T\in[1,2],[3,4],[5,6]\)). Same convention as
Figures~\ref{fig:appx-recipe-winner-depth} and
\ref{fig:appx-recipe-winner-tier}.}
\label{fig:appx-recipe-winner-9regions}
\end{figure*}

Complementing the recipe-design view in
Section~\ref{sec:recipe-design}, we report \emph{which} RL recipe achieves
the highest pass@128 in each (task, region) cell, treating the recipe choice
itself as the free axis. Figures~\ref{fig:appx-recipe-winner-depth},
\ref{fig:appx-recipe-winner-tier}, and
\ref{fig:appx-recipe-winner-9regions} show the per-cell winner for the
depth axis (\(D=1,\ldots,10\)), the complexity axis (\(T=1,\ldots,6\)), and
the nine depth-band by tier-band regions, respectively. Each cell is
colored by the winning RL recipe and annotated with that recipe's
pass@128 score in the corresponding scope.

Three patterns are visible across the three views; see also
Appendix~\ref{appx:inductive-analogy-experiments} for the
correlation-based view of the same task-family pattern.
First, the winner is \emph{task-dependent}: deductive and abductive favor
Cmplx-\(\ast\), Depth-\(\ast\), or Mix recipes depending on the region, while
inductive and analogy keep high pass@128 under several recipe families. Second,
the winner shifts along each axis: for deductive and abductive, Cmplx-Uniform
is strongest in several shallow-depth or low-complexity regions, but Depth- and
Mix-family recipes appear more often in harder cells; the absolute pass@128
also drops sharply with combined difficulty (deductive D10 \(\approx 0.35\);
abductive D10 \(\approx 0.05\)). Third, inductive and analogy values stay in a
much narrower band across regions (typically \(0.78\)--\(1.00\)), consistent
with the smaller magnitudes already reported in
Appendix~\ref{appx:inductive-analogy-experiments}; for these two families, the
winner often changes while the score changes only by a few percentage points.
Together these views support recipe selection as a region- and task-conditional
decision rather than a single best-overall recipe.

\subsection{Curriculum Supplementary Experiments}
\label{appx:curriculum-supplementary-results}

This appendix collects the curriculum analyses that complement
Section~\ref{sec:curriculum}. The main text reports the fixed-budget granularity
result; here we provide the curriculum-order and replay settings, the full
deductive/abductive trajectories, the full-grid heatmaps, and the inductive/analogy
extensions.

\subsubsection{Task Setting}
\label{appx:curriculum-order-replay-setting}

\paragraph{Task 1: Curriculum Order} On both axes, we run a 3-block
hard-to-easy (H2E) control against the easy-to-hard (E2H) curriculum.
E2H starts with the easiest cells (low \(D\) or low \(T\)) and ends with
the hardest; H2E reverses the order.

\paragraph{Task 2: Historical-Data Replay} We add 20\% historical-data
replay to the 3-block curriculum and compare it with the matched 3-block
no-replay baseline. The replay sample is drawn uniformly from already-seen
blocks, so under an E2H schedule it concentrates on easier cells.

\subsubsection{Observations}
\label{appx:curriculum-da-trajectories}

\begin{figure*}[!t]
\centering
\includegraphics[width=0.88\textwidth]{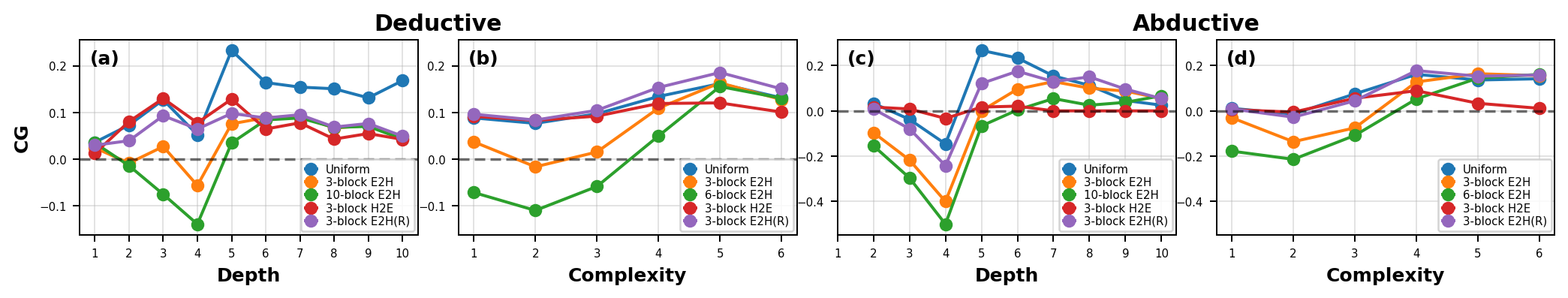}
\caption{Curriculum CG trajectories for deductive and abductive on the depth and
complexity axes. Within each task pair, the left panel shows depth schedules and the
right panel shows complexity schedules. Higher curves indicate stronger CG.}
\label{fig:curriculum-lines}
\end{figure*}

Figure~\ref{fig:curriculum-lines} shows two additional patterns. First, curriculum
order changes where gains remain: E2H favors hard regions and H2E favors easy
regions, but neither matches uniform mixing overall. Second, replay most clearly
helps earlier or easier regions, especially \(D \le 5\) on the depth axis or
\(T \le 3\) on the complexity axis; in harder regions, replay and no-replay are
closer and less consistently separated.

\subsubsection{Discussion}
\label{appx:curriculum-discussion-section}

\paragraph{Curriculum order changes retained regions but not global coverage.}
SGD-style optimization exhibits a recency bias: the final gradient updates
exert the strongest influence on the trained policy because no subsequent step
overwrites them. Easy-to-hard (E2H) curricula end on the hardest block and
preserve hard-region capability; hard-to-easy (H2E) curricula end on the
easiest block and preserve easy-region capability. The directional pattern is
a property of the optimizer rather than of the data, and curriculum order
cannot substitute for coverage. Recent E2H curriculum work on LLM RL
\citep{parashar2026curriculum} attributes the hard-task gain to scaffolding; our
data instead support a recency-based attribution, since uniform mixing---which
contains no scaffolding phase---outperforms E2H on the global grid.
Reverse-curriculum studies \citep{florensa2017reverse} report H2E advantages
in sparse-reward robotic RL through near-goal-state exposure, a mechanism that
does not transfer to dense-reward RLVR. Self-paced LLM RL
\citep{do-etal-2026-space} adaptively selects training data based on difficulty
estimates; in our framework this corresponds to a retention knob rather than
a coverage substitute.

\paragraph{Replay benefits easier regions more reliably than harder regions.}
A uniform-time replay buffer draws from already-seen blocks, which under an
E2H curriculum correspond to easier cells. The replayed 20\% therefore acts
primarily as a retention signal on those cells, while harder current-block
cells receive the same 80\% update stream as the no-replay baseline. Replay
therefore behaves more like a retention mechanism than a direct acquisition
mechanism: easier regions improve when forgetting is the main bottleneck,
whereas harder regions show smaller and less consistent changes when the
bottleneck is acquiring new behavior. Recent LLM-RL replay methods---RLEP \citep{zhang2025rlep},
Retrospective Replay \citep{dou2025improving}, ExGRPO
\citep{zhan2026exgrpo}, and EFRame \citep{wang2025eframe}---all report
accuracy gains on mathematical reasoning, but each employs prioritized or
filtered replay (verified successes, exploration-promising trajectories, or
correctness-weighted experience). Our analysis predicts that such
prioritization should accelerate hard-region learning; our deliberately
uniform-time design isolates the retention-only contribution. Classical
experience replay \citep{lin1992self,mnih2015human} is foundational to these
methods and aligns with our retention-based reading.

\subsubsection{Full-Grid and Task-Family Extensions}
\label{appx:curriculum-full-grid}

\begin{figure*}[!t]
\centering
\includegraphics[width=0.88\textwidth]{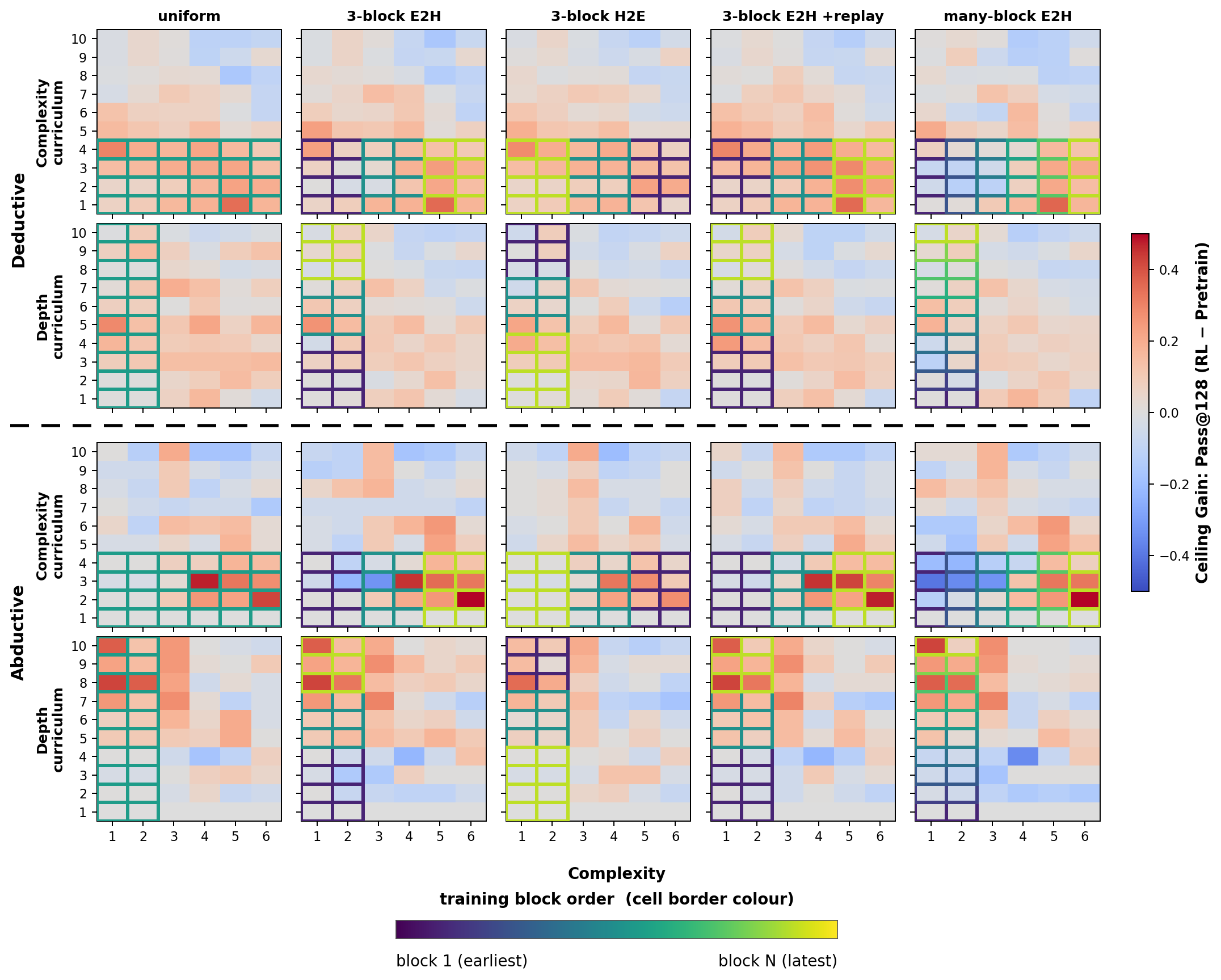}
\caption{Full-grid CG heatmaps for the depth- and complexity-axis curricula on
deductive (top two rows) and abductive (bottom two rows). Within each task block,
the top sub-row is the complexity-axis curriculum and the bottom sub-row is the
depth-axis curriculum; the five columns are uniform, 3-block E2H, 3-block H2E,
3-block E2H+replay, and many-block E2H. Cell borders are viridis-colored by
training block index (block 1 \(\rightarrow\) block N, dark purple \(\rightarrow\)
yellow). The color scale is shared across the two task families, and a bold
dashed line separates the two task blocks.}
\label{appx:curriculum-heatmaps-fig}
\end{figure*}

\begin{figure*}[!t]
\centering
\includegraphics[width=0.88\textwidth]{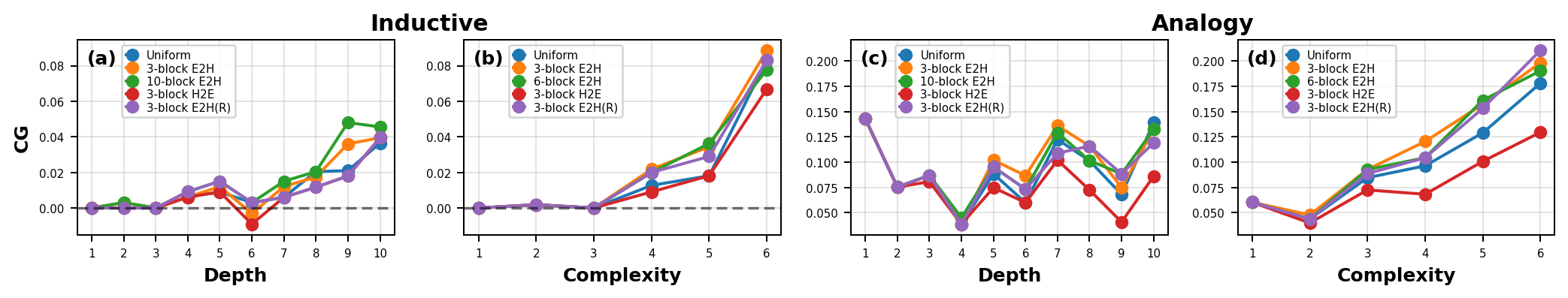}
\caption{Curriculum CG trajectories for inductive (left two panels) and analogy
(right two panels) along the depth and complexity axes. Same layout and conventions as
the deductive/abductive trajectory figure (Figure~\ref{fig:curriculum-lines}). The
inductive and analogy trajectories show the same qualitative curriculum and replay
patterns as deductive and abductive, with smaller magnitudes.}
\label{appx:curriculum-lines-fig}
\end{figure*}

The full-grid curriculum heatmaps in Figure~\ref{appx:curriculum-heatmaps-fig}
show where each curriculum variant improves or suppresses CG across the full
\((D,T)\) grid for deductive and abductive tasks. They clarify that curriculum
ordering changes retained regions, while uniform mixing remains the stronger
global default. Figure~\ref{appx:curriculum-lines-fig} extends the trajectory
analysis to inductive and analogy tasks. Together with
Figure~\ref{fig:curriculum-lines}, these results show that the granularity
ordering (uniform \(\ge\) 3-block \(\ge\) many-block) and the
replay-helps-easier-regions tendency hold as global patterns across all four
task families. The
effect is less pronounced for inductive and analogy because their CG range is
narrower, leaving less ceiling headroom for curriculum order to separate
retained from non-retained cells.

\end{document}